\PassOptionsToPackage{dvipsnames}{xcolor}

\documentclass[numbers]{assets/matterlab}

\usepackage{microtype}
\usepackage{graphicx}
\usepackage{booktabs}
\usepackage{float}
\usepackage{xurl}
\usepackage{hyperref}
\usepackage{titlesec}
\usepackage{tabularx}
\usepackage{chemformula}
\usepackage{siunitx}

\usepackage{amsmath}
\usepackage{amssymb}
\usepackage{mathtools}
\usepackage{amsthm}
\usepackage{bm}

\usepackage[noabbrev,nameinlink]{cleveref}

\usepackage{amsmath,amsfonts,bm}

\def\eqref#1{equation~\ref{#1}}

\def\1{\bm{1}}

\DeclareMathAlphabet{\mathsfit}{\encodingdefault}{\sfdefault}{m}{sl}
\SetMathAlphabet{\mathsfit}{bold}{\encodingdefault}{\sfdefault}{bx}{n}

\usepackage{orcidlink}
\usepackage{comment}
\usepackage[version=4]{mhchem}
\usepackage{longtable}
\usepackage{array}
\renewcommand{\arraystretch}{1.7}
\usepackage{multirow}           
\usepackage{amsfonts}           
\usepackage{nicefrac}
\usepackage{duckuments}         
\usepackage{thmtools,thm-restate}
\usepackage{enumitem}
\usepackage[dvipsnames]{xcolor}
\usepackage{colortbl}
\usepackage{tikz}
\usepackage{tikz-cd}
\usepackage{caption}
\usepackage{subcaption}
\usepackage{listings}
\usepackage{gensymb}
\usepackage{textcomp} 
\usepackage{newunicodechar} 
\newunicodechar{Δ}{$\Delta$}
\newunicodechar{∆}{$\Delta$}
\newunicodechar{−}{\textminus}
\newunicodechar{₀}{$_{0}$}
\newunicodechar{₁}{$_{1}$}
\newunicodechar{₂}{$_{2}$}
\newunicodechar{₃}{$_{3}$}
\newunicodechar{₄}{$_{4}$}
\newunicodechar{₅}{$_{5}$}
\newunicodechar{₆}{$_{6}$}
\newunicodechar{₇}{$_{7}$}
\newunicodechar{₈}{$_{8}$}
\newunicodechar{₉}{$_{9}$}

\newunicodechar{ₐ}{$_{a}$}
\newunicodechar{ₑ}{$_{e}$}
\newunicodechar{ₒ}{$_{o}$}
\newunicodechar{ₓ}{$_{x}$}
\newunicodechar{ₕ}{$_{h}$}
\newunicodechar{ₖ}{$_{k}$}
\newunicodechar{ₗ}{$_{l}$}
\newunicodechar{ₘ}{$_{m}$}
\newunicodechar{ₙ}{$_{n}$}
\newunicodechar{ₚ}{$_{p}$}
\newunicodechar{ₛ}{$_{s}$}
\newunicodechar{ₜ}{$_{t}$}
\newunicodechar{₋}{$_{-}$}
\newunicodechar{₍}{$_{(}$}
\newunicodechar{₎}{$_{)}$}
\newunicodechar{⁻}{$^{-}$}
\newunicodechar{α}{$\alpha$}
\newunicodechar{β}{$\beta$}
\newunicodechar{ω}{$\omega$}
\newunicodechar{σ}{$\sigma$}
\newunicodechar{≤}{$\leq$}
\newunicodechar{≥}{$\geq$}
\newunicodechar{λ}{$\lambda$}
\newunicodechar{≈}{$\approx$}
\newunicodechar{≡}{$\equiv$}

\newunicodechar{ }{\,}
\newunicodechar{─}{\textemdash}
\newunicodechar{π}{$\pi$}
\usepackage{fontawesome5}
\usepackage[toc,page,header]{appendix}
\usepackage{minitoc}

\newcommand{\addressCHEM}{Department of Chemistry, University of Toronto,  80 St. George St., Toronto, ON M5S 3H6, Canada}
\newcommand{\addressAC}{Acceleration Consortium, 700 University Ave., Toronto, ON M7A 2S4, Canada}
\newcommand{\addressCS}{Department of Computer Science, University of Toronto, 40 St George St., Toronto, ON M5S 2E4, Canada}
\newcommand{\addressVECTOR}{Vector Institute for Artificial Intelligence, W1140-108 College St., Schwartz Reisman Innovation Campus, Toronto, ON M5G 0C6, Canada}
\newcommand{\addressMSE}{Department of Materials Science \& Engineering, University of Toronto, 184 College St., Toronto, ON M5S 3E4, Canada}
\newcommand{\addressCHEMENG}{Department of Chemical Engineering \& Applied Chemistry, University of Toronto, 200 College St., Toronto, ON M5S 3E5, Canada}
\newcommand{\addressMEDICALSCI}{Institute of Medical Science, 1 King's College Circle, Medical Sciences Building, Room 2374, Toronto, ON M5S 1A8, Canada}
\newcommand{\addressCIFAR}{Canadian Institute for Advanced Research (CIFAR), 661 University Ave., Toronto,
ON M5G 1M1, Canada}
\newcommand{\addressNVIDIA}{NVIDIA, 431 King St W \#6th, Toronto, ON M5V 1K4, Canada}

\newcommand{\addressMATH}{Department of Mathematics, University of Toronto, 40 St George St., Toronto, ON M5S 2E4, Canada}

\newcommand{\acknowAC}{This research is part of the University of Toronto’s Acceleration Consortium, which receives funding from the CFREF-2022-00042 Canada First Research Excellence Fund. }
\newcommand{\acknowGEN}[1]{This research was enabled in part by support provided by #1 and the Digital Research Alliance of Canada (\url{https://www.alliancecan.ca}). }
\newcommand{\acknowSciNet}[1]{Computations were performed on the #1 supercomputer at the SciNet HPC Consortium. SciNet is funded by: the Canada Foundation for Innovation; the Government of Ontario; Ontario Research Fund - Research Excellence; and the University of Toronto. }

\newcommand{\acknowDARPA}{This work was supported by the Defense Advanced Research Projects Agency (DARPA) under Agreement No.~HR0011262E022. }

\providecommand{\doi}[1]{\href{https://doi.org/#1}{doi:#1}}

\usepackage{soul, color}
\newcommand{\Note}[1]{}
\renewcommand{\Note}[1]{#1}  

\newcommand{\jiaru}[1]{\colorbox{cyan!25}{\strut \textbf{Responsible: Jiaru}}}
\newcommand{\kenko}[1]{\colorbox{yellow!35}{\strut \textbf{Responsible: Kenko}}}
\newcommand{\thomas}[1]{\colorbox{violet!20}{\strut \textbf{Responsible: Thomas}}}
\newcommand{\yeonghun}[1]{\colorbox{green!25}{\strut \textbf{Responsible: Yeonghun}}}
\newcommand{\choi}[1]{\colorbox{orange!30}{\strut \textbf{Responsible: Changhyeok}}}

\usepackage{array}

\newcolumntype{P}[1]{>{\raggedright\arraybackslash}p{#1}}

\usepackage{lmodern}
\usepackage{cinzel}
\usepackage{quantikz}
\usepackage{adjustbox}
\usepackage{wrapfig}
\usepackage{booktabs}
\usepackage{tabularx}
\usepackage{xcolor}

\usepackage[most]{tcolorbox}
\usepackage{titlesec}
\usepackage{braket}
\usepackage[nolist]{acronym}
\usepackage{needspace}
\usepackage{titletoc}
\definecolor{LightGray1}{rgb}{0.97,0.97,0.97}
\usepackage{listings}

\lstdefinelanguage{SPARQL}{
  basicstyle=\footnotesize\ttfamily,
  backgroundcolor=\color{LightGray1},
  columns=fullflexible,
  breaklines=true,
  sensitive=true,
  frame=bt,
  aboveskip=1em,
  belowskip=1em,
  xleftmargin=.5em,
  xrightmargin=.5em,
  framexleftmargin=.5em,
  framextopmargin=.5em,
  framexbottommargin=.5em,
  framexrightmargin=.5em,
  tabsize = 2,
  showstringspaces=false,
  morecomment=[l][\color{gray}]{\#},       
  morecomment=[n][\color{blue}]{<http}{>}, 
  morestring=[b][\color{OliveGreen}]{\"},  
  keywordsprefix=?,
  classoffset=0,
  keywordstyle=\color{Sepia},
  morekeywords={},
  classoffset=1,
  keywordstyle=\color{Purple},
  morekeywords={purl,rdf,rdfs,skos,xsd,time,prov,om,owl,saref,yago,dbo,dbr,OntoCAPE_Behavior,OntoCAPE_Material,OntoCAPE_Phase_System,OntoCAPE_Reaction_Mechanism,OntoCAPE_Substance,OntoCAPE_System,OntoSpecies,OntoKin,OntoReaction,OntoDoE,OntoLab,OntoVapourtec,OntoHPLC,OntoDerivation,OntoGoal,OntoAgent,dc,mofs,grafico},
  classoffset=2,
  keywordstyle=\color{MidnightBlue},
  morekeywords={
    SELECT,CONSTRUCT,DESCRIBE,ASK,WHERE,FROM,NAMED,PREFIX,BASE,OPTIONAL,VALUES,DISTINCT,
    FILTER,GRAPH,LIMIT,OFFSET,SERVICE,UNION,EXISTS,NOT,BINDINGS,MINUS,a
  }
}

\tcbset{textmarker/.style={%
    enhanced,
    breakable,
    before=\par\Needspace{12\baselineskip}, 
    frame hidden,
    parbox=false,
    boxrule=0mm,
    boxsep=0mm,
    arc=3mm,          
    outer arc=3mm,    
    left=3mm,right=3mm,
    top=7pt,bottom=7pt,
    toptitle=1mm,bottomtitle=1mm,
    left skip=0.2cm,
    fontupper=\footnotesize
}}

\tcbset{textmarker2/.style={%
    enhanced,
    breakable,
    before=\par\Needspace{12\baselineskip}, 
    frame hidden,
    parbox=false,
    boxrule=0mm,
    boxsep=0mm,
    arc=3mm,          
    outer arc=3mm,    
    left=3mm,right=3mm,
    top=7pt,bottom=7pt,
    toptitle=1mm,bottomtitle=1mm,
    left skip=0.8cm,
    fontupper=\footnotesize
}}

\newtcolorbox{hintBox}{textmarker,
    colback=blue!10!white,
    fontupper=\footnotesize
}

\newtcolorbox{hintBox2}{textmarker2,
    colback=blue!5!white,
    fontupper=\footnotesize
}

\newtcolorbox{potenteLongBox}{textmarker,
  parbox=true,
  colback=blue!10!white,
  fontupper=\footnotesize,
  title={\faRobot\ \underline{\textbf{\potente{}}}},
  title after break={\faRobot\ \underline{\textbf{\potente{} (continued)}}},
  colbacktitle=blue!10!white,
  coltitle=black,
  fonttitle=\footnotesize
}

\newtcolorbox{thoughtBox}{
  fontupper=\footnotesize,
  colback=green!3!white,
  colframe=blue!5!white,
  boxrule=0.5pt
}

\newtcolorbox{importantBox}{textmarker,
    colback=red!10!white,
    fontupper=\footnotesize
}

\newtcolorbox{noteBox}{textmarker,
    colback=matterbg,
    fontupper=\footnotesize
}

\newtcolorbox{noteBoxSmall}{%
    textmarker,
    colback=matterbg,
    fontupper=\footnotesize
}

\newcommand{\prompt}[1]{\begin{noteBoxSmall} \underline{\textbf{Prompt}} #1 \end{noteBoxSmall}}

\newcommand{\potentereslong}[1]{\begin{potenteLongBox} #1 \end{potenteLongBox}}

\newcommand{\elagente}{{\cinzel El Agente}}
\newcommand{\elagenteQ}{{\cinzel El Agente Q}}
\newcommand{\elagenteG}{{\cinzel El Agente Gr\'afico}}
\newcommand{\elagenteP}{{\cinzel El Agente Potente}}

\newcommand{\potente}{{\cinzel Potente}}
\newcommand{\elagenteQuan}{{\cinzel El Agente Cu\'antico}}
\newcommand{\elagenteE}{{\cinzel El Agente Estructural}}
\newcommand{\elagenteSo}{{\cinzel El Agente S\'olido}}
\newcommand{\elagenteQun}{{\cinzel El Agente Quntur}}
\newcommand{\elagenteF}{{\cinzel El Agente Forjardor}}
\newcommand{\elagenteS}{{\cinzel El Agente Seguro}}

\title{El Agente Potente: High-Throughput Agentic Atomistic Simulations}

\author[1,2,\dagger,\orcidlink{0000-0002-0802-9559}]{Tsz~Wai~Ko}
\author[1,2,\dagger,\orcidlink{0000-0002-1246-1993}]{Jiaru~Bai}
\author[3,4,\dagger,\orcidlink{0009-0009-1341-5831}]{Thomas~Swanick}
\author[1,2,\orcidlink{0000-0001-5191-5735}]{Yeonghun~Kang}
\author[1,2,\orcidlink{0009-0003-1903-2184}]{Changhyeok~Choi}
\author[3]{Angelina Qihong Jiang}
\author[3,\orcidlink{0009-0004-6626-6714}]{Aiwei~Yin}
\author[1,3,8,*,\orcidlink{0000-0002-8446-7956}]{Varinia~Bernales}
\author[1,2,3,5,6,7,8,9,10,*,\orcidlink{0000-0002-8277-4434}]{Al\'an~Aspuru-Guzik}

\affiliation[1]{\addressCHEM}
\affiliation[2]{\addressVECTOR}
\affiliation[3]{\addressCS}
\affiliation[4]{\addressMATH}
\affiliation[5]{\addressMSE}
\affiliation[6]{\addressCHEMENG}
\affiliation[7]{\addressMEDICALSCI}
\affiliation[8]{\addressAC}
\affiliation[9]{\addressCIFAR}
\affiliation[10]{\addressNVIDIA}

\contribution[\dagger]{Contributed equally to this work.}

\abstract{
Foundational \acp{mlip} are transforming atomistic simulations by achieving near-\textit{ab initio} accuracy across large chemical spaces at a fraction of the computational cost.
A central challenge in using these tools for high-throughput property calculations is translating high-level scientific intent into adaptive simulation campaigns without compromising workflow rigour.
We introduce \elagenteP{}, an agentic system that combines typed execution graphs with a complementary coding mode for \acp{mlip}-driven atomistic simulations. Typed execution graphs provide structured and provenance-aware execution for standardized workflows, with \acp{llm} restricted to planning and routing while deterministic Python components perform scientific computation and validation. Complementing this structured execution, a coding agent constructs customized workflows for tasks requiring greater procedural flexibility while invoking existing \potente{} functions for supported calculations. We demonstrate \elagenteP{} across computational materials discovery, molecular energy-landscape exploration, adsorption, and catalytic reaction workflows, together with systematic benchmarks of reproducibility and \ac{llm} token cost. These results establish typed execution graphs and code-based workflow construction as complementary mechanisms for agentic scientific computing, combining controlled, auditable execution with the flexibility required for customized atomistic simulations.

}

\date{\today}
\correspondence{
\email{varinia@bernales.org} and \email{alan@aspuru.com}}

\begin{document}

\acrodef{dft}[DFT]{density functional theory}
\acrodef{dftb}[DFTB]{density functional tight-binding}
\acrodef{td-dft}[TD-DFT]{time-dependent density functional theory}
\acrodef{mlip}[MLIP]{machine-learning interatomic potential}
\acrodef{wft}[WFT]{wave function theory}
\acrodef{hf}[HF]{Hartree-Fock}
\acrodef{qcg}[QCG]{quantum cluster growth}
\acrodef{td}[TD]{time-dependent}
\acrodef{tddft}[TDDFT]{time-dependent density functional theory}
\acrodef{md}[MD]{molecular dynamics}
\acrodef{rmsd}[RMSD]{root mean square deviation}
\acrodef{qc}[QC]{quantum chemistry}
\acrodef{mof}[MOF]{metal-organic framework}
\acrodef{DAC}[DAC]{direct air capture}
\acrodef{kg}[KG]{knowledge graph}
\acrodef{ogm}[OGM]{object graph mapper}
\acrodef{iri}[IRI]{internationalized resource identifier}

\acrodef{llm}[LLM]{large language model}
\acrodef{coala}[CoALA]{cognitive architectures for language agents}
\acrodef{ai}[AI]{artificial intelligence}
\acrodef{ttl}[TTL]{time-to-live}
\acrodef{api}[API]{application programming interface}

\maketitle


\newpage
\section{Introduction}

Machine learning interatomic potentials (\acp{mlip}) are rapidly expanding the accessible scale and scope of atomistic simulation by approximating \textit{ab initio} potential-energy surfaces at substantially reduced computational cost~\cite{behler2021machine,friederich2021machine,ko2023recent,zhang2025roadmap}. Recent models with broad chemical coverage~\cite{chen2022universal,deng2023chgnet,batatia2025foundation, ko2025fast} further enable a common computational infrastructure across diverse materials and molecular systems. However, deploying these foundational \acp{mlip} at scale requires more than efficient energy and force evaluation: practical studies must coordinate structure preparation, relaxation, sampling, property calculations, validation, failure handling, and data aggregation, often through workflows that differ substantially between scientific problems.

Most existing high-throughput workflow frameworks, such as AFLOW~\cite{curtarolo2012aflow}, AiiDA~\cite{huber2020aiida, uhrin2021workflows}, Pyiron~\cite{janssen2019pyiron, menon2024electrons}, and Atomate2~\cite{ganose2025atomate2}, provide powerful infrastructure for automating atomistic simulations. However, using these frameworks often requires substantial expertise across multiple simulation packages, as well as domain knowledge of computational materials science and chemistry. 
 Large language model (\ac{llm})-based scientific agents offer a complementary form of flexibility by translating high-level objectives into tool use, workflow decisions, or generated code~\cite{Pham2026ChemGraph,wang2025dreams,hu2025aitomia,Nduma2025Crystalyse,Liu2026CatGo,kim2026materealize}. Recent systems targeting atomistic simulation, including LAMMPS-Agent~\cite{vriza2026multi}, Masgent~\cite{liu2026masgent}, and AtomisticSkills~\cite{deng2026harnessing}, further demonstrate the potential of this approach. Yet greater runtime flexibility raises a complementary challenge: scientific autonomy must be introduced without sacrificing reproducibility, provenance, and explicit control over numerical execution.

Within the \elagente{} family, successive systems have explored different ways of introducing autonomy into scientific workflows. \elagenteQ{}~\cite{zou2025agente} and its successor \elagenteQun{}~\cite{perez2026agente}, as well as other members such as \elagenteSo{}~\cite{kumar2026agente}, \elagenteE{}~\cite{choi2026agente} and \elagenteQuan{}~\cite{gustin2026agente}, use hierarchical multi-agent reasoning to dynamically decompose and delegate computational tasks across quantum-chemistry, solid-state, and quantum simulations. \elagenteG{}~\cite{grafico} instead structures execution through typed execution graphs, while retaining flexibility through a coding agent that can construct task-specific graphs. \elagenteF{}~\cite{zhang2026agente} extends adaptability to the capability layer by dynamically generating and refining computational tools. Building on \elagenteG{}, \elagenteS{}~\cite{kang2026seguro} embeds specialized agents directly within structured execution graphs, combining multi-agent reasoning with explicit control over scientific state and execution. Together, these systems explore how to introduce flexibility at the levels of reasoning, workflow construction, and tool construction while preserving reliable scientific execution.

Here, we introduce \elagenteP{}, an agentic framework for \ac{mlip}-driven atomistic simulations that integrates structured workflow execution with programmable agentic flexibility. Rather than relying exclusively on either predefined workflows or open-ended code generation, \potente{} supports two complementary execution modes. Standardized scientific procedures are represented as typed execution graphs, providing explicit control over data flow, validation, branching, provenance, and concurrent execution as intermediate results become available. For tasks requiring customized procedural logic, a coding agent constructs task-specific Python workflows and invokes existing \potente{} functions when the requested calculations are already supported. This hybrid design combines controlled and auditable execution with the flexibility to adapt workflow logic beyond predefined graph components.

We assess \elagenteP{} through three systematic benchmarks examining reproducibility, the performance of foundational \acp{mlip} and \ac{llm} token cost (including comparisons with open-ended skill-based agents), and seven representative scientific case studies. These span high-throughput property evaluation, composition-driven materials discovery, molecular energy-landscape exploration, adsorption, and catalytic reaction pathways, providing progressively more demanding tests of structured execution, adaptive decision-making, and customized workflow construction.

\begin{figure}
    \centering
    \includegraphics[width=\linewidth]{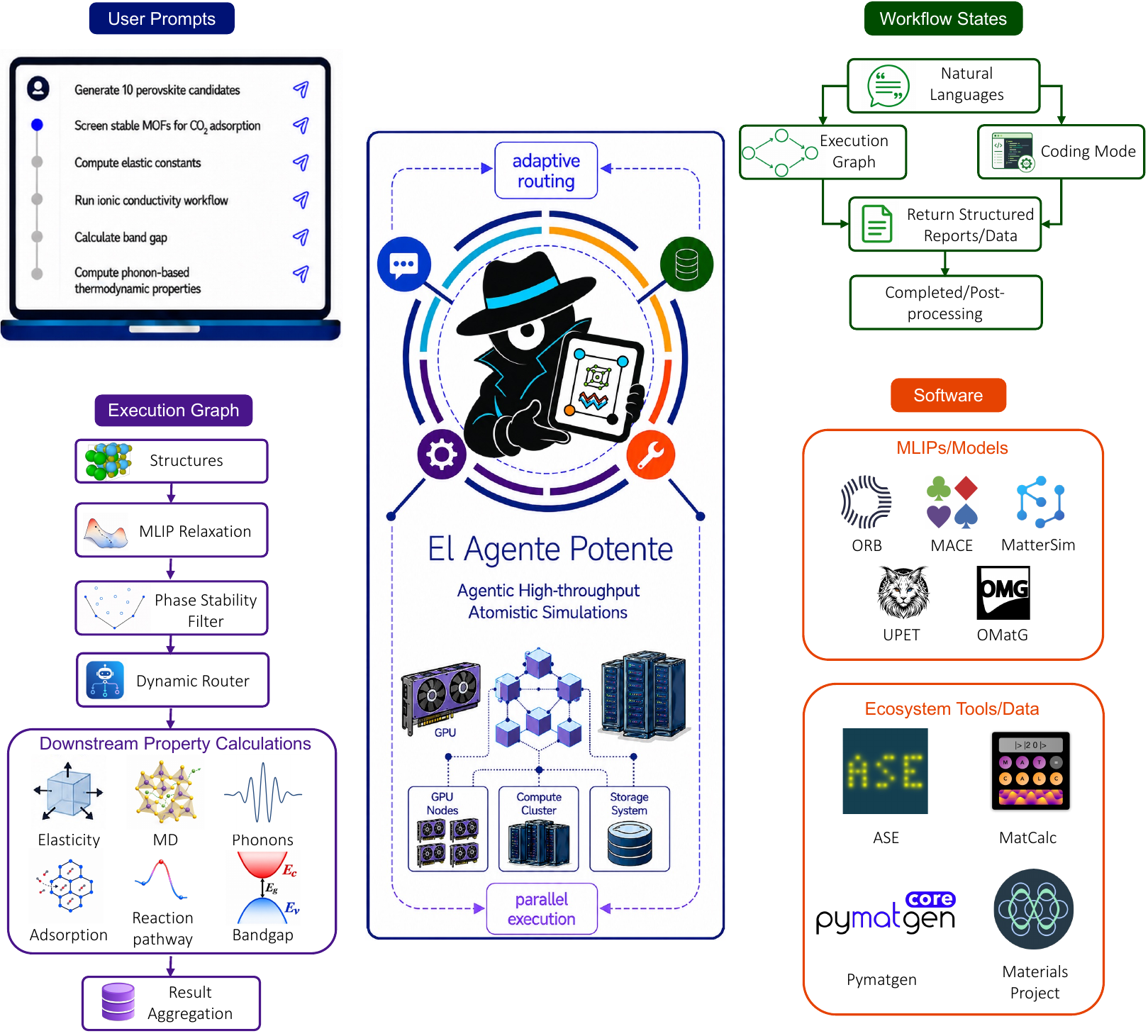}
    \caption{\textbf{Architecture of \elagenteP{}:} Natural-language scientific requests are adaptively routed to reusable computational capabilities for MLIP-driven atomistic simulation. Standardized high-throughput tasks use a typed execution graph supporting parallel structure processing, relaxation, routing, and downstream property calculations, while customized tasks can be implemented through agent-generated Python workflows that compose existing functions, tools, and the execution graph itself. Structured results and provenance are returned for downstream analysis and reuse.
    }
    \label{fig:potente_architecture}
\end{figure}

\section{Results}

\subsection{Architecture of \elagenteP{}}

Figure~\ref{fig:potente_architecture} summarizes the architecture of \potente{}, an agentic framework for \ac{mlip}-driven atomistic simulations across molecules, bulk materials, and interfaces. Given a natural-language scientific objective, \potente{} plans and orchestrates computational operations using a hierarchy of reusable capabilities. Its central design principle separates agentic scientific reasoning and workflow composition from validated numerical execution, allowing the agent to adapt task orchestration without regenerating established computational procedures.

A central capability exposed to \potente{} is a typed execution graph for standardized high-throughput simulation campaigns. The graph coordinates computational nodes for structure retrieval or generation, \ac{mlip}-based relaxation, optional filtering, downstream property calculations, and result aggregation. Within the graph, an \ac{llm}-based router can make constrained scientific decisions, such as selecting downstream calculations from the user objective and structural metadata, while deterministic Python functions perform the underlying simulations and data processing.

The graph is designed to execute many related calculations efficiently. Mapping operations fan computations across structures, streaming lets downstream tasks begin as soon as their dependencies are satisfied, and join operations synchronize results only when required. Outputs, routing and filtering decisions, failures, simulation artifacts, and provenance are retained as structured workflow state, with selected semantic provenance optionally persisted to a knowledge graph.

When a task requires customized procedural logic, \potente{} can instead construct and execute a task-specific Python workflow. The generated code can compose reusable \potente{} functions, external ecosystem tools, and the execution graph itself, allowing the agent to customize control flow, iteration, data dependencies, and post-processing without reimplementing established numerical procedures.

This layered architecture allows \potente{} operate at different levels of abstraction, invoking complete high-throughput workflows when appropriate while composing lower-level capabilities when greater procedural flexibility is needed. Detailed graph topology, supported \ac{mlip} models, reusable simulation functions, and parallel execution implementation are provided in the Supplementary Information, Sec.~\nameref{si_sec:execution_graph}.

\subsection{Benchmarks}
\subsubsection{Reproducibility of high-throughput calculations}

\begin{figure}
    \centering
    \includegraphics[width=1.0\linewidth]{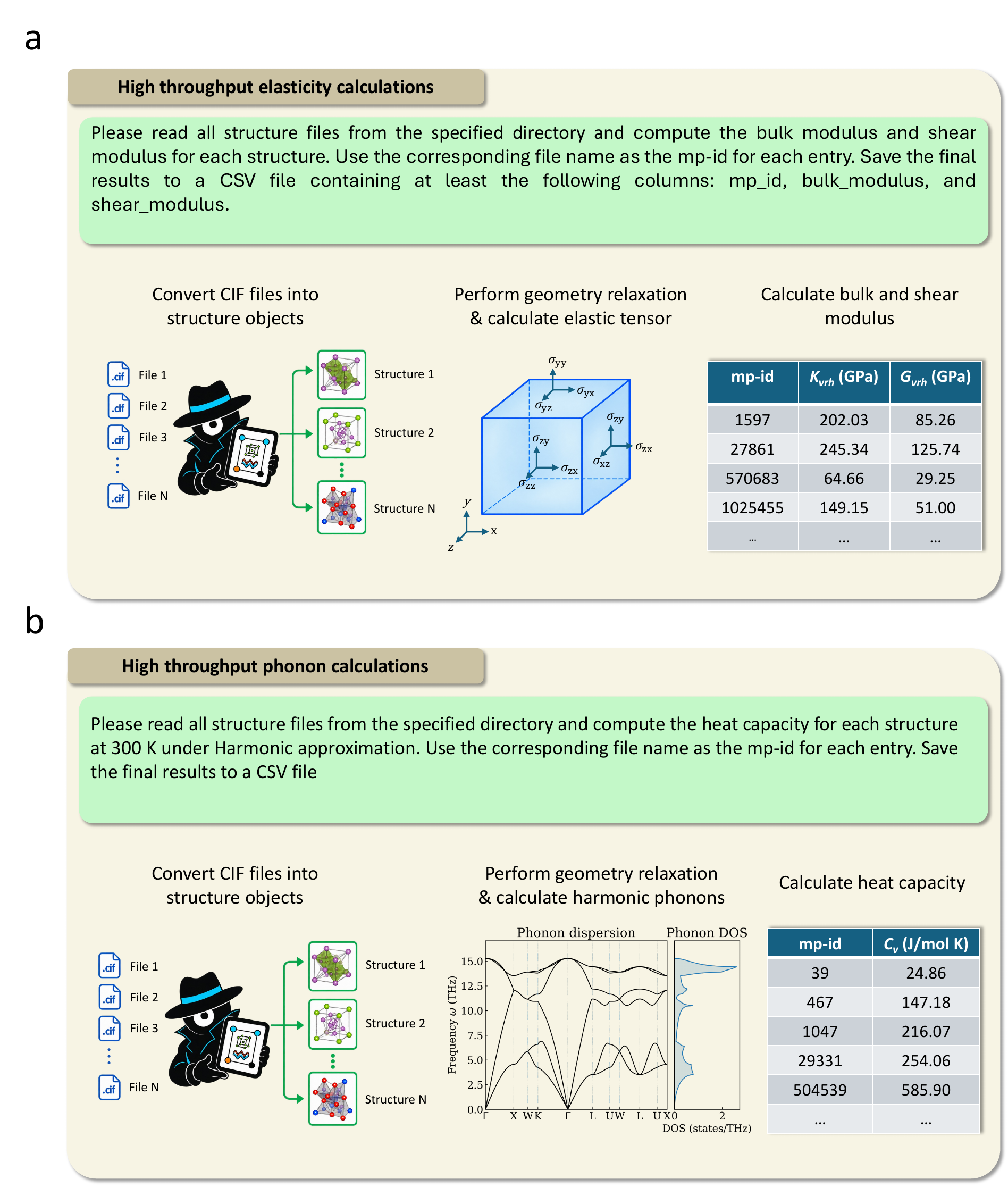}
    \caption{\textbf{Agentic workflows for high-throughput property calculations.} Examples of \elagenteP{} workflows for automated high-throughput property prediction. a, Elasticity workflow in which CIF files are converted into structure objects, relaxed with an MLIP, and used to compute elastic tensors, bulk moduli, and shear moduli. b, Phonon workflow in which structures are relaxed, harmonic phonon calculations are performed, and constant-volume heat capacities are computed at 300 K. In both examples, natural-language instructions are converted into executable workflows that process multiple structures and export the final results as CSV files.}
    \label{fig:high_throughput_prompts}
\end{figure}

To evaluate the reproducibility of high-throughput execution, we considered two workflows applied to 500 structures retrieved from the Materials Project~\cite{dunn2020benchmarking}. Both workflows used the OMAT head of the multi-head MACE model (MACE-OMAT)~\cite{batatia2025cross}. The first workflow calculated the Voigt--Reuss--Hill bulk and shear moduli ($K_{\mathrm{VRH}}$ and $G_{\mathrm{VRH}}$), whereas the second calculated the harmonic constant-volume heat capacity ($C_V$) at 300~K. For each task, we compared the \elagenteP{} execution graph with a manually scripted Python workflow using identical input structures, \ac{mlip}, and computational settings. Each workflow was executed independently five times.

In the elasticity workflow (Figure~\ref{fig:high_throughput_prompts}a), \potente{} reads the structures directly from the input directory, relaxes each structure, and evaluates the elastic tensor from stress--strain calculations to obtain $K_{\mathrm{VRH}}$ and $G_{\mathrm{VRH}}$. In the phonon workflow (Figure~\ref{fig:high_throughput_prompts}b), the relaxed structures are passed to harmonic phonon calculations from which $C_V$ is evaluated at 300~K. Across the five repeated executions, the YbZn$_2$P$_2$ structure (Materials Project ID mp-9582) was the only structure that occasionally failed in both the manually scripted and \elagenteP{} workflows. The failure occurred during Niggli reduction of the MLIP-relaxed unit cell, a tolerance-sensitive crystallographic standardization procedure~\cite{grossekunstleve2004numerically}. When a relaxed lattice lies close to a reduction boundary, small numerical variations in its lattice parameters can cause the reduction algorithm to fail, rather than indicating a failure of the \ac{mlip} force evaluation or harmonic phonon calculation. In \elagenteP{}, such cases are returned as typed failures and retained for subsequent inspection rather than interrupting the remaining high-throughput campaign.

We first quantified run-to-run reproducibility by calculating, for each material, the standard deviation of $K_{\mathrm{VRH}}$, $G_{\mathrm{VRH}}$, and $C_V$ across the five independent executions. The manually scripted workflow shows narrowly distributed standard deviations for all three quantities (Figure~\ref{fig:human_repro_histogram}), with most materials showing only small numerical variations between repeated runs. The same analysis for the \elagenteP{} execution graph yields comparably narrow distributions (Figure~\ref{fig:potente_repro_histogram}). The median and 95th-percentile standard deviations remain small relative to the overall ranges of the corresponding properties, indicating that repeated agentic execution does not introduce appreciable variability into the underlying numerical calculations.

We next compared the scientific outputs produced by the two approaches. For each material, the values of $K_{\mathrm{VRH}}$, $G_{\mathrm{VRH}}$, and $C_V$ were averaged over the five executions of each workflow. The resulting parity plots (Figure~\ref{fig:parity_plots_high_throughout}) lie essentially along the $y=x$ line across the full property ranges, showing that \elagenteP{} reproduces the results of the manually scripted reference workflow while autonomously constructing and executing the corresponding high-throughput calculation graph.

Together, these benchmarks show that the typed execution graph preserves the numerical reproducibility of a conventional scripted workflow while providing automated workflow construction, structured failure handling, and provenance capture. The agreement between repeated executions and between independently implemented workflow interfaces is particularly important for agentic high-throughput simulations, where adaptive planning should not compromise the determinism of the underlying scientific calculations.

\subsubsection{Agentic benchmarking of foundational machine learning interatomic potentials}

Having established the reproducibility of the execution-graph workflow, we next examined whether \elagenteP{} could support systematic benchmarking of different pretrained foundational \acp{mlip}. These models differ in architecture, training data, and chemical coverage, so their accuracy can vary substantially across downstream properties. Reliable comparison under identical simulation settings is consequently important before deploying a foundational \ac{mlip} in large-scale screening campaigns.

For this benchmark, we evaluated MatterSim-v1.0.0-1M (MatterSim)~\cite{yang2024mattersim}, conservative Orb-V3-OMAT (Orb-OMAT)~\cite{rhodes2025orb}, and MACE-OMAT~\cite{batatia2025cross} on 1,000 bulk crystal structures with reference bulk and shear moduli. In contrast to the preceding benchmark, which used the predefined execution graph, here the user requested a comparative high-throughput benchmark, and \potente{} employed its coding mode. The coding agent constructed a Python workflow that iterated over the structures and models while invoking the existing model-loading, relaxation, and elasticity functions implemented in \potente{}. Thus, the agent customized the benchmarking logic without reimplementing the underlying numerical procedures. All 3,000 model--structure calculations completed successfully.

The three models exhibited similar performance for the Voigt--Reuss--Hill bulk modulus as shown in Figure~\ref{fig:elasticity_parity_plots}. MatterSim achieved the lowest mean absolute error (MAE) of 13.99~GPa, followed by MACE-OMAT at 15.69~GPa and Orb-OMAT at 16.88~GPa, with corresponding $R^2$ values of 0.873, 0.871, and 0.863, respectively. Larger differences emerged for the shear modulus. Orb-OMAT gave the lowest MAE of 13.94~GPa, closely followed by MatterSim (14.05~GPa) and MACE-OMAT (15.24~GPa). MatterSim nevertheless achieved the highest $R^2$ for the shear modulus (0.663), compared with 0.624 for Orb-OMAT and 0.439 for MACE-OMAT, indicating larger structure-dependent deviations for the latter.

This benchmark illustrates two aspects of \elagenteP{}. First, the coding mode can construct customized comparative workflows while reusing the same validated computational functions employed elsewhere in the framework. Second, even closely performing foundational \acp{mlip} can exhibit property-dependent differences in accuracy and ranking, emphasizing the importance of benchmarking candidate models for the specific observables and materials domains in which they will be deployed.

\subsubsection{Comparison with a skill-based coding agent}


To assess the practical trade-off between a ``fixed'' execution-graph workflow and a more open-ended skill-based coding agent, we compared \potente{} with Claude Code equipped with atomistic skills~\cite{deng2026harnessing}. Both systems were given the same high-level GCMC task for CO$_2$ adsorption in HKUST-1: read the input CIF structure, set up the CO$_2$ GCMC simulation with an \ac{mlip} backend, run 5,000 equilibration and 5,000 production steps at 298 K and 1 bar, and aggregate the adsorption results. We repeated the task five times for each system and compared the adsorption outputs against metrics that reflect execution behaviour.

\begin{table}[H]
\centering
\caption{\textbf{Benchmark comparison of agentic workflows for GCMC simulations.} CO$_2$ adsorption in HKUST-1 was evaluated using GCMC simulations at 298 K and 1 bar, with 5000 equilibration steps and 5000 production steps. The predictions from \potente{} and Claude Code equipped with AtomisticSkills~\cite{deng2026harnessing} were obtained using the same high-level GCMC task. Each workflow was repeated five times, and values are reported as mean $\pm$ standard deviation. The uptake and interaction energy evaluate adsorption predictions, while the token cost reflects the execution overhead of each agent trace. Detailed metrics see Supplementary Information, Sec.~\ref{si_subsec:atomisticskills}.}
\label{tab:skills}
\begin{tabular}{c|c|c|c}
\toprule
\textbf{Agent} & \textbf{Uptake (mmol/g)} & \textbf{Heat of adsorption (kJ/mol per CO$_2$)} & \textbf{Token cost (USD)} \\
\midrule
Claude Code with skills & 1.25 $\pm$ 0.57 & 13.81 $\pm$ 0.77 & 0.638 $\pm$ 0.139 \\
\potente{} & 1.88 $\pm$ 0.12 & 22.79 $\pm$ 0.53 & 0.152 $\pm$ 0.003 \\
\bottomrule
\end{tabular}
\end{table}

As summarized in Table~\ref{tab:skills}, both systems completed the HKUST-1 CO$_2$ GCMC task, but with distinct execution profiles. Because the requested simulations used only 5,000 production steps, interpret the adsorption values as screening-level predictions.  For context, experimental CO$_2$ uptakes of approximately 4.1--4.2~mmol~g$^{-1}$ have been reported for pristine HKUST-1 near 298--300~K and 1~bar ~\cite{vrtovec2020structural,chen2018high}. Reported isosteric heats of CO$_2$ adsorption for pristine HKUST-1 are approximately 20--30~kJ~mol$^{-1}$, depending on loading and measurement protocol~\cite{vrtovec2020structural,cortessuarez2019swcnt}.

These experimental values are provided only as physical context, because the two systems used different \ac{mlip} backends and GCMC sampling protocols, the comparison should be viewed as a workflow-level assessment and does not constitute a direct benchmark of adsorption accuracy. In \potente{}, the execution graph reduced runtime decision-making and produced reproducible adsorption estimates with stable resource usage. By contrast, skill-based Claude Code performed workflow construction in addition to running the calculation in each independent run. Only two of the five runs considered pre-relaxing the framework before GCMC (one relaxed it with the \ac{mlip} calculator, one explicitly checked and opted to skip it). This comparison points to a complementary use case: skill-based coding agents can accelerate workflow prototyping, while typed execution graphs can amortize this effort for repeated computation. Detailed discussions on each run can be found in SI~Sec.~\ref{si_subsec:atomisticskills}.

\subsection{Case studies}

We evaluated \elagenteP{} on seven representative research problems formulated through tailored user prompts, demonstrating its ability to execute robust atomistic simulations across diverse application scenarios.

\subsubsection{Materials Discovery}
\begin{figure}
    \centering
    \includegraphics[width=1.0\linewidth]{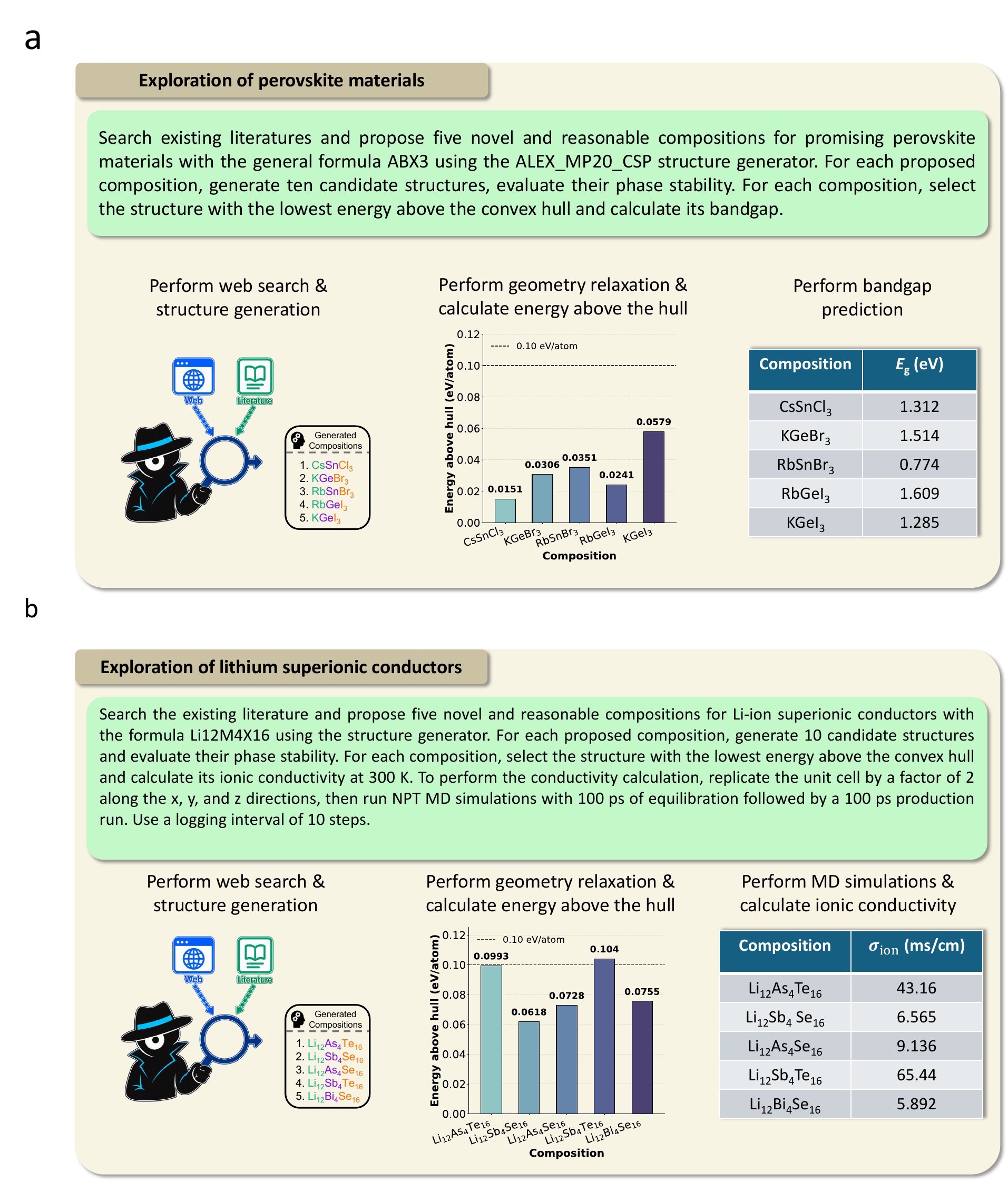}
    \caption{\textbf{Agentic workflows for materials discovery:} Exploration of perovskite materials \textbf{a} and lithium superionic conductors \textbf{b} using \acp{mlip} and generative models. In both cases, natural-language instructions are translated into executable pipelines that combine literature-guided composition proposal, structure generation, MLIP-based relaxation, phase-stability evaluation, and property prediction, including bandgap estimation for perovskites and MD-based ionic conductivity calculations for Li conductors. }
    \label{fig:discovery_prompts}
\end{figure}
Computational materials discovery often relies on high-throughput calculations to efficiently screen large chemical spaces and identify promising candidates. A central challenge is the end-to-end integration of the autonomous discovery workflow, from generating candidate structures to evaluating their thermodynamic stability and then performing downstream property calculations for targeted applications. For composition-driven discovery, thermodynamic screening can be performed using the energy above the convex hull. Competing phases in the corresponding chemical system are retrieved from the Materials Project~\cite{jain2013commentary,Horton2025accelerated} and evaluated using the same \ac{mlip} as the generated candidates, providing an internally consistent phase diagram. Candidates satisfying the user-defined stability criterion proceed to downstream calculations, whereas filtered or failed structures retain typed records describing their execution outcome. Downstream calculations are implemented as modular nodes spanning structural, vibrational, mechanical, electronic, dynamical, adsorption, and reaction-pathway properties. Because routing runs independently for each structure, different members of the same campaign can be assigned different combinations of calculations based on the scientific objective.

\paragraph{Perovskite discovery} 
In the first case study (see Figure~\ref{fig:discovery_prompts}a), via the user prompt, the user asked \potente{} to explore promising ABX$_3$ perovskite materials. The agent first performed a literature and web search on lead-free halide perovskites, motivated by the broader interest in replacing Pb-based halide absorbers with chemically reasonable Ge- and Sn-based alternatives \cite{mitro2022electronic, rahman2022rbsnx, rahman2025first, ayalew2026exploring}. Based on this search, the agent selected five charge-balanced alkali-metal halide perovskite candidates: \(\mathrm{KGeI}_3\), \(\mathrm{RbGeI}_3\), \(\mathrm{KGeBr}_3\), \(\mathrm{RbSnBr}_3\), and \(\mathrm{CsSnCl}_3\).

For each composition, \potente{} generated ten candidate structures using the OMatG model trained on the ALEX-MP20-CSP database~\cite{zeni2025generative}. The generated structures and competing phases in the corresponding chemical systems were then relaxed using MACE-OMAT~\cite{batatia2025cross}, which was fine-tuned across different domains and large datasets spanning bulk materials, molecules, and surfaces~\cite{barros2026open, hollmer2025open,chanussot2021open,kaplan2025foundational}. The relaxed energies were used to evaluate thermodynamic stability by computing the energy above the \ac{mlip} convex hull. For each composition, only the lowest-energy-above-hull structure was retained for electronic-property prediction. Band gaps were then predicted using PET-MAD-DOS~\cite{how2026universal}, a point-edge transformer~\cite{pozdnyakov2023smooth} trained on density-of-states data at the Perdew-Burke-Ernzerhof revised for solids (PBE-sol) level from the Massive Atomic Density dataset~\cite{mazitov2025pet}.

The selected candidates occupied a relatively narrow metastability window of \(0.0151\)--\(0.0579\)~eV~atom\(^{-1}\) above the \ac{mlip} convex hull, with predicted band gaps ranging from \(0.774\) to \(1.609\)~eV. \(\mathrm{CsSnCl}_3\) was the most stable candidate, lying only \(0.0151\)~eV~atom\(^{-1}\) above the hull with a predicted band gap of \(1.312\)~eV. The Ge-based halides also showed low hull distances and optoelectronically relevant gaps: \(\mathrm{RbGeI}_3\), \(\mathrm{KGeBr}_3\), and \(\mathrm{KGeI}_3\) were \(0.0241\), \(0.0306\), and \(0.0579\)~eV~atom\(^{-1}\) above hull, with predicted band gaps of \(1.609\), \(1.514\), and \(1.285\)~eV, respectively. \(\mathrm{RbSnBr}_3\) was also close to the hull, at \(0.0351\)~eV~atom\(^{-1}\), but had the smallest predicted band gap of \(0.774\)~eV.

Overall, these results illustrate how \potente{} integrates literature-guided composition proposal, structure generation, MLIP relaxation, convex-hull-based down-selection, and electronic-property prediction within a single reproducible workflow. The selected Ge- and Sn-based halide perovskites are qualitatively consistent with prior first-principles studies that identified \(\mathrm{KGeI}_3\), \(\mathrm{RbGeI}_3\), \(\mathrm{KGeX}_3\) \((X=\mathrm{Br},\mathrm{I})\), \(\mathrm{RbSnX}_3\) \((X=\mathrm{Cl},\mathrm{Br},\mathrm{I})\), and \(\mathrm{CsSnCl}_3\) as potential lead-free photovoltaic or optoelectronic candidates. We note that the predicted band gaps slightly deviate from reported \textit{ab initio} values~\cite{mitro2022electronic,rahman2025first,rahman2022rbsnx,ayalew2026exploring}, which can be attributed to differences in first-principles settings, exchange--correlation functionals, structural phases, and the intrinsic accuracy of the machine-learned band-gap model. Nevertheless, the qualitative agreement indicates that \potente{} can already serve as an effective preliminary screening tool for rapidly identifying plausible lead-free halide perovskite candidates, with higher-fidelity DFT reserved for final validation.

\paragraph{Superionic lithium-ion conductors} 
In the second task, we prompted \potente{} to propose lithium-ion superionic conductors with the 
\(\mathrm{Li}_{12}\mathrm{M}_4\mathrm{X}_{16}\) stoichiometry as shown in Figure~\ref{fig:discovery_prompts}b. This composition can be viewed as a four-formula-unit representation of the \(\mathrm{Li}_3\mathrm{M}\mathrm{X}_4\) family, which is closely related to thio-LISICON-type solid electrolytes such as \(\mathrm{Li}_3\mathrm{PS}_4\). Following the protocol used in the perovskite-discovery task, \potente{} first performed literature and web searches to identify chemically related lithium- and alkali-ion-conducting chalcogenides and plausible substitution trends within this chemical space. Prior work has shown that Li-sublattice disorder in these frameworks can promote fast Li-ion transport~\cite{dathar2017li}. The chemical rationale for the agent-proposed compositions is further supported by reported As- and Sb-based chalcogenide conductors. 

For example, ($\mathrm{Li}_3\mathrm{SbS}_4$) glass and glass-ceramic electrolytes have been synthesized and shown to contain ($\mathrm{SbS}_4$) units, demonstrating that heavier group-15 tetrahedral chalcogenide frameworks can support Li-ion transport~\cite{kimura2019li3sbs4}. In addition, cubic ($\mathrm{Na}_3\mathrm{SbSe}_4$) exhibits fast Na-ion conduction at room temperature, indicating that ($\mathrm{SbSe}_4^{3-}$)-based selenide frameworks are also compatible with superionic alkali-ion transport~\cite{xiong2018na3sbse4}. More broadly, previous computational studies have shown that anion chemistry, framework topology, and cation/anion substitution can substantially affect phase stability and ion migration in lithium superionic conductors~\cite{ong2013lgps_family,wang2015design_principles}.
Therefore, the proposed \(\mathrm{Li}_{12}\mathrm{As}_4\mathrm{Se}_{16}\), 
\(\mathrm{Li}_{12}\mathrm{Sb}_4\mathrm{Se}_{16}\), 
\(\mathrm{Li}_{12}\mathrm{Bi}_4\mathrm{Se}_{16}\), 
\(\mathrm{Li}_{12}\mathrm{As}_4\mathrm{Te}_{16}\), and 
\(\mathrm{Li}_{12}\mathrm{Sb}_4\mathrm{Te}_{16}\) compositions can be interpreted 
as literature-inspired extensions of known 
\(\mathrm{Li}_3\mathrm{M}\mathrm{X}_4\) chalcogenide solid-electrolyte chemistry.

Using the same protocol for structure generation and thermodynamic-stability evaluation as in the first prompt, \potente{} generated ten candidate structures for each of five charge-balanced compositions: \( \mathrm{Li}_{12}\mathrm{Sb}_4\mathrm{Se}_{16} \), \( \mathrm{Li}_{12}\mathrm{As}_4\mathrm{Se}_{16} \), \( \mathrm{Li}_{12}\mathrm{Bi}_4\mathrm{Se}_{16} \), \( \mathrm{Li}_{12}\mathrm{As}_4\mathrm{Te}_{16} \), and \( \mathrm{Li}_{12}\mathrm{Sb}_4\mathrm{Te}_{16} \). Each structure was relaxed with MACE-OMAT, evaluated against the \ac{mlip} convex hull, and the lowest-energy-above-hull structure for each composition was selected for Li-ion transport calculations. Conductivities were obtained from \(2\times2\times2\) replicated supercells using \textit{NPT} Nose-Hoover-chain molecular dynamics at 300~K, with 100~ps equilibration, 100~ps production, a 1~fs timestep, and a logging interval of ten steps.

The selected candidates spanned a narrow metastability window of \(0.0618\)--\(0.1038\)~eV~atom\(^{-1}\) above the \ac{mlip} convex hull, with predicted 300~K Li-ion conductivities from \(5.89\) to \(65.44\)~mS~cm\(^{-1}\). The most stable candidate was \( \mathrm{Li}_{12}\mathrm{Sb}_4\mathrm{Se}_{16} \), at \(0.0618\)~eV~atom\(^{-1}\) above hull, with a predicted conductivity of \(6.56\)~mS~cm\(^{-1}\). \( \mathrm{Li}_{12}\mathrm{As}_4\mathrm{Se}_{16} \) and \( \mathrm{Li}_{12}\mathrm{Bi}_4\mathrm{Se}_{16} \) were similarly close to the hull, with hull distances of \(0.0728\) and \(0.0755\)~eV~atom\(^{-1}\), and conductivities of \(9.14\) and \(5.89\)~mS~cm\(^{-1}\), respectively. The telluride candidates showed higher predicted transport but larger metastability: \( \mathrm{Li}_{12}\mathrm{As}_4\mathrm{Te}_{16} \) reached \(43.16\)~mS~cm\(^{-1}\) at \(0.0993\)~eV~atom\(^{-1}\) above hull, while \( \mathrm{Li}_{12}\mathrm{Sb}_4\mathrm{Te}_{16} \) gave the highest conductivity, \(65.44\)~mS~cm\(^{-1}\), at \(0.1038\)~eV~atom\(^{-1}\) above hull.

These results identify selenide \( \mathrm{Li}_3\mathrm{M}\mathrm{Se}_4 \) chemistries, especially \( \mathrm{Li}_{12}\mathrm{Sb}_4\mathrm{Se}_{16} \) and \( \mathrm{Li}_{12}\mathrm{As}_4\mathrm{Se}_{16} \), as the most balanced candidates in terms of metastability and room-temperature Li-ion mobility. The tellurides \( \mathrm{Li}_{12}\mathrm{As}_4\mathrm{Te}_{16} \) and \( \mathrm{Li}_{12}\mathrm{Sb}_4\mathrm{Te}_{16} \) are higher-risk but potentially high-conductivity candidates: their softer anion frameworks may promote Li mobility, but their larger energy above hull suggests a greater thermodynamic penalty. The large conductivities typically arise from the short MD trajectories, finite supercell sizes, and limitations of both the MLIP and the generative model. These results should therefore be interpreted as screening-level indicators and as a demonstration of the autonomous workflow, rather than as quantitatively validated conductivities.

\subsubsection{Molecular Energy Landscape Exploration}
Molecular energy-landscape exploration presents distinct challenges for autonomous agents. First, practical workflows often span multiple computational domains, including MLIPs and quantum-chemistry methods, and connecting these domains within a unified framework remains challenging. Second, certain types of simulations require chemical intuition rather than direct translation of user instructions, such as identifying relevant conformational degrees of freedom or selecting appropriate collective variables.

\begin{figure}
    \centering
    \includegraphics[width=1.0\linewidth]{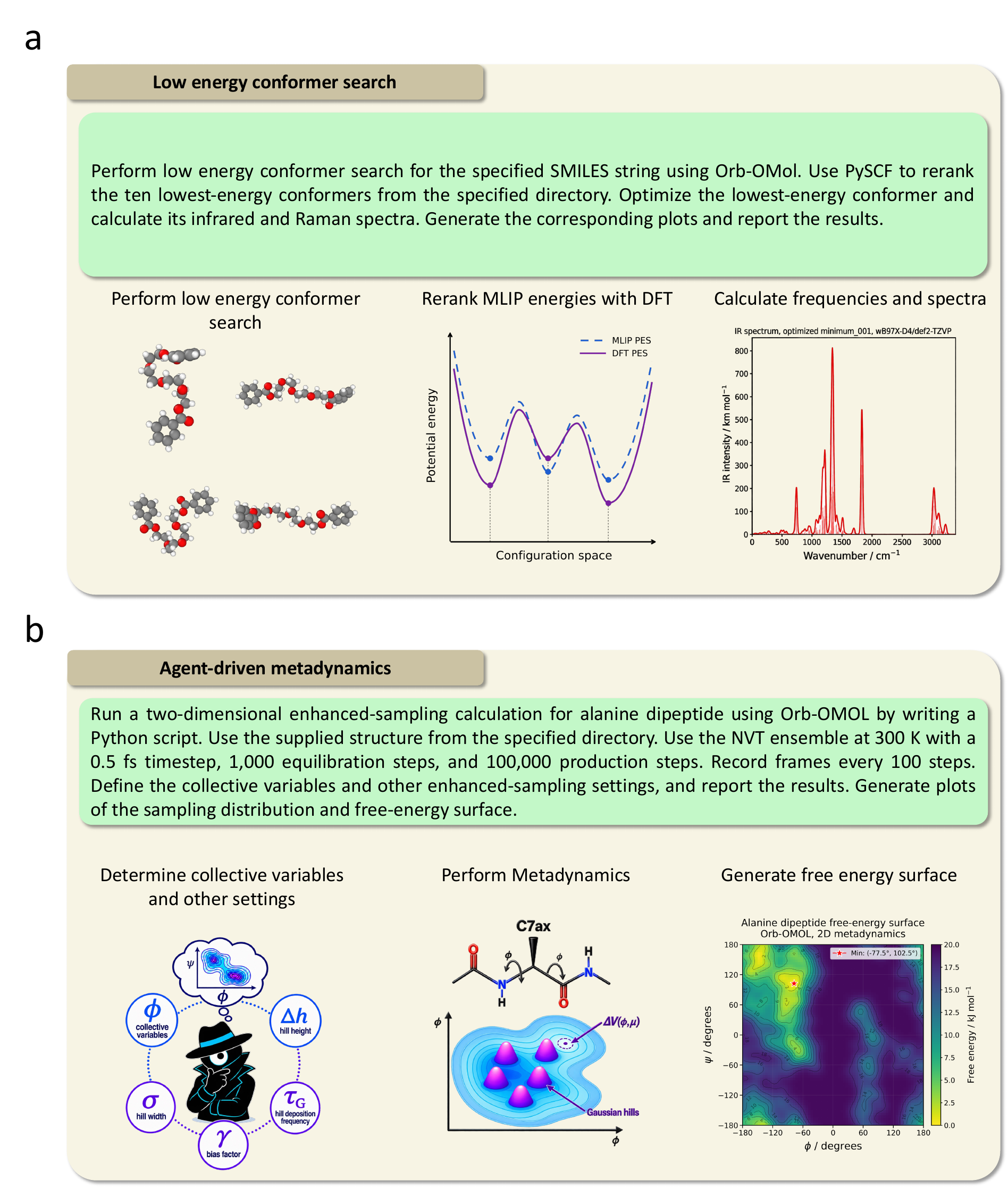}
    \caption{\textbf{Agentic workflows for molecular energy-landscape exploration.}
    \textbf{a,} Low-energy conformer-search workflow. Starting from a user-specified molecular structure, the agent samples conformers with MLIPs, identifies low-energy candidates, reranks MLIP energies with more accurate DFT calculations, and performs vibrational-frequency calculations to generate infrared and Raman spectra. The molecular structures shown are representative conformers obtained from the actual conformer search, and the IR spectrum is the calculated spectrum of the lowest-energy conformer after DFT reranking.
    \textbf{b,} Agent-driven metadynamics workflow. Given a molecular system and a natural-language simulation request, the agent determines appropriate collective variables and enhanced-sampling parameters, executes metadynamics simulations, and analyses the resulting trajectory to construct the multidimensional free-energy surface. The collective-variable-selection and Gaussian-hill panels are schematic representations of the workflow, whereas the displayed two-dimensional free-energy surface is the actual result obtained from the alanine-dipeptide metadynamics simulation. }

    \label{fig:molecular_cases}
\end{figure}
\paragraph{Low-energy conformer search and validation with quantum chemistry}

In the first molecular user case, we evaluated whether \elagenteP{} could coordinate MLIP and quantum-chemistry workflows within \elagenteG{} through a sequence of natural language instructions. Low-energy conformer search is well suited to this demonstration because flexible molecules may contain many local minima separated by small energy differences, making exhaustive exploration directly with DFT computationally demanding. We therefore adopted a hierarchical strategy in which the MLIPs were used for efficient conformational sampling, followed by DFT for reranking and validation. Tetraethylene glycol dibenzoate, C$_{22}$H$_{26}$O$_7$, was selected because its flexible polyether chain contains multiple rotatable C--C and C--O bonds, while its terminal benzoate groups introduce rigid aromatic moieties and intramolecular dispersion interactions. These features give rise to a nontrivial landscape of folded and extended conformations.

We first provided the SMILES string
\texttt{O=C(OCCOCCOCCOCCOC(=O)c1ccccc1)c1ccccc1}
and requested a low-energy conformer search using the conservative Orb-v3~\cite{rhodes2025orb} trained on the OMol25 database~\cite{levine2025open} (Orb-OMOL). The agent interpreted the system as an isolated, neutral, closed-shell molecule and routed the request to a multiple-minimum Monte Carlo conformer-search workflow. The exploration identified 84 distinct local minima distributed over an energy range of approximately $0.535~\mathrm{eV}$ ($12.34~\mathrm{kcal,mol^{-1}}$) relative to the lowest-energy conformer. The ten lowest-energy conformers were concentrated within only $0.129~\mathrm{eV}$ ($2.97~\mathrm{kcal,mol^{-1}}$), indicating a dense manifold of competing low-energy structures. Using computationally efficient MLIPs for broad PES exploration substantially reduces the computational cost compared with performing geometry optimization and energy evaluation for all sampled configurations directly with electronic-structure methods, while enabling much more extensive configurational sampling.

We subsequently prompted the agent to rerank the 10 lowest-energy conformers using PySCF~\cite{sun2020recent}, which offers GPU acceleration for quantum chemistry workflows covering a wide range of electronic structure methods for materials and molecules\cite{li2025introducing}. The lowest-energy conformer was further optimized with the same level of theory, followed by the calculation of infrared (IR) and Raman spectra. The agent launched ten independent single-point calculations in parallel at the $\omega$B97X-D4/def2-TZVP level. After DFT reranking, the conformer ranked lowest in energy by Orb-OMOL remained the lowest-energy structure, while the ten selected structures spanned approximately $0.135\ \mathrm{eV}$ ($3.12\ \mathrm{kcal\,mol^{-1}}$). This DFT-ranked lowest-energy conformer was subsequently optimized to an electronic energy of $-1380.74048\ E_{\mathrm{h}}$, followed by harmonic vibrational analysis of its 159 normal modes. No significant imaginary frequencies above the $50\ \mathrm{cm^{-1}}$ threshold were identified, supporting its assignment as a local minimum.
The calculated IR spectrum exhibited its strongest bands at approximately $1820$ and $1825~\mathrm{cm^{-1}}$, with intensities of $299.7$ and $258.1~\mathrm{km,mol^{-1}}$, respectively, together with a group of intense modes between approximately $1316$ and $1349~\mathrm{cm^{-1}}$. In contrast, the Raman spectrum was dominated by high-frequency modes, with the strongest Raman activities at approximately $3238~\mathrm{cm^{-1}}$ ($306.4$ and $256.3~\text{\AA}^{4},\mathrm{amu}^{-1}$) and $3031~\mathrm{cm^{-1}}$ ($228.9~\text{\AA}^{4},\mathrm{amu}^{-1}$). The corresponding IR and Raman spectra are shown in
Figure~\ref{fig:low_energy_conformer_search_si}. \elagenteP{} automatically generated broadened IR and Raman spectra together with machine-readable peak tables. This example demonstrates how \elagenteP{} can combine efficient MLFF-based exploration of a complex molecular PES with targeted higher-level electronic-structure refinement and spectroscopic characterization, thereby reducing the number of expensive quantum-chemical calculations required for conformational screening.

\paragraph{Agent-driven metadynamics for conformational free-energy exploration}

To test the chemical intuition of state-of-the-art LLM models, we evaluated whether \elagenteP{} could translate a partially specified natural-language request into a complete free-energy sampling workflow through the coding mode. Conventional MD simulations can remain trapped within metastable conformational states when transitions are separated by free-energy barriers that are rarely crossed on accessible simulation timescales. Metadynamics\cite{laio2002escaping} addresses this limitation by introducing a history-dependent bias along selected collective variables (CVs), thereby accelerating transitions between metastable states and enabling reconstruction of the underlying free-energy landscape. 

We asked \potente{} to perform a two-dimensional enhanced-sampling calculation for neutral, singlet alanine dipeptide using Orb-OMOL and supplied an initial C7$_{\mathrm{ax}}$ conformer. The MD conditions were explicitly specified as the NVT ensemble at $300~\mathrm{K}$, a timestep of $0.5~\mathrm{fs}$, 1,000 equilibration steps, 100,000 production steps, and a trajectory-recording interval of 100 steps. However, we intentionally did not specify either the collective variables or the enhanced-sampling parameters, instead asking \elagenteP{} to determine these settings. The agent first parsed the molecular structure and constructed its bonding topology using covalent connectivity. From this topology, it identified the conventional alanine-dipeptide Ramachandran torsions as chemically meaningful coordinates for describing the dominant backbone conformational changes. The $\phi$ torsion was identified as C${\mathrm{acetyl}}$--N--C${\alpha}$--C${\mathrm{carbonyl}}$, corresponding to zero-based atom indices $[4,6,8,14]$, whereas $\psi$ was identified as N--C${\alpha}$--C${\mathrm{carbonyl}}$--N${\mathrm{methylamide}}$, corresponding to $[6,8,14,16]$. For the supplied starting structure, the corresponding torsional angles were $\phi=74.3^{\circ}$ and $\psi=-57.4^{\circ}$, placing the initial configuration in the expected C7$_{\mathrm{ax}}$ region of the alanine-dipeptide Ramachandran surface.

Based on these CVs, \elagenteP{} selected two-dimensional well-tempered metadynamics~\cite{barducci2008well} and generated a Python script that constructed and executed the corresponding typed MLFF workflow request. Both torsional CVs were treated as periodic coordinates over the complete $[-180^{\circ},180^{\circ}]$ domain. The two-dimensional free-energy grid was discretized using $5^{\circ}$ bins along both coordinates, resulting in $72\times72=5184$ grid points. The agent selected Gaussian hills with a height of $1.0~\mathrm{kJ,mol^{-1}}$ and a standard deviation of $10^{\circ}$ along both $\phi$ and $\psi$. Bias was deposited every 100 MD steps, corresponding to an interval of $50~\mathrm{fs}$ at the chosen timestep. A well-tempered bias factor of 10 was used, corresponding to a bias-temperature increment of $2700~\mathrm{K}$ at the $300~\mathrm{K}$ simulation temperature. No additional harmonic confinement was applied because both torsions were explicitly treated as periodic variables over their full angular domains.

The MD was propagated using an NVT Bussi thermostat~\cite{bussi2007canonical} at $300~\mathrm{K}$. After 1,000 steps of equilibration, the system was sampled for 100,000 production steps, corresponding to a total production time of $50~\mathrm{ps}$. Configurations were recorded every 100 steps, yielding 1,001 trajectory frames when the initial production configuration was included. The enhanced-sampling calculation accumulated 100,001 CV samples and visited 2,429 of the 5,184 grid bins, corresponding to approximately $47\%$ of the discretized two-dimensional Ramachandran space. Thus, within a relatively short MLIP trajectory, the metadynamics bias drove the system away from the initial C7$_{\mathrm{ax}}$ configuration and enabled exploration of multiple conformational regions that would be more difficult to sample using conventional MD on the same timescale.

The resulting two-dimensional free-energy surface exhibited its lowest-energy region around $(\phi,\psi)=(-77.5^{\circ},102.5^{\circ})$. Several neighboring grid points, including $(-82.5^{\circ},102.5^{\circ})$, $(-82.5^{\circ},97.5^{\circ})$, and $(-77.5^{\circ},107.5^{\circ})$, occurred within approximately $0.3~\mathrm{kJ,mol^{-1}}$ of the minimum, indicating a broad low-free-energy basin rather than a set of independent minima. A second prominent low-energy region was identified around $(\phi,\psi)=(-147.5^{\circ},157.5^{\circ})$. These regions can be associated with the C7${\mathrm{eq}}$-like and C5-like conformational basins, respectively, while the supplied starting structure at $(74.3^{\circ},-57.4^{\circ})$ corresponds closely to the C7${\mathrm{ax}}$ region. The corresponding free energy surface and sampling distribution are shown in Figure~\ref{fig:metadynamics_si}.

Notably, the locations of the conformational regions sampled by Orb-OMOL are in good qualitative agreement with previous electronic-structure studies of gas-phase alanine dipeptide, which consistently identify C7${\mathrm{eq}}$, C5, and C7${\mathrm{ax}}$ as principal low-energy conformers \cite{Vargas2002Alanine,Fadda2013Alanine}. At the B97-D/def2-TZVP level, Fadda and Woods reported optimized $(\phi,\psi)$ coordinates of $(-82.3^{\circ},75.4^{\circ})$, $(-158.4^{\circ},154.7^{\circ})$, and $(74.1^{\circ},-58.6^{\circ})$ for C7${\mathrm{eq}}$, C5, and C7${\mathrm{ax}}$, respectively \cite{Fadda2013Alanine}. The corresponding MP2/aug-cc-pVDZ structures reported in the same study, $(-82.6^{\circ},75.8^{\circ})$, $(-161.1^{\circ},155.5^{\circ})$, and $(73.7^{\circ},-53.7^{\circ})$, show that the locations of these conformational minima are relatively robust with respect to the electronic-structure method.

The C7${\mathrm{ax}}$ starting configuration at $(74.3^{\circ},-57.4^{\circ})$ is therefore in particularly close agreement with the electronic-structure results, while the sampled C5-like basin around $(-147.5^{\circ},157.5^{\circ})$ also lies close to the corresponding DFT and MP2 minima. The lowest free-energy region sampled by Orb-OMOL around $(-77.5^{\circ},102.5^{\circ})$ can be associated with the C7${\mathrm{eq}}$ region, although it is shifted along $\psi$ relative to the optimized gas-phase DFT minimum. First-principles enhanced-sampling calculations have also been used to construct free-energy and potential-energy surfaces of alanine dipeptide using both local and hybrid DFT functionals \cite{Sevgen2018Hierarchical}, further establishing this system as a benchmark for electronic-structure-based conformational sampling. Overall, the present Orb-OMOL results reproduce the qualitative topology and locations of the principal C7${\mathrm{eq}}$, C5, and C7${\mathrm{ax}}$ conformational regions, with particularly good agreement for C7$_{\mathrm{ax}}$ and C5. We emphasize that this comparison concerns the conformational landscape rather than direct quantitative equivalence of relative energies, because the Orb-OMOL metadynamics calculation yields a finite-temperature free-energy surface at $300~\mathrm{K}$, whereas the optimized DFT and MP2 values correspond to static potential-energy minima.

To facilitate analysis, \elagenteP{} subsequently generated a two-dimensional histogram of the sampled $\phi$--$\psi$ distribution and the corresponding free-energy-surface plot. In addition to the graphical outputs, the workflow preserved the complete biased trajectory, molecular-dynamics log, free-energy grid in CSV format, the generated Python script, the exact enhanced-sampling configuration, and JSON summaries describing the calculation and resulting artifacts.

This example demonstrates that \elagenteP{} can resolve several scientifically important choices that were deliberately omitted from the user's request. Rather than requiring the user to manually identify atom indices or specify detailed biasing parameters, the agent inferred chemically meaningful CVs directly from molecular connectivity, mapped them to the corresponding atoms, recognized their periodic nature, selected an appropriate enhanced-sampling method, and defined the grid resolution, Gaussian hill height and width, deposition frequency, and well-tempered bias factor. The numerical molecular dynamics and bias-force evaluations remained within the Orb-OMOL enhanced-sampling backend, while \elagenteP{} translated the higher-level scientific intent into a complete, executable, and provenance-preserving enhanced-sampling workflow.

\subsubsection{Adsorption and Catalytic Reaction Workflows}

Adsorption and catalytic simulations depend strongly on the chemical environment and physical process under study. Different problems may require distinct strategies for structure construction, configurational sampling, reaction-pathway generation, and the treatment of structural constraints, making it challenging to translate high-level user instructions into robust and physically meaningful simulations.

\begin{figure}
    \centering
    \includegraphics[width=1.0\linewidth]{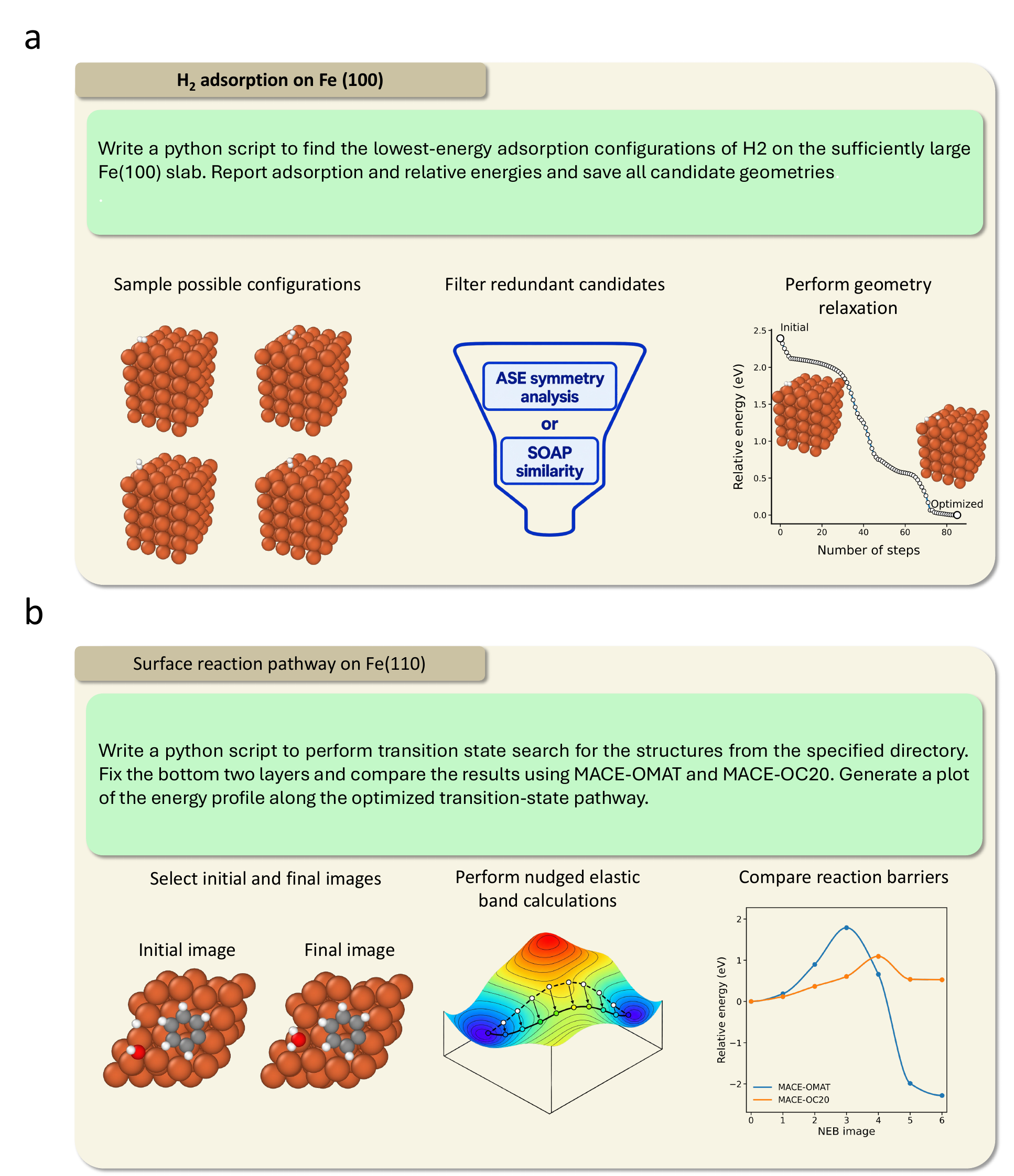}
    \caption{\textbf{Agentic workflows for surface catalysis.}
    \textbf{a,} Identification of low-energy H$_2$ adsorption configurations on Fe(100). The agent generates and executes a workflow that samples candidate adsorption geometries, removes redundant structures using symmetry analysis or SOAP-based structural similarity, and relaxes the remaining candidates to determine their relative adsorption energies and lowest-energy configurations.
    \textbf{b,} Determination of the surface reaction pathway on Fe(110). The agent selects the initial and final images, constrains the bottom two slab layers, performs a nudged elastic band (NEB) calculation to determine the minimum-energy reaction pathway and transition state, and compares the resulting energy profiles and reaction barriers predicted by MACE-OMAT and MACE-OC20.}
    \label{fig:surface_cases}
\end{figure}

\paragraph{Autonomous construction and adsorption-site search of a surface model}
As a surface-science user case, we evaluated whether \elagenteP{} could translate a qualitative request for the ``lowest-energy adsorption configurations'' into a complete adsorption-site search through the coding mode. We requested \potente{} to generate a Python script to identify low-energy \ce{H2} adsorption configurations on Fe(100) using multi-head MACE model with the OC20 head (MACE-OC20), but did not provide an explicit slab structure, adsorption sites, molecular orientations, or structural constraints. 

\elagenteP{} automatically constructed an ideal bcc $\alpha$-Fe(100) slab with a lattice parameter of $2.866~\text{\AA}$, a $4\times4$ lateral supercell, five Fe layers, and $15~\text{\AA}$ of vacuum, resulting in a 160-atom surface model. The cell was kept fixed, the lower three Fe layers were constrained, and the upper two layers together with all adsorbate atoms were allowed to relax. The same relaxation protocol was applied to the clean and adsorbate-covered slabs to ensure a consistent adsorption-energy reference.

The agent represented the adsorbate as neutral, singlet molecular \ce{H2} and automatically constructed the configurational search space. The adsorption-site analysis initially identified 49 candidate surface positions. Because the slab represents an ideal periodic Fe(100) surface, \elagenteP{} applied crystallographic symmetry analysis using ASE and reduced these 49 positions to four symmetry-distinct adsorption environments before performing the more expensive geometry optimizations. The configurational search was not restricted to a single orientation of \ce{H2}. Instead, the workflow considered polar orientations of $0^{\circ}$, $45^{\circ}$, and $90^{\circ}$ relative to the surface normal together with eight azimuthal rotations, thereby sampling upright, tilted, and approximately surface-parallel starting configurations. Both H atoms were considered possible anchoring atoms. Combining adsorption-site and orientational sampling generated 12 candidates for local relaxation, from which 10 distinct relaxed minima were retained.

A key feature of the workflow is that the chemical identity of the initial adsorbate does not constrain the final optimized structure. Although each candidate was initialized using molecular \ce{H2}, the two H atoms remained fully mobile throughout geometry optimization. In the lowest-energy configurations, the H--H bond dissociated spontaneously during relaxation, and the two H atoms formed separate surface-bound configurations. The resulting structures therefore correspond to dissociative adsorption,
[
\ce{H2(g) + 2* -> 2H*},
]
rather than molecular physisorption of intact \ce{H2}. This distinction illustrates an important feature of the adsorption workflow: \elagenteP{} defines and systematically explores chemically reasonable initial configurations, while the underlying MLIP relaxation remains free to identify a different local chemical state when bond breaking or bond formation is energetically favourable.

After ensuring a consistent treatment of the clean and adsorbate-covered surfaces, the lowest-energy relaxed configuration obtained with MACE-OC20 had a dissociative adsorption energy of approximately $-2.47~\mathrm{eV}$ per \ce{H2}. The second-lowest configuration was nearly degenerate, differing by only approximately $2~\mathrm{meV}$, while the next minimum occurred approximately $0.25~\mathrm{eV}$ above the lowest-energy structure. Higher-lying configurations extended to more than $1~\mathrm{eV}$ above the minimum, demonstrating that the systematic site and orientation search sampled both highly favourable and substantially less favourable arrangements of the two surface-bound H atoms. The near degeneracy of the two lowest structures also illustrates why enumeration of multiple adsorption configurations is important: independent initial placements can converge to closely related or symmetry-connected minima that would not necessarily be identified from a single manually selected adsorption geometry.

The absolute MACE-OC20 adsorption energy should, however, be interpreted as a prediction at the level of the underlying MLIP rather than as a quantitatively validated reference value for \ce{H2}/Fe(100). The OC20 electronic-structure data\cite{chanussot2021open} underlying this class of catalyst potentials were generated using RPBE without spin polarization or dispersion corrections. The absence of spin polarization is particularly relevant for Fe-containing catalysts because $\alpha$-Fe is magnetic and adsorption energetics can depend strongly on the magnetic state of the substrate. In addition, molecular $*\mathrm{H2}$ was not included among the original OC20 adsorbates and appears only in subsequent extensions of the Open Catalyst adsorbate space. Consequently, although the spontaneous dissociation and the ranking of alternative surface configurations provide a chemically meaningful demonstration of the workflow, the magnitude of the predicted $-2.47~\mathrm{eV}$ adsorption energy should not by itself be interpreted as quantitative agreement with spin-polarized DFT or experiment. A higher-level validation calculation on the selected minima would be required for that purpose.

The workflow additionally preserved the complete computational provenance rather than returning only the lowest adsorption energy. For each locally optimized candidate, \elagenteP{} retained the initial adsorption configuration, the relaxed structure, adsorption and relative energies, and the full geometry-optimization trajectory. We generated machine-readable CSV and JSON summaries, along with the Fe(100) slab, candidate configurations, and the Python script used to run the calculations. These artifacts make it possible to reconstruct how individual molecular \ce{H2} starting configurations evolved into the final dissociatively adsorbed structures and to determine whether different initial sites or orientations converged to the same local minimum.

This use case therefore demonstrates several levels of scientific workflow automation. Starting from only the identity of the adsorbate and the Fe(100) surface, \elagenteP{} constructed an appropriate periodic slab model, selected its dimensions and vacuum spacing, imposed physically motivated surface constraints, generated the molecular adsorbate, enumerated candidate adsorption sites, removed crystallographically redundant environments, sampled molecular orientations and anchoring choices, and configured and executed the subsequent MACE-OC20 relaxations. Importantly, the workflow did not constrain the final structures to preserve the molecular form assumed in the input and consequently identified dissociatively adsorbed states during relaxation. \elagenteP{} thus translated a high-level request to ``find the lowest-energy adsorption configurations'' into a systematic, executable, and provenance-preserving exploration of the surface potential-energy landscape, while leaving the underlying numerical energy and force evaluations to the selected MLIP backend.

\paragraph{Transition-state search in surface reactions}
Finally, we evaluated whether \elagenteP{} could translate a high-level request for a transition-state search into a complete double-ended reaction-pathway calculation. We provided the reactant and product structures for a surface reaction on Fe(110) and requested a Python script to compare MACE-OMAT and MACE-OC20 while fixing the bottom two surface layers and generating the corresponding energy profiles. \elagenteP{} automatically identified and validated the endpoint structures, verified their atomic correspondence, and determined the bottom two Fe layers from their $z$ coordinates, constraining 32 Fe atoms while keeping the simulation cell fixed. Because the user did not specify a chain-of-states algorithm, the agent constructed a seven-image nudged elastic band (NEB) pathway followed by Sella refinement and independently executed the same workflow using MACE-OMAT and MACE-OC20.

The resulting pathway corresponds to the C$\rightarrow$E elementary step reported in DFT studies~\cite{hensley2015phenol} for phenol deoxygenation on Fe(110). This step is a side reaction of the DHOx(H) pathway in which coadsorbed surface H and OH species combine to form water. The DFT minimum-energy-path calculation reported an activation barrier of $1.41~\mathrm{eV}$ and an endothermic reaction energy of $+0.85~\mathrm{eV}$~\cite{hensley2015phenol}. In comparison, MACE-OC20 predicted an activation barrier of $1.09~\mathrm{eV}$ and a reaction energy of $+0.53~\mathrm{eV}$, underestimating both quantities by approximately $0.32~\mathrm{eV}$ while correctly reproducing the endothermic character of the reaction. MACE-OMAT predicted a higher activation barrier of $1.79~\mathrm{eV}$, differing from the DFT value by approximately $0.38~\mathrm{eV}$, but yielded a strongly exothermic reaction energy of $-2.28~\mathrm{eV}$, opposite to the DFT thermodynamics.

The comparison therefore reveals substantial model dependence even when the same endpoint structures and transition-state-search protocol are used. MACE-OC20 provides reasonable agreement with the DFT reference for this elementary step, reproducing both the approximate magnitude of the activation barrier and the endothermic reaction energy, whereas MACE-OMAT gives a comparable error in the barrier but fails to reproduce the reaction thermodynamics. Beyond the numerical results, this example demonstrates that \elagenteP{} can translate the general instruction ``transition-state search'' into a complete reaction-pathway workflow without requiring the user to manually specify the NEB procedure. The agent automatically constructed the structural constraints and intermediate images, executed multiple MLIP backends, extracted the pathway energetics, and generated comparative energy-profile plots. Comparison with an independently reported DFT minimum-energy pathway further shows how the workflow can assess the reliability and model dependence of foundational MLIPs for reactive surface processes.

\paragraph{Widom insertion for metal-organic frameworks}
\begin{figure}
    \centering
    \includegraphics[width=1.0\linewidth]{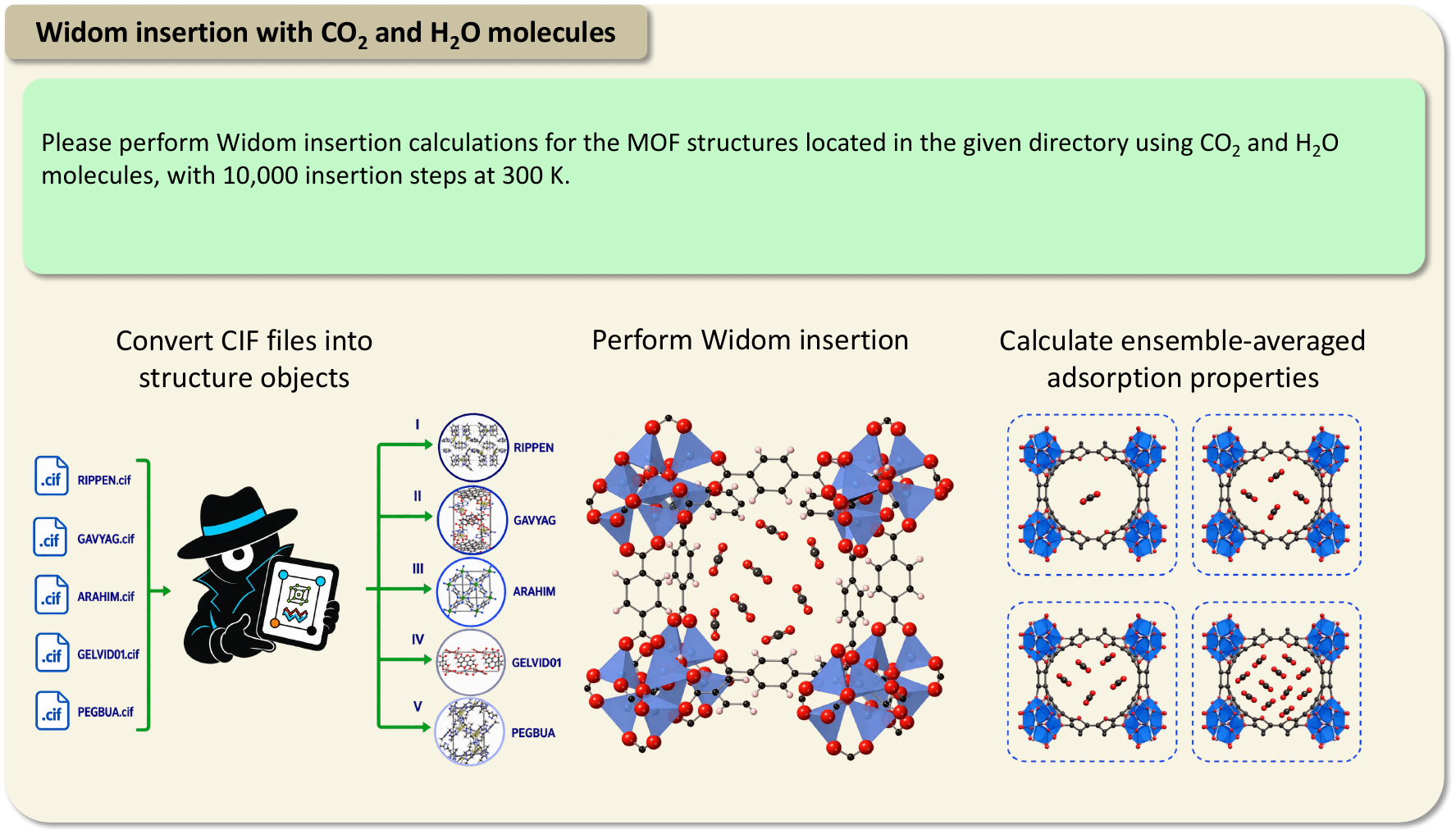}
    \caption{\textbf{Agentic workflow for adsorption-property prediction in metal-organic frameworks.}  CIF files are first converted into structure objects, followed by Widom insertion simulations using CO$_2$ and H$_2$O probe molecules at 300 K. The resulting insertion energies are then ensemble averaged to estimate adsorption properties for each MOF, enabling automated screening of host--guest interactions across multiple framework structures.}
    \label{fig:widom_insertion}
\end{figure}

\Ac{DAC} is a challenging separation problem because CO$_2$ must be captured from ambient air at only about 400~ppm, while H$_2$O competes as an adsorbate. For \acp{mof}, an effective \ac{DAC} material must therefore combine sufficient CO$_2$ affinity with suppressed H$_2$O adsorption, since water can block adsorption sites, reduce working capacity, and increase regeneration costs. Both the zero-loading CO$_2$ heat of adsorption and the Henry-coefficient-based CO$_2$/H$_2$O selectivity are therefore important screening descriptors.

To test \potente{} on a literature-grounded adsorption task, we selected five CoREMOF structures, RIPPEN, GAVYAG, ARAHIM, GELVID01, and PEGBUA, from a previous MLIP-based \ac{DAC} screening study~\cite{lim2025accelerating} as shown in Figure\ref{fig:widom_insertion}. These MOFs contain diverse adsorption environments, including aromatic rings, metal--heteroatom motifs, electronegative functional groups, and methyl groups, making them a compact but chemically varied test set. Starting from the provided CIF structures, \potente{} relaxed each MOF using MACE-DAC-1 (MACE-DAC) from DAC-SIM~\cite{lim2025accelerating} and performed separate Widom insertion calculations for CO$_2$ and H$_2$O at 300~K, using two folds of 10,000 trial insertions per MOF--adsorbate pair (20,000 attempted insertions in total for each pair). All calculations completed successfully and produced zero-loading adsorption descriptors.

For CO$_2$, the computed Henry coefficients, $K_\mathrm{H}^{\mathrm{CO_2}}$, spanned nearly four orders of magnitude, from $4.66\times10^{-3}$ for ARAHIM to $17.82$ for PEGBUA. The corresponding average CO$_2$--framework interaction energies ranged from $-0.438$ to $-0.631$~eV, with adsorption heats of $44.7$--$63.4$~kJ~mol$^{-1}$. PEGBUA showed the largest CO$_2$ Henry coefficient, whereas GELVID01 showed the most exothermic average CO$_2$ interaction energy. This non-monotonic ordering indicates that dilute CO$_2$ uptake depends not only on binding strength, but also on the number and accessibility of favourable insertion sites.

For H$_2$O, $K_\mathrm{H}^{\mathrm{H_2O}}$ varied from $3.23\times10^{-4}$ for RIPPEN to $1.66$ for GELVID01, with adsorption heats of $31.0$--$62.2$~kJ~mol$^{-1}$. GELVID01 was the strongest water-binding framework, while RIPPEN showed the weakest H$_2$O uptake. The CO$_2$/H$_2$O selectivity was then computed as
\[
S_{\mathrm{CO_2}/\mathrm{H_2O}} =
\frac{K_\mathrm{H}^{\mathrm{CO_2}}}{K_\mathrm{H}^{\mathrm{H_2O}}},
\]
and reported as $\log_{10} S_{\mathrm{CO_2}/\mathrm{H_2O}}$ because the selectivity spans several orders of magnitude. As shown in Figure~\ref{fig:widom_insertion_qst_vs_selectivity}, RIPPEN showed the highest selectivity, with $\log_{10}S_{\mathrm{CO_2}/\mathrm{H_2O}}=3.07$, followed by PEGBUA ($2.52$) and GAVYAG ($2.36$). In contrast, GELVID01 was nearly non-selective ($0.04$) because it binds both CO$_2$ and H$_2$O strongly, while ARAHIM was water-selective ($-0.37$).

Overall, these results show that a high CO$_2$ heat of adsorption alone is insufficient for identifying promising \ac{DAC} materials. RIPPEN and PEGBUA best satisfy the combined criterion of appreciable CO$_2$ affinity and suppressed H$_2$O uptake, whereas GELVID01 and ARAHIM are less favourable because of strong H$_2$O binding or weak CO$_2$ uptake. Compared with Ref.~\citenum{lim2025accelerating}, the workflow reproduces the favourable behaviour of RIPPEN and GAVYAG and the strong CO$_2$ affinity of PEGBUA, while showing larger deviations for ARAHIM and GELVID01. These deviations likely arise from differences in geometry-relaxation settings and Widom insertion statistics, including the two-fold sampling protocol in the present workflow (10,000 trial insertions per fold), compared with the single-run calculations reported in the reference. In addition, Widom-derived $K_\mathrm{H}$ values are sensitive to rare low-energy insertion sites and small changes in relaxed framework geometry, particularly in narrow pores or localized adsorption pockets.

\section{Discussion}

In this work, we introduced \elagenteP{} as an agentic framework for \ac{mlip}-driven atomistic simulations, combining structured execution with programmable workflow construction. Typed execution graphs provide explicit control over scientific state, validation, branching, failure handling, provenance, and parallel execution for standardized workflows, whereas coding mode enables task-specific procedural logic when predefined graph components are insufficient. Importantly, the coding agent reuses existing \potente{} functions whenever the requested calculations are already supported, limiting the need to regenerate established numerical procedures. This separation allows \ac{llm}-based reasoning to determine and organize scientific actions while validated computational components remain responsible for their numerical execution.

The benchmarks illustrate the practical consequences of this architecture. Execution-graph workflows reproduced the numerical behaviour of manually scripted calculations while providing structured provenance and failure handling, whereas coding mode supported customized comparative workflows using the same validated computational functions. Comparison with an open-ended skill-based coding agent further highlights a trade-off between flexibility and repeated execution: open-ended coding is useful for building new procedures, while a validated execution graph can reduce repeated runtime reasoning once a workflow is established. These results therefore support using structured execution and programmatic workflow construction as complementary mechanisms rather than competing approaches to scientific automation.

The scientific case studies extend this distinction across workflows with substantially different requirements, from standardized high-throughput screening to problems requiring task-specific decisions such as conformational sampling, collective-variable selection, adsorption-site enumeration, and reaction-pathway construction. However, reliable execution should not be conflated with reliable physical prediction. The accuracy of the resulting calculations remains limited by the transferability of the underlying \acp{mlip}, generative and property models, and by numerical choices such as system size, sampling length, structural relaxation, and convergence criteria. Likewise, structured execution does not guarantee that an \ac{llm} will select the scientifically optimal workflow or simulation parameters. Typed schemas, validated functions, and provenance make such decisions more constrained, reproducible, and inspectable, but higher-fidelity calculations and domain-specific validation remain necessary when quantitative accuracy is required.

An important next step is therefore to make both workflow selection and computational fidelity adaptive. Uncertainty estimates, disagreement among foundational \acp{mlip}, failed validation criteria, or sensitivity to simulation settings could identify cases that require higher-fidelity electronic-structure calculations. Such calculations could then refine screening decisions or provide new data to improve the model. Coupling these capabilities with experimental automation could ultimately support closed-loop workflows in which generative models, foundational \acp{mlip} calculations, and experiments operate at different fidelity levels within a common provenance-aware framework. More broadly, the results suggest a design principle for agentic high-throughput scientific computing: structured execution and procedural flexibility can be implemented as complementary layers, enabling greater scientific autonomy without relinquishing explicit control over numerical computation.

\section{Methods}

\subsection{Computational materials science and chemistry software}
\label{sec}

The project employed \texttt{uv} to manage a Python 3.12 environment. All experiments and benchmarks used ChatGPT~5.5 as the primary \ac{llm}, while ChatGPT~4.1 powered the routing agents. All computational materials science and chemistry packages were installed in a single environment to enable single-agent execution via a shared Python runtime. The computational software stack includes electronic-structure packages, molecular and crystal structure toolkits, \acp{mlip}, phonon and atomistic simulation tools, and MOF-specific analysis packages. Where necessary, forked versions of selected packages were used to resolve dependency conflicts or expose programmatic interfaces required by the workflow.

\begin{enumerate}

\item \textbf{ASE}: The Atomic Simulation Environment, \texttt{ase} (v3.26.0), served as the common atomic-structure representation throughout the workflow~\cite{hjorth2017atomic}. Geometry relaxations and molecular dynamics simulations were also performed through ASE. In addition, ASE provided the calculator interface for MLIPs to evaluate energies, forces, and stresses for input structures, which were extensively used in downstream property calculations.

\item \textbf{Pymatgen}: The Python Materials Genomics, \texttt{pymatgen} (v2025.6.14) was used for querying materials from MP databases through \texttt{mp-api} (v0.45.13) to construct a phase diagram using MLIP energies and to convert structures between ASE and pymatgen objects.

\item \textbf{Machine learning interatomic potentials}: All property calculations were supported by multiple \ac{mlip} backends. \texttt{mace-torch} (v0.3.15)~\cite{batatia2022mace}, with \texttt{e3nn} pinned at v0.5.0 for compatibility with the broader environment. MatterSim-v1.0.0-1M was available through \texttt{mattersim} (v0.1.dev136)~\cite{yang2024mattersim}, and Orb models were available through \texttt{orb-models} (v0.5.5)~\cite{rhodes2025orb}. The PET models for MLIPs~\cite{bigi2026pushing} and band gap prediction~\cite{how2026universal} were available through \texttt{upet} (v2026.1).

\item \textbf{MatCalc}: \texttt{matcalc} (v0.4.5) was included as a high-level materials-property calculation interface for connecting machine-learned interatomic potentials to standard atomistic simulation workflows. In this work, \texttt{matcalc} primarily served as a wrapper for MLIP force evaluators, ASE-compatible calculators, and structure representations from ASE and \texttt{pymatgen}. Downstream property workflows were then executed through specialized materials-science backends: elastic properties were computed from stress--strain calculations using \texttt{pymatgen}-based elasticity analysis, while harmonic and anharmonic lattice-dynamics workflows used \texttt{phonopy} (v3.1.0), \texttt{phono3py} (v3.28.0), and \texttt{symfc} (v1.5.4) for displacement generation, force-constant construction/symmetrization, and phonon-derived thermal properties. This design allowed \potente{} to couple MLIP-based energy, force, and stress predictions with established property-analysis tools in a consistent automated workflow.

\item \textbf{Dispersion corrections}: The dispersion correction for MACE-DAC-1 was available through \texttt{torch-dftd} (v0.5.3).

\item \textbf{Visualization and molecular rendering}: Molecular and crystal-structure visualization was supported by \texttt{py3Dmol} (v2.5.2), \texttt{matplotlib} (v3.10.6), and \texttt{plotly} (v6.3.0). These tools were used for interactive molecular visualization and post-processing of computational results.

\end{enumerate}

\section*{Data availability}
The benchmark results, generated user prompts, and workflow outputs used in this work are available at \url{https://github.com/kenko911/User_Cases_El_Agente_Potente}. Additional data required to reproduce the computational workflows, including input structures, generated scripts, and post-processing outputs, are provided in the same repository where applicable.

\section*{Code availability}
The main computational materials science and chemistry packages used in this work were obtained from the following source-code repositories:  \texttt{ase} (\url{https://gitlab.com/ase/ase}), \texttt{pymatgen} (\url{https://github.com/materialsproject/pymatgen}),   \texttt{mace-torch} (\url{https://github.com/ACEsuit/mace}), \texttt{mattersim} (\url{https://github.com/microsoft/mattersim}), \texttt{orb-models} (\url{https://github.com/orbital-materials/orb-models}), \texttt{matcalc} (\url{https://github.com/materialyzeai/matcalc}), 
\texttt{torch-dftd} (\url{https://github.com/pfnet-research/torch-dftd}),
\texttt{upet}(\url{https://github.com/lab-cosmo/upet}), \texttt{phonopy} (\url{https://github.com/phonopy/phonopy}), \texttt{phono3py} (\url{https://github.com/phonopy/phono3py}). A complete package list and environment specification will be provided with the repository to support reproducibility. Third-party packages, pretrained machine learning interatomic-potential models, and external databases were used according to their respective licenses.



\section*{Acknowledgments}

We gratefully acknowledge the longstanding contributions of the Matter Lab’s current and
past group members (https://www.matter.toronto.edu/), in particular El Agente team. T.W.K. acknowledges the support of the Vector Distinguished Postdoctoral Fellowship.
J.B. acknowledges funding from the Eric and Wendy Schmidt AI in Science Postdoctoral Fellowship Program, a program by Schmidt Futures.
Y.K. acknowledges support from the CIFAR AI Safety Catalyst Award (Catalyst Fund Project \#CF26-AI-001).
A.A.-G. thanks Anders~G.~Fr{\o}seth for his generous support. A.A.-G. also acknowledges the generous support of Natural Resources Canada and the Canada 150 Research Chairs program. This work was supported by the AI2050 program of Schmidt Sciences.
\acknowDARPA
\acknowAC
\acknowGEN{SciNet HPC Consortium (\url{https://scinethpc.ca/})}
\acknowSciNet{Trillium}



{
\small
\bibliography{references}

\begin{thebibliography}{72}
\providecommand{\natexlab}[1]{#1}
\providecommand{\url}[1]{\texttt{#1}}
\expandafter\ifx\csname urlstyle\endcsname\relax
  \providecommand{\doi}[1]{doi: #1}\else
  \providecommand{\doi}{doi: \begingroup \urlstyle{rm}\Url}\fi

\bibitem[Behler and Cs{\'a}nyi(2021)]{behler2021machine}
J{\"o}rg Behler and G{\'a}bor Cs{\'a}nyi.
\newblock Machine learning potentials for extended systems: a perspective.
\newblock \emph{Eur. Phys. J. B}, 94\penalty0 (7):\penalty0 142, 2021.
\newblock \doi{10.1140/epjb/s10051-021-00156-1}.

\bibitem[Friederich et~al.(2021)Friederich, H{\"a}se, Proppe, and
  Aspuru-Guzik]{friederich2021machine}
Pascal Friederich, Florian H{\"a}se, Jonny Proppe, and Al{\'a}n Aspuru-Guzik.
\newblock Machine-learned potentials for next-generation matter simulations.
\newblock \emph{Nat. Mater.}, 20\penalty0 (6):\penalty0 750--761, 2021.
\newblock \doi{10.1038/s41563-020-0777-6}.

\bibitem[Ko and Ong(2023)]{ko2023recent}
Tsz~Wai Ko and Shyue~Ping Ong.
\newblock Recent advances and outstanding challenges for machine learning
  interatomic potentials.
\newblock \emph{Nat. Comput. Sci.}, 3\penalty0 (12):\penalty0 998--1000, 2023.
\newblock \doi{10.1038/s43588-023-00561-9}.

\bibitem[Zhang et~al.(2025)Zhang, Sorkin, Aitken, Politano, Behler, Thompson,
  Ko, Ong, Chalykh, Korogod, et~al.]{zhang2025roadmap}
Yong-Wei Zhang, Viacheslav Sorkin, Zachary~H Aitken, Antonio Politano, J{\"o}rg
  Behler, Aidan~P Thompson, Tsz~Wai Ko, Shyue~Ping Ong, Olga Chalykh, Dmitry
  Korogod, et~al.
\newblock Roadmap for the development of machine learning-based interatomic
  potentials.
\newblock \emph{Model. Simul. Mater. Sci. Eng.}, 33\penalty0 (2):\penalty0
  023301, 2025.
\newblock \doi{10.1088/1361-651X/ad9d63}.

\bibitem[Chen and Ong(2022)]{chen2022universal}
Chi Chen and Shyue~Ping Ong.
\newblock A universal graph deep learning interatomic potential for the
  periodic table.
\newblock \emph{Nat. Comput. Sci.}, 2\penalty0 (11):\penalty0 718--728, 2022.
\newblock \doi{10.1038/s43588-022-00349-3}.

\bibitem[Deng et~al.(2023)Deng, Zhong, Jun, Riebesell, Han, Bartel, and
  Ceder]{deng2023chgnet}
Bowen Deng, Peichen Zhong, KyuJung Jun, Janosh Riebesell, Kevin Han,
  Christopher~J Bartel, and Gerbrand Ceder.
\newblock Chgnet as a pretrained universal neural network potential for
  charge-informed atomistic modelling.
\newblock \emph{Nat. Mach. Intell.}, 5\penalty0 (9):\penalty0 1031--1041, 2023.
\newblock \doi{10.1038/s42256-023-00716-3}.

\bibitem[Batatia et~al.(2025)Batatia, Benner, Chiang, Elena, Kov{\'a}cs,
  Riebesell, Advincula, Asta, Avaylon, Baldwin, et~al.]{batatia2025foundation}
Ilyes Batatia, Philipp Benner, Yuan Chiang, Alin~M Elena, D{\'a}vid~P
  Kov{\'a}cs, Janosh Riebesell, Xavier~R Advincula, Mark Asta, Matthew Avaylon,
  William~J Baldwin, et~al.
\newblock A foundation model for atomistic materials chemistry.
\newblock \emph{The Journal of chemical physics}, 163\penalty0 (18):\penalty0
  184110, 2025.
\newblock \doi{10.1063/5.0297006}.

\bibitem[Ko et~al.(2025)Ko, Liu, Mishra, Yu, Qi, and Ong]{ko2025fast}
Tsz~Wai Ko, Runze Liu, Adesh~Rohan Mishra, Zihan Yu, Ji~Qi, and Shyue~Ping Ong.
\newblock A fast, accurate, and reactive equivariant foundation potential.
\newblock \emph{arXiv preprint arXiv:2511.07249}, 2025.
\newblock \doi{10.48550/arXiv.2511.07249}.

\bibitem[Curtarolo et~al.(2012)Curtarolo, Setyawan, Hart, Jahnatek, Chepulskii,
  Taylor, Wang, Xue, Yang, Levy, et~al.]{curtarolo2012aflow}
Stefano Curtarolo, Wahyu Setyawan, Gus~LW Hart, Michal Jahnatek, Roman~V
  Chepulskii, Richard~H Taylor, Shidong Wang, Junkai Xue, Kesong Yang, Ohad
  Levy, et~al.
\newblock Aflow: An automatic framework for high-throughput materials
  discovery.
\newblock \emph{Computational Materials Science}, 58:\penalty0 218--226, 2012.
\newblock \doi{10.1016/j.commatsci.2012.02.005}.

\bibitem[Huber et~al.(2020)Huber, Zoupanos, Uhrin, Talirz, Kahle,
  H{\"a}uselmann, Gresch, M{\"u}ller, Yakutovich, Andersen,
  et~al.]{huber2020aiida}
Sebastiaan~P Huber, Spyros Zoupanos, Martin Uhrin, Leopold Talirz, Leonid
  Kahle, Rico H{\"a}uselmann, Dominik Gresch, Tiziano M{\"u}ller, Aliaksandr~V
  Yakutovich, Casper~W Andersen, et~al.
\newblock Aiida 1.0, a scalable computational infrastructure for automated
  reproducible workflows and data provenance.
\newblock \emph{Scientific data}, 7\penalty0 (1):\penalty0 300, 2020.
\newblock \doi{10.1038/s41597-020-00638-4}.

\bibitem[Uhrin et~al.(2021)Uhrin, Huber, Yu, Marzari, and
  Pizzi]{uhrin2021workflows}
Martin Uhrin, Sebastiaan~P Huber, Jusong Yu, Nicola Marzari, and Giovanni
  Pizzi.
\newblock Workflows in aiida: Engineering a high-throughput, event-based engine
  for robust and modular computational workflows.
\newblock \emph{Computational Materials Science}, 187:\penalty0 110086, 2021.
\newblock \doi{10.1016/j.commatsci.2020.110086}.

\bibitem[Janssen et~al.(2019)Janssen, Surendralal, Lysogorskiy, Todorova,
  Hickel, Drautz, and Neugebauer]{janssen2019pyiron}
Jan Janssen, Sudarsan Surendralal, Yury Lysogorskiy, Mira Todorova, Tilmann
  Hickel, Ralf Drautz, and J{\"o}rg Neugebauer.
\newblock pyiron: An integrated development environment for computational
  materials science.
\newblock \emph{Computational Materials Science}, 163:\penalty0 24--36, 2019.
\newblock \doi{10.1016/j.commatsci.2018.07.043}.

\bibitem[Menon et~al.(2024)Menon, Lysogorskiy, Knoll, Leimeroth, Poul, Qamar,
  Janssen, Mrovec, Rohrer, Albe, et~al.]{menon2024electrons}
Sarath Menon, Yury Lysogorskiy, Alexander~LM Knoll, Niklas Leimeroth, Marvin
  Poul, Minaam Qamar, Jan Janssen, Matous Mrovec, Jochen Rohrer, Karsten Albe,
  et~al.
\newblock From electrons to phase diagrams with machine learning potentials
  using pyiron based automated workflows.
\newblock \emph{npj Computational Materials}, 10\penalty0 (1):\penalty0 261,
  2024.
\newblock \doi{10.1038/s41524-024-01441-0}.

\bibitem[Ganose et~al.(2025)Ganose, Sahasrabuddhe, Asta, Beck, Biswas,
  Bonkowski, Bustamante, Chen, Chiang, Chrzan, et~al.]{ganose2025atomate2}
Alex~M Ganose, Hrushikesh Sahasrabuddhe, Mark Asta, Kevin Beck, Tathagata
  Biswas, Alexander Bonkowski, Joana Bustamante, Xin Chen, Yuan Chiang, Daryl~C
  Chrzan, et~al.
\newblock Atomate2: modular workflows for materials science.
\newblock \emph{Digital Discovery}, 4\penalty0 (7):\penalty0 1944--1973, 2025.
\newblock \doi{10.1039/D5DD00019J}.

\bibitem[Pham et~al.(2026)Pham, Tanikanti, and Ke\c{c}eli]{Pham2026ChemGraph}
Thang~D. Pham, Aditya Tanikanti, and Murat Ke\c{c}eli.
\newblock Chemgraph as an agentic framework for computational chemistry
  workflows.
\newblock \emph{Communications Chemistry}, 9\penalty0 (1), January 2026.
\newblock ISSN 2399-3669.
\newblock \doi{10.1038/s42004-025-01776-9}.
\newblock \url{http://dx.doi.org/10.1038/s42004-025-01776-9}.

\bibitem[Wang et~al.(2025)Wang, Huang, Zhao, Xu, Zhu, Janssen, and
  Viswanathan]{wang2025dreams}
Ziqi Wang, Hongshuo Huang, Hancheng Zhao, Changwen Xu, Shang Zhu, Jan Janssen,
  and Venkatasubramanian Viswanathan.
\newblock Dreams: Density functional theory based research engine for agentic
  materials simulation.
\newblock \emph{arXiv preprint arXiv:2507.14267}, 2025.
\newblock \doi{10.48550/arXiv.2507.14267}.

\bibitem[Hu et~al.(2025)Hu, Nawaz, Hou, Rui, Chi, Chen, Ullah, and
  Dral]{hu2025aitomia}
Jinming Hu, Hassan Nawaz, Yi-Fan Hou, Yuting Rui, Lijie Chi, Yuxinxin Chen,
  Arif Ullah, and Pavlo~O Dral.
\newblock Aitomia: Your intelligent assistant for ai-driven atomistic and
  quantum chemical simulations.
\newblock \emph{arXiv preprint arXiv:2505.08195}, 2025.
\newblock \doi{10.48550/arXiv.2505.08195}.

\bibitem[Nduma et~al.(2025)Nduma, Park, and Walsh]{Nduma2025Crystalyse}
Ryan Nduma, Hyunsoo Park, and Aron Walsh.
\newblock Crystalyse: a multi-tool agent for materials design, 2025.
\newblock \url{https://arxiv.org/abs/2512.00977}.

\bibitem[Liu et~al.(2026)Liu, Ma, Zhang, Pascasio, Yang, Chen, and
  Li]{Liu2026CatGo}
Guangsheng Liu, Xiao Ma, Leshen Zhang, Jenedith Pascasio, Jonathan Yang,
  Yuxiang Chen, and Wan-Lu Li.
\newblock Catgo: Bridging cli coding agents with interactive structure and
  workflow management for computational chemistry.
\newblock \emph{Digital Discovery}, 2026.
\newblock \doi{10.1039/D6DD00273K}.

\bibitem[Kim et~al.(2026)Kim, Choi, Jang, Park, Bernales, Aspuru-Guzik, and
  Jung]{kim2026materealize}
Seongmin Kim, Jaehwan Choi, Kunik Jang, Junkil Park, Varinia Bernales, Al{\'a}n
  Aspuru-Guzik, and Yousung Jung.
\newblock Materealize: a multi-agent deliberation system for end-to-end
  material design and synthesis.
\newblock \emph{arXiv preprint arXiv:2601.15743}, 2026.
\newblock \doi{10.48550/arXiv.2601.15743}.

\bibitem[Vriza et~al.(2026)Vriza, Kornu, Koneru, Chan, and
  Sankaranarayanan]{vriza2026multi}
Aikaterini Vriza, Uma Kornu, Aditya Koneru, Henry Chan, and Subramanian~KRS
  Sankaranarayanan.
\newblock Multi-agentic ai framework for end-to-end atomistic simulations.
\newblock \emph{Digital Discovery}, 5\penalty0 (1):\penalty0 440--452, 2026.
\newblock \doi{10.1039/D5DD00435G}.

\bibitem[Liu et~al.(2026)Liu, Yang, and Zhong]{liu2026masgent}
Guangchen Liu, Songge Yang, and Yu~Zhong.
\newblock Masgent: an ai-assisted materials simulation agent.
\newblock \emph{Digital Discovery}, 5\penalty0 (5):\penalty0 2151--2171, 2026.
\newblock \doi{10.1039/D6DD00043F}.

\bibitem[Deng et~al.(2026)Deng, Li, Cox, Chun, Nam, Lyssenko, Edamadaka, Ruza,
  Du, Segal, et~al.]{deng2026harnessing}
Bowen Deng, Bohan Li, Matthew Cox, Hoje Chun, Juno Nam, Artur Lyssenko, Sathya
  Edamadaka, Jurgis Ruza, Xiaochen Du, Nofit Segal, et~al.
\newblock Harnessing atomisticskills for agentic atomistic research.
\newblock \emph{arXiv preprint arXiv:2605.24002}, 2026.
\newblock \doi{10.48550/arXiv.2605.24002}.

\bibitem[Zou et~al.(2025)Zou, Cheng, Aldossary, Bai, Leong,
  Campos-Gonzalez-Angulo, Choi, Ser, Tom, Wang, et~al.]{zou2025agente}
Yunheng Zou, Austin~H Cheng, Abdulrahman Aldossary, Jiaru Bai, Shi~Xuan Leong,
  Jorge~Arturo Campos-Gonzalez-Angulo, Changhyeok Choi, Cher~Tian Ser, Gary
  Tom, Andrew Wang, et~al.
\newblock El agente: An autonomous agent for quantum chemistry.
\newblock \emph{Matter}, 8\penalty0 (7):\penalty0 102263, 2025.
\newblock \doi{10.1016/j.matt.2025.102263}.

\bibitem[P{\'e}rez-S{\'a}nchez et~al.(2026)P{\'e}rez-S{\'a}nchez, Zou,
  Campos-Gonzalez-Angulo, M{\"u}ller, Gustin, Wang, Hao, Ko, Choi, Isbrandt,
  et~al.]{perez2026agente}
Juan~B P{\'e}rez-S{\'a}nchez, Yunheng Zou, Jorge~A Campos-Gonzalez-Angulo,
  Marcel M{\"u}ller, Ignacio Gustin, Andrew Wang, Han Hao, Tsz~Wai Ko,
  Changhyeok Choi, Eric~S Isbrandt, et~al.
\newblock El agente quntur: A research collaborator agent for quantum
  chemistry.
\newblock \emph{arXiv preprint arXiv:2602.04850}, 2026.
\newblock \doi{10.48550/arXiv.2602.04850}.

\bibitem[Kumar et~al.(2026)Kumar, Zou, Wang, Vald{\'e}s-Hern{\'a}ndez, Ko, Yue,
  Leng, Xu, Crebolder, Aspuru-Guzik, et~al.]{kumar2026agente}
Sai Govind~Hari Kumar, Yunheng Zou, Andrew Wang, Jes{\'u}s
  Vald{\'e}s-Hern{\'a}ndez, Tsz~Wai Ko, Nathan Yue, Olivia Leng, Hanyong Xu,
  Chris Crebolder, Al{\'a}n Aspuru-Guzik, et~al.
\newblock El agente s{\'o}lido: A new age(nt) for solid state simulations.
\newblock \emph{arXiv preprint arXiv:2602.17886}, 2026.
\newblock \doi{10.48550/arXiv.2602.17886}.

\bibitem[Choi et~al.(2026)Choi, Zou, M{\"u}ller, Hao, Kang,
  P{\'e}rez-S{\'a}nchez, Gustin, Xu, Wang, Vakili, et~al.]{choi2026agente}
Changhyeok Choi, Yunheng Zou, Marcel M{\"u}ller, Han Hao, Yeonghun Kang, Juan~B
  P{\'e}rez-S{\'a}nchez, Ignacio Gustin, Hanyong Xu, Andrew Wang,
  Mohammad~Ghazi Vakili, et~al.
\newblock El agente estructural: An artificially intelligent molecular editor.
\newblock \emph{arXiv preprint arXiv:2602.04849}, 2026.
\newblock \doi{10.48550/arXiv.2602.04849}.

\bibitem[Gustin et~al.(2026)Gustin, Mantilla~Calder{\'o}n,
  P{\'e}rez-S{\'a}nchez, Crebolder, Gonthier, Vakili, Nakamura, Panicker,
  Ramprasad, Yin, et~al.]{gustin2026agente}
Ignacio Gustin, Luis Mantilla~Calder{\'o}n, Juan~B P{\'e}rez-S{\'a}nchez, Chris
  Crebolder, J{\'e}r{\^o}me~F Gonthier, Mohammad~Ghazi Vakili, Yuma Nakamura,
  Karthik Panicker, Manav Ramprasad, Aiwei Yin, et~al.
\newblock El agente cu{\'a}ntico: automating quantum simulations.
\newblock \emph{Reports on Progress in Physics}, 89\penalty0 (7):\penalty0
  077602, 2026.
\newblock \doi{10.1088/1361-6633/ae8933}.

\bibitem[Bai et~al.(2026)Bai, Aldossary, Swanick, M\"{u}ller, Kang, Zhang, Lee,
  Ko, Vakili, Bernales, and Aspuru-Guzik]{grafico}
Jiaru Bai, Abdulrahman Aldossary, Thomas Swanick, Marcel M\"{u}ller, Yeonghun
  Kang, Zijian Zhang, Jin~Won Lee, Tsz~Wai Ko, Mohammad~Ghazi Vakili, Varinia
  Bernales, and Al{\'a}n Aspuru-Guzik.
\newblock {El Agente Gr{\'a}fico}: Structured execution graphs for scientific
  agents.
\newblock \emph{arXiv preprint arXiv:2602.17902}, 2026.
\newblock \doi{10.48550/arXiv.2602.17902}.
\newblock \url{https://arxiv.org/abs/2602.17902}.

\bibitem[Zhang et~al.(2026)Zhang, Yin, Baweja, Bai, Gustin, Bernales, and
  Aspuru-Guzik]{zhang2026agente}
Zijian Zhang, Aiwei Yin, Amaan Baweja, Jiaru Bai, Ignacio Gustin, Varinia
  Bernales, and Al{\'a}n Aspuru-Guzik.
\newblock El agente forjador: Task-driven agent generation for quantum
  simulation.
\newblock \emph{arXiv preprint arXiv:2604.14609}, 2026.
\newblock \doi{10.48550/arXiv.2604.14609}.

\bibitem[Kang et~al.(2026)Kang, Leong, Park, Bai, Papidocha, Zhao, Zhang,
  Gimeno, Lo, Isbrandt, Hao, Cao, Katzenburg, Du, Aspuru-Guzik, and
  Bernales]{kang2026seguro}
Yeonghun Kang, Shi~Xuan Leong, Hyun~Suk Park, Jiaru Bai, Sven Papidocha, Yuchi
  Zhao, Rui Zhang, Magali Gimeno, Stanley Lo, Eric~S. Isbrandt, Han Hao, Yang
  Cao, Felix Katzenburg, Fangshi Du, Al{\'a}n Aspuru-Guzik, and Varinia
  Bernales.
\newblock El agente seguro: An agent for chemical safety.
\newblock \emph{ChemRxiv}, 2026\penalty0 (0610), 2026.
\newblock \doi{10.26434/chemrxiv.15004378/v2}.
\newblock \url{https://chemrxiv.org/doi/abs/10.26434/chemrxiv.15004378/v2}.

\bibitem[Dunn et~al.(2020)Dunn, Wang, Ganose, Dopp, and
  Jain]{dunn2020benchmarking}
Alexander Dunn, Qi~Wang, Alex Ganose, Daniel Dopp, and Anubhav Jain.
\newblock Benchmarking materials property prediction methods: the matbench test
  set and automatminer reference algorithm.
\newblock \emph{npj Computational Materials}, 6\penalty0 (1):\penalty0 138,
  2020.
\newblock \doi{10.1038/s41524-020-00406-3}.

\bibitem[Batatia et~al.(2025)Batatia, Lin, Hart, Kasoar, Elena, Norwood, Wolf,
  and Cs{\'a}nyi]{batatia2025cross}
Ilyes Batatia, Chen Lin, Joseph Hart, Elliott Kasoar, Alin~M Elena, Sam~Walton
  Norwood, Thomas Wolf, and G{\'a}bor Cs{\'a}nyi.
\newblock Cross learning between electronic structure theories for unifying
  molecular, surface, and inorganic crystal foundation force fields.
\newblock \emph{arXiv preprint arXiv:2510.25380}, 2025.
\newblock \doi{10.48550/arXiv.2510.25380}.

\bibitem[Grosse-Kunstleve et~al.(2004)Grosse-Kunstleve, Sauter, and
  Adams]{grossekunstleve2004numerically}
R.~W. Grosse-Kunstleve, N.~K. Sauter, and P.~D. Adams.
\newblock Numerically stable algorithms for the computation of reduced unit
  cells.
\newblock \emph{Acta Crystallographica Section A}, 60:\penalty0 1--6, 2004.
\newblock \doi{10.1107/S010876730302186X}.

\bibitem[Yang et~al.(2024)Yang, Hu, Zhou, Liu, Shi, Li, Li, Chen, Chen, Zeni,
  et~al.]{yang2024mattersim}
Han Yang, Chenxi Hu, Yichi Zhou, Xixian Liu, Yu~Shi, Jielan Li, Guanzhi Li,
  Zekun Chen, Shuizhou Chen, Claudio Zeni, et~al.
\newblock Mattersim: A deep learning atomistic model across elements,
  temperatures and pressures.
\newblock \emph{arXiv preprint arXiv:2405.04967}, 2024.
\newblock \doi{10.48550/arXiv.2405.04967}.

\bibitem[Rhodes et~al.(2025)Rhodes, Vandenhaute, {\v{S}}imkus, Gin, Godwin,
  Duignan, and Neumann]{rhodes2025orb}
Benjamin Rhodes, Sander Vandenhaute, Vaidotas {\v{S}}imkus, James Gin, Jonathan
  Godwin, Tim Duignan, and Mark Neumann.
\newblock Orb-v3: atomistic simulation at scale.
\newblock \emph{arXiv preprint arXiv:2504.06231}, 2025.
\newblock \doi{10.48550/arXiv.2504.06231}.

\bibitem[Vrtovec et~al.(2020)Vrtovec, Mazaj, Buscarino, Terracina, Agnello,
  Arcon, Kovac, and Zabukovec~Logar]{vrtovec2020structural}
Nika Vrtovec, Matjaz Mazaj, Gianpiero Buscarino, Angela Terracina, Simonpietro
  Agnello, Iztok Arcon, Janez Kovac, and Natasa Zabukovec~Logar.
\newblock Structural and co2 capture properties of ethylenediamine-modified
  hkust-1 metal--organic framework.
\newblock \emph{Crystal Growth \& Design}, 20\penalty0 (8):\penalty0
  5455--5465, 2020.
\newblock \doi{10.1021/acs.cgd.0c00667}.

\bibitem[Chen et~al.(2018)Chen, Mu, Lester, and Wu]{chen2018high}
Yipei Chen, Xueliang Mu, Edward Lester, and Tao Wu.
\newblock High efficiency synthesis of hkust-1 under mild conditions with high
  bet surface area and co2 uptake capacity.
\newblock \emph{Progress in Natural Science: Materials International},
  28\penalty0 (5):\penalty0 584--589, 2018.
\newblock \doi{10.1016/j.pnsc.2018.08.002}.

\bibitem[Cort{\'e}s-S{\'u}arez et~al.(2019)Cort{\'e}s-S{\'u}arez, Celis-Arias,
  Beltr{\'a}n, et~al.]{cortessuarez2019swcnt}
Jonathan Cort{\'e}s-S{\'u}arez, Vanessa Celis-Arias, Hiram~I. Beltr{\'a}n,
  et~al.
\newblock Synthesis and characterization of an swcnt@hkust-1 composite:
  Enhancing the co2 adsorption properties of hkust-1.
\newblock \emph{ACS Omega}, 4\penalty0 (3):\penalty0 5275--5282, 2019.
\newblock \doi{10.1021/acsomega.9b00330}.

\bibitem[Jain et~al.(2013)Jain, Ong, Hautier, Chen, Richards, Dacek, Cholia,
  Gunter, Skinner, Ceder, et~al.]{jain2013commentary}
Anubhav Jain, Shyue~Ping Ong, Geoffroy Hautier, Wei Chen, William~Davidson
  Richards, Stephen Dacek, Shreyas Cholia, Dan Gunter, David Skinner, Gerbrand
  Ceder, et~al.
\newblock Commentary: The materials project: A materials genome approach to
  accelerating materials innovation.
\newblock \emph{APL Mater.}, 1\penalty0 (1), 2013.
\newblock \doi{10.1063/1.4812323}.

\bibitem[Horton et~al.(2025)Horton, Huck, Yang, Munro, Dwaraknath, Ganose,
  Kingsbury, Wen, Shen, Mathis, Kaplan, Berket, Riebesell, George, Rosen,
  Spotte-Smith, McDermott, Cohen, Dunn, Kuner, Rignanese, Petretto, Waroquiers,
  Griffin, Neaton, Chrzan, Asta, Hautier, Cholia, Ceder, Ong, Jain, and
  Persson]{Horton2025accelerated}
Matthew~K. Horton, Patrick Huck, Ruo~Xi Yang, Jason~M. Munro, Shyam Dwaraknath,
  Alex~M. Ganose, Ryan~S. Kingsbury, Mingjian Wen, Jimmy~X. Shen, Tyler~S.
  Mathis, Aaron~D. Kaplan, Karlo Berket, Janosh Riebesell, Janine George,
  Andrew~S. Rosen, Evan W.~C. Spotte-Smith, Matthew~J. McDermott, Orion~A.
  Cohen, Alex Dunn, Matthew~C. Kuner, Gian-Marco Rignanese, Guido Petretto,
  David Waroquiers, Sinead~M. Griffin, Jeffrey~B. Neaton, Daryl~C. Chrzan, Mark
  Asta, Geoffroy Hautier, Shreyas Cholia, Gerbrand Ceder, Shyue~Ping Ong,
  Anubhav Jain, and Kristin~A. Persson.
\newblock Accelerated data-driven materials science with the materials project.
\newblock \emph{Nature Materials}, 24\penalty0 (10):\penalty0 1522–1532,
  2025.
\newblock ISSN 1476-4660.
\newblock \doi{10.1038/s41563-025-02272-0}.
\newblock \url{http://dx.doi.org/10.1038/s41563-025-02272-0}.

\bibitem[Mitro et~al.(2022)Mitro, Saiduzzaman, Biswas, Sultana, and
  Hossain]{mitro2022electronic}
SK~Mitro, Md~Saiduzzaman, Arpon Biswas, Aldina Sultana, and Khandaker~Monower
  Hossain.
\newblock Electronic phase transition and enhanced optoelectronic performance
  of lead-free halide perovskites agei3 (a= rb, k) under pressure.
\newblock \emph{Materials Today Communications}, 31:\penalty0 103532, 2022.
\newblock \doi{10.1016/j.mtcomm.2022.103532}.

\bibitem[Rahman et~al.(2022)Rahman, Jubair, Rahaman, Ahasan, Ostrikov, and
  Roknuzzaman]{rahman2022rbsnx}
Md~Habibur Rahman, Md~Jubair, Md~Zahidur Rahaman, Md~Shamim Ahasan, Kostya~Ken
  Ostrikov, and Md~Roknuzzaman.
\newblock Rbsnx 3 (x= cl, br, i): promising lead-free metal halide perovskites
  for photovoltaics and optoelectronics.
\newblock \emph{RSC advances}, 12\penalty0 (12):\penalty0 7497--7505, 2022.
\newblock \doi{10.1039/D2RA00414C}.

\bibitem[Rahman and Shorowordi(2025)]{rahman2025first}
Abu Sadat Md~Sayem Rahman and Kazi~Md Shorowordi.
\newblock First-principles study of novel non-toxic trigonal kgex3 (x= br, i)
  perovskites: A potential for optoelectronic applications.
\newblock \emph{Materials Science in Semiconductor Processing}, 186:\penalty0
  109114, 2025.
\newblock \doi{10.1016/j.mssp.2024.109114}.

\bibitem[Ayalew and Kuma(2026)]{ayalew2026exploring}
Beyene~Tesfaw Ayalew and Shiferaw~Gadisa Kuma.
\newblock Exploring cssncl3 as a lead-free halide perovskite: Insights from
  density functional theory.
\newblock \emph{Advances in Condensed Matter Physics}, 2026\penalty0
  (1):\penalty0 2502469, 2026.
\newblock \doi{10.1155/acmp/2502469}.

\bibitem[Zeni et~al.(2025)Zeni, Pinsler, Z{\"u}gner, Fowler, Horton, Fu, Wang,
  Shysheya, Crabb{\'e}, Ueda, et~al.]{zeni2025generative}
Claudio Zeni, Robert Pinsler, Daniel Z{\"u}gner, Andrew Fowler, Matthew Horton,
  Xiang Fu, Zilong Wang, Aliaksandra Shysheya, Jonathan Crabb{\'e}, Shoko Ueda,
  et~al.
\newblock A generative model for inorganic materials design.
\newblock \emph{Nature}, 639\penalty0 (8055):\penalty0 624--632, 2025.
\newblock \doi{10.1038/s41586-025-08628-5}.

\bibitem[Barros-Luque et~al.(2026)Barros-Luque, Shuaibi, Fu, Wood, Dzamba, Gao,
  Rizvi, Uyttendaele, Zitnick, and Ulissi]{barros2026open}
Luis Barros-Luque, Muhammed Shuaibi, Xiang Fu, Brandon~M Wood, Misko Dzamba,
  Meng Gao, Ammar Rizvi, Matt Uyttendaele, C~Lawrence Zitnick, and Zachary~W
  Ulissi.
\newblock The open materials 2024 (omat24) inorganic materials dataset and
  models.
\newblock \emph{Nature Computational Science}, 6\penalty0 (6):\penalty0
  642--652, 2026.
\newblock \doi{10.1038/s43588-026-00996-w}.

\bibitem[H{\"o}llmer et~al.(2025)H{\"o}llmer, Egg, Martirossyan, Fuemmeler,
  Shui, Gupta, Prakash, Roitberg, Liu, Karypis, et~al.]{hollmer2025open}
Philipp H{\"o}llmer, Thomas Egg, Maya~M Martirossyan, Eric Fuemmeler, Zeren
  Shui, Amit Gupta, Pawan Prakash, Adrian Roitberg, Mingjie Liu, George
  Karypis, et~al.
\newblock Open materials generation with stochastic interpolants.
\newblock \emph{arXiv preprint arXiv:2502.02582}, 2025.
\newblock \doi{10.48550/arXiv.2502.02582}.

\bibitem[Chanussot et~al.(2021)Chanussot, Das, Goyal, Lavril, Shuaibi, Riviere,
  Tran, Heras-Domingo, Ho, Hu, et~al.]{chanussot2021open}
Lowik Chanussot, Abhishek Das, Siddharth Goyal, Thibaut Lavril, Muhammed
  Shuaibi, Morgane Riviere, Kevin Tran, Javier Heras-Domingo, Caleb Ho, Weihua
  Hu, et~al.
\newblock Open catalyst 2020 (oc20) dataset and community challenges.
\newblock \emph{Acs Catalysis}, 11\penalty0 (10):\penalty0 6059--6072, 2021.
\newblock \doi{10.1021/acscatal.0c04525}.

\bibitem[Kaplan et~al.(2025)Kaplan, Liu, Qi, Ko, Deng, Riebesell, Ceder,
  Persson, and Ong]{kaplan2025foundational}
Aaron~D Kaplan, Runze Liu, Ji~Qi, Tsz~Wai Ko, Bowen Deng, Janosh Riebesell,
  Gerbrand Ceder, Kristin~A Persson, and Shyue~Ping Ong.
\newblock A foundational potential energy surface dataset for materials.
\newblock \emph{arXiv preprint arXiv:2503.04070}, 2025.
\newblock \doi{10.48550/arXiv.2503.04070}.

\bibitem[How et~al.(2026)How, Febrer, Chong, Mazitov, Bigi, Kellner,
  Pozdnyakov, and Ceriotti]{how2026universal}
Wei~Bin How, Pol Febrer, Sanggyu Chong, Arslan Mazitov, Filippo Bigi, Matthias
  Kellner, Sergey Pozdnyakov, and Michele Ceriotti.
\newblock A universal machine learning model for the electronic density of
  states.
\newblock \emph{Digital Discovery}, 5\penalty0 (4):\penalty0 1635--1649, 2026.
\newblock \doi{10.1039/D5DD00557D}.

\bibitem[Pozdnyakov and Ceriotti(2023)]{pozdnyakov2023smooth}
Sergey Pozdnyakov and Michele Ceriotti.
\newblock Smooth, exact rotational symmetrization for deep learning on point
  clouds.
\newblock \emph{Advances in Neural Information Processing Systems},
  36:\penalty0 79469--79501, 2023.
\newblock \doi{10.52202/075280-3478}.

\bibitem[Mazitov et~al.(2025)Mazitov, Bigi, Kellner, Pegolo, Tisi, Fraux,
  Pozdnyakov, Loche, and Ceriotti]{mazitov2025pet}
Arslan Mazitov, Filippo Bigi, Matthias Kellner, Paolo Pegolo, Davide Tisi,
  Guillaume Fraux, Sergey Pozdnyakov, Philip Loche, and Michele Ceriotti.
\newblock Pet-mad as a lightweight universal interatomic potential for advanced
  materials modeling.
\newblock \emph{Nature Communications}, 16\penalty0 (1):\penalty0 10653, 2025.
\newblock \doi{10.1038/s41467-025-65662-7}.

\bibitem[Dathar et~al.(2017)Dathar, Balachandran, Kent, Rondinone, and
  Ganesh]{dathar2017li}
Gopi Krishna~Phani Dathar, Janakiraman Balachandran, Paul~RC Kent, Adam~J
  Rondinone, and P~Ganesh.
\newblock Li-ion site disorder driven superionic conductivity in solid
  electrolytes: a first-principles investigation of $\beta$-li 3 ps 4.
\newblock \emph{Journal of Materials Chemistry A}, 5\penalty0 (3):\penalty0
  1153--1159, 2017.
\newblock \doi{10.1039/C6TA07713G}.

\bibitem[Kimura et~al.(2019)Kimura, Kato, Hotehama, Sakuda, Hayashi, and
  Tatsumisago]{kimura2019li3sbs4}
Takuya Kimura, Atsutaka Kato, Chie Hotehama, Atsushi Sakuda, Akitoshi Hayashi,
  and Masahiro Tatsumisago.
\newblock Preparation and characterization of lithium ion conductive {Li3SbS4}
  glass and glass-ceramic electrolytes.
\newblock \emph{Solid State Ionics}, 333:\penalty0 45--49, 2019.
\newblock \doi{10.1016/j.ssi.2019.01.017}.

\bibitem[Xiong et~al.(2018)Xiong, Liu, Rong, Wang, McDaniel, and
  Chen]{xiong2018na3sbse4}
Shan Xiong, Zhantao Liu, Haibo Rong, Hai Wang, Malte McDaniel, and Hailong
  Chen.
\newblock {Na3SbSe4-xSx} as sodium superionic conductors.
\newblock \emph{Scientific Reports}, 8:\penalty0 9146, 2018.
\newblock \doi{10.1038/s41598-018-27301-8}.

\bibitem[Ong et~al.(2013)Ong, Mo, Richards, Miara, Lee, and
  Ceder]{ong2013lgps_family}
Shyue~Ping Ong, Yifei Mo, William~Davidson Richards, Lincoln Miara, Hyung~Suk
  Lee, and Gerbrand Ceder.
\newblock Phase stability, electrochemical stability and ionic conductivity of
  the {Li10{\(\pm\)}1MP2X12} ({M = Ge, Si, Sn, Al or P}, and {X = O, S or Se})
  family of superionic conductors.
\newblock \emph{Energy \& Environmental Science}, 6:\penalty0 148--156, 2013.
\newblock \doi{10.1039/C2EE23355J}.

\bibitem[Wang et~al.(2015)Wang, Richards, Ong, Miara, Kim, Mo, and
  Ceder]{wang2015design_principles}
Yan Wang, William~Davidson Richards, Shyue~Ping Ong, Lincoln~J. Miara, Jae~Chul
  Kim, Yifei Mo, and Gerbrand Ceder.
\newblock Design principles for solid-state lithium superionic conductors.
\newblock \emph{Nature Materials}, 14:\penalty0 1026--1031, 2015.
\newblock \doi{10.1038/nmat4369}.

\bibitem[Levine et~al.(2025)Levine, Shuaibi, Spotte-Smith, Taylor, Hasyim,
  Michel, Batatia, Cs{\'a}nyi, Dzamba, Eastman, Frey, Fu, Gharakhanyan,
  Krishnapriyan, Rackers, Raja, Rizvi, Rosen, Ulissi, Vargas, Zitnick, Blau,
  and Wood]{levine2025open}
Daniel~S. Levine, Muhammed Shuaibi, Evan Walter~Clark Spotte-Smith, Michael~G.
  Taylor, Muhammad~R. Hasyim, Kyle Michel, Ilyes Batatia, G{\'a}bor Cs{\'a}nyi,
  Misko Dzamba, Peter Eastman, Nathan~C. Frey, Xiang Fu, Vahe Gharakhanyan,
  Aditi~S. Krishnapriyan, Joshua~A. Rackers, Sanjeev Raja, Ammar Rizvi,
  Andrew~S. Rosen, Zachary Ulissi, Santiago Vargas, C.~Lawrence Zitnick,
  Samuel~M. Blau, and Brandon~M. Wood.
\newblock The open molecules 2025 ({OMol25}) dataset, evaluations, and models.
\newblock \emph{arXiv preprint arXiv:2505.08762}, 2025.
\newblock \doi{10.48550/arXiv.2505.08762}.
\newblock \url{https://arxiv.org/abs/2505.08762}.

\bibitem[Sun et~al.(2020)Sun, Zhang, Banerjee, Bao, Barbry, Blunt, Bogdanov,
  Booth, Chen, Cui, et~al.]{sun2020recent}
Qiming Sun, Xing Zhang, Samragni Banerjee, Peng Bao, Marc Barbry, Nick~S Blunt,
  Nikolay~A Bogdanov, George~H Booth, Jia Chen, Zhi-Hao Cui, et~al.
\newblock Recent developments in the pyscf program package.
\newblock \emph{J. Chem. Phys.}, 153\penalty0 (2):\penalty0 024109, 2020.
\newblock \doi{10.1063/5.0006074}.

\bibitem[Li et~al.(2025)Li, Sun, Zhang, and Chan]{li2025introducing}
Rui Li, Qiming Sun, Xing Zhang, and Garnet Kin-Lic Chan.
\newblock Introducing gpu acceleration into the python-based simulations of
  chemistry framework.
\newblock \emph{J. Phys. Chem. A}, 129\penalty0 (5):\penalty0 1459--1468, 2025.
\newblock \doi{10.1021/acs.jpca.4c05876}.

\bibitem[Laio and Parrinello(2002)]{laio2002escaping}
Alessandro Laio and Michele Parrinello.
\newblock Escaping free-energy minima.
\newblock \emph{Proc. Natl. Acad. Sci.}, 99\penalty0 (20):\penalty0
  12562--12566, 2002.
\newblock \doi{10.1073/pnas.202427399}.

\bibitem[Barducci et~al.(2008)Barducci, Bussi, and
  Parrinello]{barducci2008well}
Alessandro Barducci, Giovanni Bussi, and Michele Parrinello.
\newblock Well-tempered metadynamics: a smoothly converging and tunable
  free-energy method.
\newblock \emph{Phys. Rev. Lett.}, 100\penalty0 (2):\penalty0 020603, 2008.
\newblock \doi{10.1103/PhysRevLett.100.020603}.

\bibitem[Bussi et~al.(2007)Bussi, Donadio, and Parrinello]{bussi2007canonical}
Giovanni Bussi, Davide Donadio, and Michele Parrinello.
\newblock Canonical sampling through velocity rescaling.
\newblock \emph{J. Chem. Phys.}, 126\penalty0 (1):\penalty0 014101, 2007.
\newblock \doi{10.1063/1.2408420}.

\bibitem[Vargas et~al.(2002)Vargas, Garza, Hay, and Dixon]{Vargas2002Alanine}
Rubicelia Vargas, Jorge Garza, Benjamin~P. Hay, and David~A. Dixon.
\newblock Conformational study of the alanine dipeptide at the {MP2} and {DFT}
  levels.
\newblock \emph{J. Phys. Chem. A}, 106\penalty0 (13):\penalty0 3213--3218,
  2002.
\newblock \doi{10.1021/jp013952f}.

\bibitem[Fadda and Woods(2013)]{Fadda2013Alanine}
Elisa Fadda and Robert~J. Woods.
\newblock Contribution of the empirical dispersion correction on the
  conformation of short alanine peptides obtained by gas-phase {QM}
  calculations.
\newblock \emph{Can. J. Chem.}, 91\penalty0 (9):\penalty0 859--865, 2013.
\newblock \doi{10.1139/cjc-2012-0542}.

\bibitem[Sevgen et~al.(2018)Sevgen, Giberti, Sidky, Whitmer, Galli, Gygi, and
  de~Pablo]{Sevgen2018Hierarchical}
Emre Sevgen, Federico Giberti, Hythem Sidky, Jonathan~K. Whitmer, Giulia Galli,
  Francois Gygi, and Juan~J. de~Pablo.
\newblock Hierarchical coupling of first-principles molecular dynamics with
  advanced sampling methods.
\newblock \emph{J. Chem. Theory Comput.}, 14\penalty0 (6):\penalty0 2881--2888,
  2018.
\newblock \doi{10.1021/acs.jctc.8b00192}.

\bibitem[Hensley et~al.(2015)Hensley, Wang, and McEwen]{hensley2015phenol}
Alyssa J.~R. Hensley, Yong Wang, and Jean-Sabin McEwen.
\newblock Phenol deoxygenation mechanisms on fe(110) and pd(111).
\newblock \emph{ACS Catal.}, 5\penalty0 (2):\penalty0 523--536, 2015.
\newblock \doi{10.1021/cs501403w}.

\bibitem[Lim et~al.(2025)Lim, Park, Walsh, and Kim]{lim2025accelerating}
Yunsung Lim, Hyunsoo Park, Aron Walsh, and Jihan Kim.
\newblock Accelerating co$_2$ direct air capture screening for metal-organic
  frameworks with a transferable machine learning force field.
\newblock \emph{Matter}, 8\penalty0 (7):\penalty0 102203, 2025.
\newblock \doi{10.1016/j.matt.2025.102203}.

\bibitem[Hjorth~Larsen et~al.(2017)Hjorth~Larsen, Mortensen, Blomqvist,
  Castelli, Christensen, Du{\l}ak, Friis, Groves, Hammer, Hargus,
  et~al.]{hjorth2017atomic}
Ask Hjorth~Larsen, Jens~J{\o}rgen Mortensen, Jakob Blomqvist, Ivano~E Castelli,
  Rune Christensen, Marcin Du{\l}ak, Jesper Friis, Michael~N Groves, Bj{\o}rk
  Hammer, Cory Hargus, et~al.
\newblock The atomic simulation environment—a python library for working with
  atoms.
\newblock \emph{Journal of Physics: Condensed Matter}, 29\penalty0
  (27):\penalty0 273002, 2017.
\newblock \doi{10.1088/1361-648X/aa680e}.

\bibitem[Batatia et~al.(2022)Batatia, Kovacs, Simm, Ortner, and
  Cs{\'a}nyi]{batatia2022mace}
Ilyes Batatia, David~P Kovacs, Gregor Simm, Christoph Ortner, and G{\'a}bor
  Cs{\'a}nyi.
\newblock Mace: Higher order equivariant message passing neural networks for
  fast and accurate force fields.
\newblock \emph{Adv. Neural Inf. Process. Syst.}, 35:\penalty0 11423--11436,
  2022.
\newblock \doi{10.48550/arXiv.2206.07697}.

\bibitem[Bigi et~al.(2026)Bigi, Pegolo, Mazitov, Schmidt, and
  Ceriotti]{bigi2026pushing}
Filippo Bigi, Paolo Pegolo, Arslan Mazitov, Jonathan Schmidt, and Michele
  Ceriotti.
\newblock Pushing the limits of unconstrained machine-learned interatomic
  potentials.
\newblock \emph{arXiv preprint arXiv:2601.16195}, 2026.
\newblock \doi{10.48550/arXiv.2601.16195}.

\end{thebibliography}
\bibliographystyle{assets/plainnat}
}


\clearpage

\appendix

{\Huge \textbf{Supporting Information}}
\startcontents[appendices] 
\section*{Contents}
\printcontents[appendices]{l}{1}{\setcounter{tocdepth}{3}}

\newpage
\setcounter{table}{0}
\renewcommand{\thetable}{S\arabic{table}}%
\setcounter{figure}{0}
\renewcommand{\thefigure}{S\arabic{figure}}%
\setcounter{lstlisting}{0}
\renewcommand{\thelstlisting}{S\arabic{lstlisting}}%

\section{Execution graph implementation}
\label{si_sec:execution_graph}

\begin{figure}[!p]
    \centering
    \includegraphics[
        width=\linewidth,
        height=\textheight,
        keepaspectratio
    ]{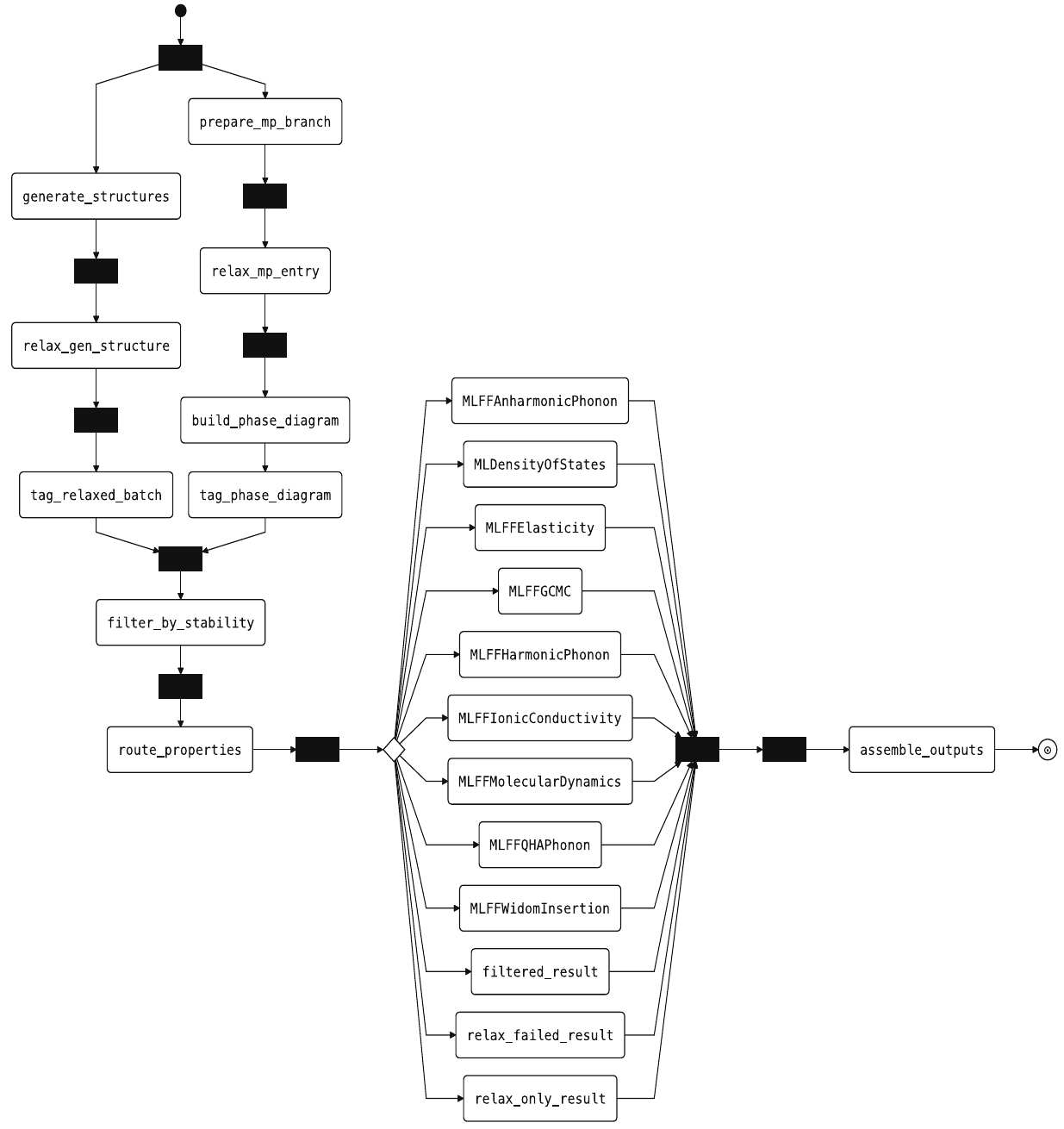}
    \caption{\textbf{Execution graph for the \ac{mlip} workflow in \elagenteP{}.}
      White rectangles represent executable graph nodes. Circles mark the graph start and end points. Filled black rectangles denote graph-control operators, such as fan-out from streamed inputs or synchronization of parallel branches. The diamond denotes the decision point that dispatches routed requests to property or terminal result nodes. The workflow starts with a candidate-structure branch and an optional Materials Project reference branch. Candidate structures are relaxed with the selected \ac{mlip}; reference entries, when enabled, are relaxed with the same \ac{mlip} and used to construct an internally consistent phase diagram. The two branches synchronize before phase-stability filtering. Passing structures are routed by an \ac{llm} property router to typed property-node configurations, while failed, filtered, or relaxation-only cases emit terminal result objects. Property outputs are grouped by relaxed structure, assembled into per-structure results, written to deterministic artifacts, and optionally persisted to the knowledge graph after graph completion.}
    \label{si_fig:mlip_execution_graph}
\end{figure}

This section describes the implemented execution graph (Figure~\ref{si_fig:mlip_execution_graph}) for the \ac{mlip} workflow, including candidate and reference structure streaming, branch synchronization, phase-stability filtering, property routing, result aggregation, artifact writing, and optional knowledge-graph persistence.

\subsection{Graph inputs and shared state}

The \ac{mlip} workflow is exposed to the outer conversational agent as the typed tool \texttt{run\_mlff\_workflow}. The tool receives an \texttt{MLFFWorkflowRequest}, whose fields specify the structure source or target compositions, \ac{mlip} backend, density-of-states backend, relaxation settings, optional phase-stability filtering settings, structure-generation settings, and whether selected semantic outputs should be pushed to the knowledge graph. The graph therefore does not execute directly from free text. The outer agent first converts the user request into a typed workflow invocation; the graph then executes deterministic Python nodes, with multiple instances of an inner \ac{llm} router used only to select downstream property nodes for each relaxed structure.

At runtime, the graph stores a shared workflow state containing the summarized request from the original user prompt, the resolved \ac{mlip} input object, selected calculator settings, relaxation parameters, phase-stability thresholds, relaxed-structure outcomes, routing decisions, phase-diagram build records, workflow summary messages, and workspace artifact locations. This state is graph-local execution state. Runtime streams, calculator instances, and transient scheduling objects are not serialized as durable records.

\subsection{Candidate and reference streams}

The upstream graph contains two branches from the start node: the candidate branch and the reference branch. They start concurrently if both are triggered from user intent.

The candidate branch begins at the stream node \texttt{generate\_structures}. Supported identifier types include \texttt{cif}, \texttt{xyz}, \texttt{name}, \texttt{smiles}, \texttt{material\_formula}, and \texttt{mp\_id}. Given an input path, the graph yields each \texttt{.cif} or \texttt{.xyz} file in the agent's workspace directory. When the \texttt{structure\_generator\_composition} composition is specified, it invokes the configured OMatG structure generator, flattens the resulting batches, and yields each candidate structure individually. A mapped edge then spawns one \texttt{relax\_gen\_structure} task for each emitted candidate. The relaxed candidate outcomes are collected and tagged as a \texttt{TaggedRelaxedBatch} before entering the downstream synchronization join.

The reference branch begins at \texttt{prepare\_mp\_branch}. This branch is active only when \texttt{enable\_phase\_stability\_filter=True} and \texttt{identifier\_type=structure\_generator\_composition}. In that case, the branch queries the Materials Project for entries for the candidate chemical system, converts each entry's structure to the internal structure representation, and streams one reference entry at a time. A mapped edge fans out one \texttt{relax\_mp\_entry} task per reference entry. The resulting relaxed objects are collected and used by \texttt{build\_phase\_diagram} to construct a \texttt{pymatgen.PhaseDiagram} from \ac{mlip}-evaluated reference energies. The result is tagged as a \texttt{TaggedPhaseDiagram}.

When phase-stability filtering is disabled, or when the input type is not \texttt{structure\_generator\_composition}, the reference branch emits a placeholder token indicating that no phase diagram is active. This allows the same join topology to complete, after which the downstream stability-filtering node acts as a pass-through. Thus, high-throughput runs over supplied files or materials identifiers still use the same graph topology, but they do not build a Materials Project reference hull in the current implementation.

\subsection{Join synchronization and phase-stability filtering}

The candidate and reference branches are synchronized by a single join that receives two tagged inputs: a \texttt{TaggedRelaxedBatch} from the candidate branch and a \texttt{TaggedPhaseDiagram} from the reference branch. The explicit tagged union prevents the stability filter from running before both the relaxed candidate batch and the phase-diagram token (or disabled-phase-diagram token) are available.

The \texttt{filter\_by\_stability} stream node then emits a relaxed outcome for each candidate. If no phase diagram is enabled, each relaxed outcome is passed downstream unchanged. If a phase diagram is enabled, the node evaluates the relaxed candidate with the same \ac{mlip} calculator, computes the energy above the \ac{mlip}-consistent hull, and compares it with \texttt{e\_above\_hull\_cutoff} (the cutoff parameter defined by the agent based on the user request). Structures above the cutoff emit filtered outcomes and skip property calculations; structures below the cutoff continue to routing. When \texttt{max\_structures\_per\_composition\_by\_e\_above\_hull} is set, the graph can additionally retain only the lowest-energy-above-hull structures for each reduced composition. Relaxation failures also continue through the graph as typed failure outcomes rather than being silently dropped.

\subsection{Property routing and decision dispatch}

Each outcome emitted by \texttt{filter\_by\_stability} is mapped to the stream node \texttt{route\_properties}. Filtered structures yield the marker \texttt{\_\_FILTERED\_\_}; failed relaxations yield \texttt{\_\_RELAX\_FAILED\_\_}. Successful relaxed structures are passed to the inner \ac{llm} property router together with the summarized user request and relaxed-structure metadata. The router returns a \texttt{RouterDecision} containing a list of typed property-node configurations. These configurations form a discriminated union whose discriminator is the \texttt{node\_name} field, so node-specific inputs are validated before the corresponding calculation node is executed.

The decision edge dispatches each routed property configuration to the matching property node. The current dispatch targets are: \texttt{MLFFMolecularDynamics}, \texttt{MLFFIonicConductivity}, \texttt{MLFFElasticity}, \texttt{MLFFHarmonicPhonon}, \texttt{MLFFQHAPhonon}, \texttt{MLFFAnharmonicPhonon}, \texttt{MLDensityOfStates}, \texttt{MLFFWidomInsertion}, and \texttt{MLFFGCMC}. The same decision edge also routes terminal markers to \texttt{relax\_failed\_result}, \texttt{filtered\_result}, or \texttt{relax\_only\_result}. This ensures that each input structure produces a typed output, even when relaxation fails, phase-stability filtering rejects the structure, or the router selects no downstream property calculation.

\subsection{Aggregation, artifacts, and knowledge-graph persistence}

Every terminal or property node returns a \texttt{PropertyResultCarrier} containing the structure \ac{iri}, chemical formula, typed result payload, and summary notes. A \texttt{property\_collect} join groups these carriers by structure \ac{iri}. The \texttt{final\_collect} join merges the grouped dictionaries from parallel branches, and \texttt{assemble\_outputs} constructs a single \texttt{MLFFOutput} for each relaxed or terminal structure.

The workflow writes deterministic workspace artifacts after graph aggregation. For successful relaxed structures, the output includes a per-structure \texttt{output.json} bundle that contains the relaxed structure and the realized property payloads. The run can also produce a structure-level \texttt{property\_summary.csv}, pressure-resolved \texttt{gcmc\_results.csv}, phase-diagram support files, trajectories, logs, and failure snapshots. When \texttt{update\_graph=True}, selected semantic roots are pushed to the knowledge graph after the graph completes: the workflow run, relaxation outcomes, routing decisions, phase-diagram build steps, and typed property results. Runtime streams and calculator objects are intentionally not persisted as knowledge-graph entities.

\subsection{Parallelism model}

The graph exposes parallelism at three levels. First, candidate structures are streamed by \texttt{generate\_structures}, and a mapped edge launches one candidate-relaxation task per emitted structure. Second, when phase-stability filtering is active, Materials Project reference entries are streamed by \texttt{prepare\_mp\_branch}, and a mapped edge launches a reference relaxation task for each entry. These reference relaxations run concurrently with the candidate branch before synchronization at \texttt{phase\_diagram\_collect}. Third, after filtering, \texttt{route\_properties} can emit multiple property configurations for each relaxed structure, and the mapped decision edge fans these out to independent property nodes.

For \(N\) candidate structures, \(M\) Materials Project reference entries, and \(K_i\) routed property nodes for candidate \(i\), the graph can expose up to \(N\) candidate-relaxation tasks, \(M\) reference-relaxation tasks, and \(\sum_i K_i\) property tasks to the asynchronous execution layer. Actual hardware concurrency is bounded by the Python runtime, scheduler limits, available accelerator slots, and calculator-safety guards. The implementation therefore separates graph-level parallelism from physical resource assignment: the graph defines independent work units and synchronization points, while the runtime controls calculator reuse, GPU-slot assignment, and thread-safe model initialization.

\begin{table}[H]
\centering
\small
\caption{Execution semantics of the main \ac{mlip} graph stages.}
\label{si_tab:mlip_graph_semantics}
\begin{tabularx}{\linewidth}{p{0.20\linewidth}p{0.25\linewidth}X}
\toprule
\textbf{Graph stage} & \textbf{Main node(s)} & \textbf{Execution semantics} \\
\midrule
Candidate streaming &
\texttt{generate\_structures} &
Streams one or more candidate structures from identifiers, directories, or composition-conditioned OMatG generation. \\
Candidate relaxation &
\texttt{relax\_gen\_structure}, \texttt{relax\_collect} &
Mapped relaxation fans out over emitted candidates; a join collects relaxed candidate outcomes. \\
Reference branch &
\texttt{prepare\_mp\_branch}, \texttt{relax\_mp\_entry}, \texttt{mp\_collect} &
When enabled for generated compositions, streams Materials Project reference entries and relaxes them with the same \ac{mlip}; otherwise emits a sentinel. \\
Phase diagram &
\texttt{build\_phase\_diagram}, \texttt{phase\_diagram\_collect} &
Builds an \ac{mlip}-consistent phase diagram from relaxed reference entries and synchronizes the reference and candidate branches through tagged join tokens. \\
Stability filtering &
\texttt{filter\_by\_stability} &
Passes candidates through when no phase diagram is enabled; otherwise emits passing or filtered outcomes based on energy above hull. \\
Property routing &
\texttt{route\_properties} &
Uses the inner \ac{llm} router to produce typed property-node configurations for each successful relaxed structure; emits markers for failed or filtered cases. \\
Decision dispatch &
Decision edge to property and terminal nodes &
Dispatches each routed configuration to the matching property node, or terminal markers to failure, filtered, or relaxation-only result nodes. \\
Aggregation and output &
\texttt{property\_collect}, \texttt{final\_collect}, \texttt{assemble\_outputs} &
Groups property results by structure \ac{iri}, assembles one output object per structure, writes artifacts, and optionally pushes selected semantic roots to the knowledge graph. \\
\bottomrule
\end{tabularx}
\end{table}

\section{Additional information for case studies and benchmarks}
\label{si_sec:case_studies}
\subsection{Phase stability analyses for materials discovery}

\begin{figure}[H]
    \centering
    \includegraphics[width=0.9\linewidth]{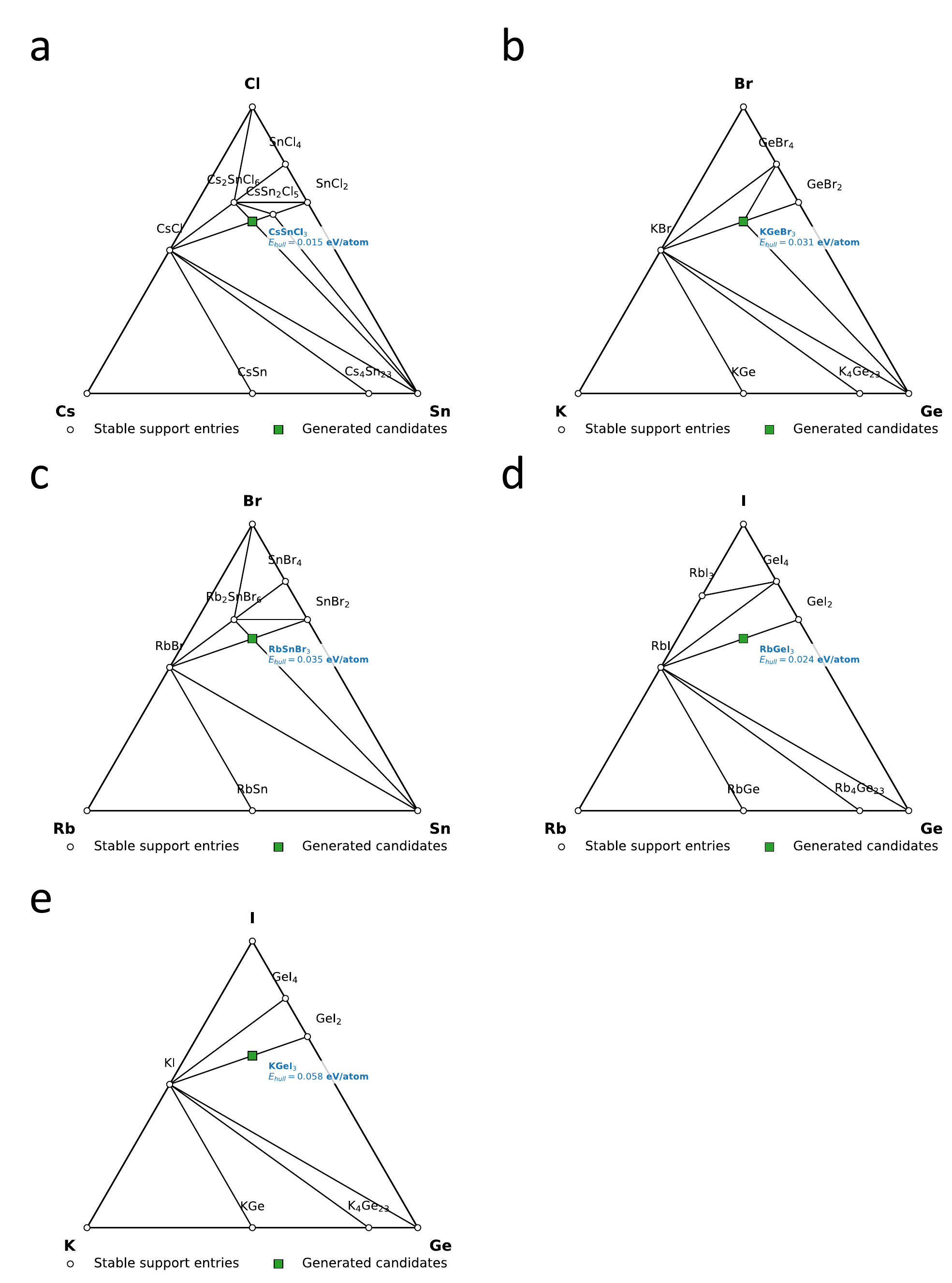}
    \caption{\textbf{Convex-hull analysis of generated lead-free halide perovskite candidates.} Ternary phase diagrams for the lowest-\(E_\mathrm{hull}\) generated candidates: \textbf{a}, \(\mathrm{CsSnCl}_3\); \textbf{b}, \(\mathrm{KGeBr}_3\); \textbf{c}, \(\mathrm{RbSnBr}_3\); \textbf{d}, \(\mathrm{RbGeI}_3\); and \textbf{e}, \(\mathrm{KGeI}_3\). Open circles denote stable support entries, black lines denote \ac{mlip} hull tie lines, and green squares mark the selected generated candidates.}
    \label{fig:phase_diagram_pervoskites}
\end{figure}

\begin{figure}[H]
    \centering
    \includegraphics[width=0.9\linewidth]{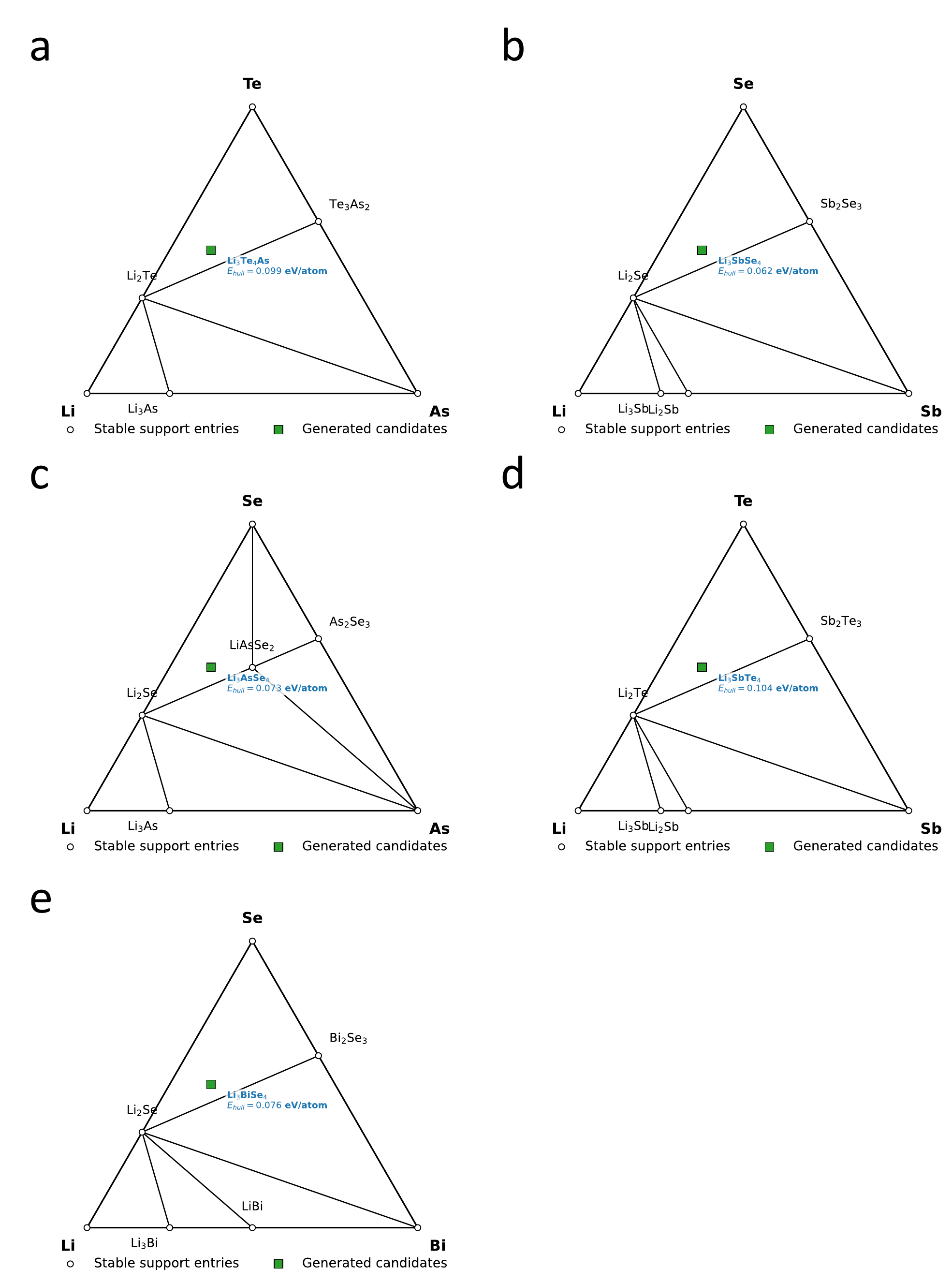}
    \caption{\textbf{Convex-hull analysis of generated Li-ion conductor candidates.}
    Ternary phase diagrams for the lowest-\(E_\mathrm{hull}\) generated candidates: \textbf{a}, \(\mathrm{Li}_{12}\mathrm{As}_4\mathrm{Te}_{16}\); \textbf{b}, \(\mathrm{Li}_{12}\mathrm{Sb}_4\mathrm{Se}_{16}\); \textbf{c}, \(\mathrm{Li}_{12}\mathrm{As}_4\mathrm{Se}_{16}\); \textbf{d}, \(\mathrm{Li}_{12}\mathrm{Sb}_4\mathrm{Te}_{16}\); and \textbf{e}, \(\mathrm{Li}_{12}\mathrm{Bi}_4\mathrm{Se}_{16}\). Open circles denote stable support entries, black lines denote \ac{mlip} hull tie lines, and green squares mark the selected generated candidates.
    }
    \label{fig:phase_diagram_ionnic_conductors}
\end{figure}

\subsection{Molecular case studies }

\begin{figure}[H]
    \centering
    \includegraphics[width=1.0\linewidth]{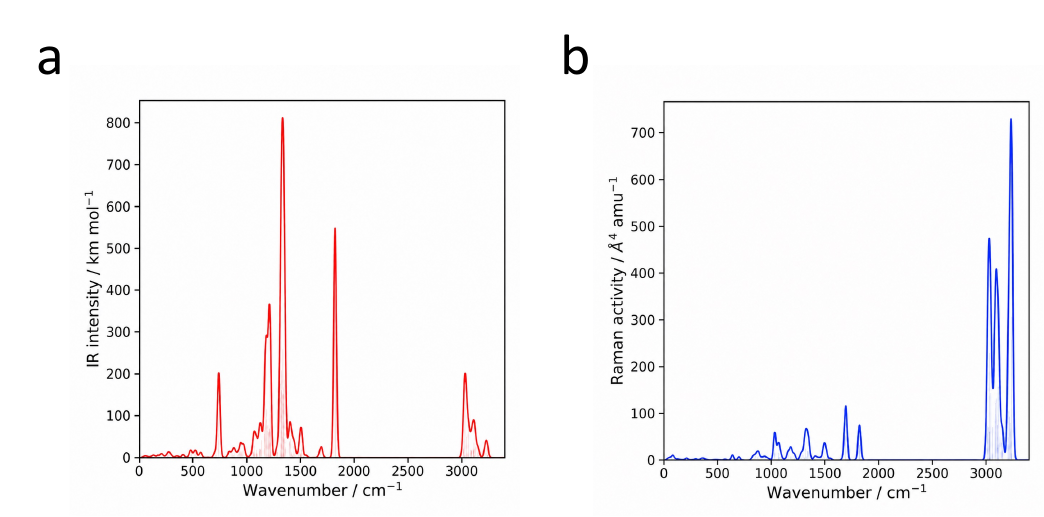}
    \caption{\textbf{DFT vibrational spectra of the lowest-energy reranked conformer.}
    \textbf{a,} Harmonic infrared spectrum and \textbf{b,} Raman spectrum of the lowest-energy conformer obtained after DFT reranking of conformers generated using the MLIP-based conformational search. The selected structure was subsequently optimized and its harmonic vibrational frequencies, IR intensities, and Raman activities were calculated with PySCF at the $\omega$B97X-D4/def2-TZVP level of theory. Gaussian broadening was applied to the discrete vibrational transitions for visualization.}
    \label{fig:low_energy_conformer_search_si}
\end{figure}

\begin{figure}[H]
    \centering
    \includegraphics[width=1.0\linewidth]{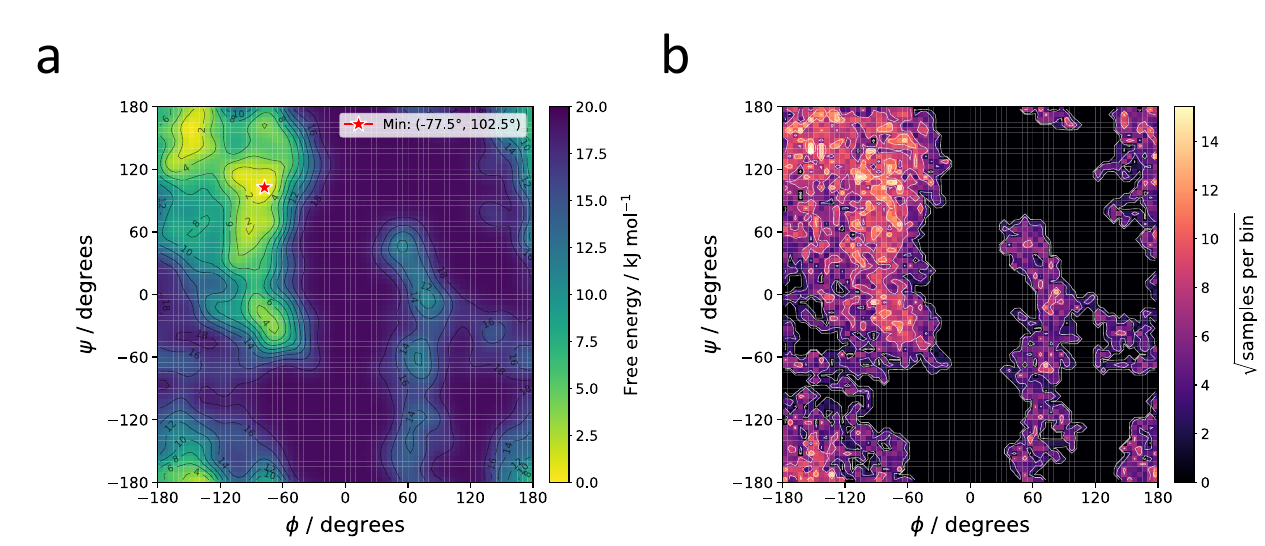}
    \caption{\textbf{Agentic-driven metadynamics of alanine dipeptide.}
    \textbf{a,} Two-dimensional free-energy surface projected onto the backbone dihedral angles $\phi$ and $\psi$, obtained from well-tempered metadynamics at 300~K using Orb-OMOL. The free energy is reported relative to the global minimum, located at approximately $(\phi,\psi)=(-77.5^\circ,102.5^\circ)$ and marked by the red star.
    \textbf{b,} Corresponding sampling distribution in the $\phi$--$\psi$ space. The color scale reports the square root of the number of sampled configurations per bin, highlighting the regions explored during the metadynamics trajectory.} 
    \label{fig:metadynamics_si}
\end{figure}

\subsection{Widom insertion results}



\begin{figure}[H]
    \centering
    \includegraphics[width=0.5\linewidth]{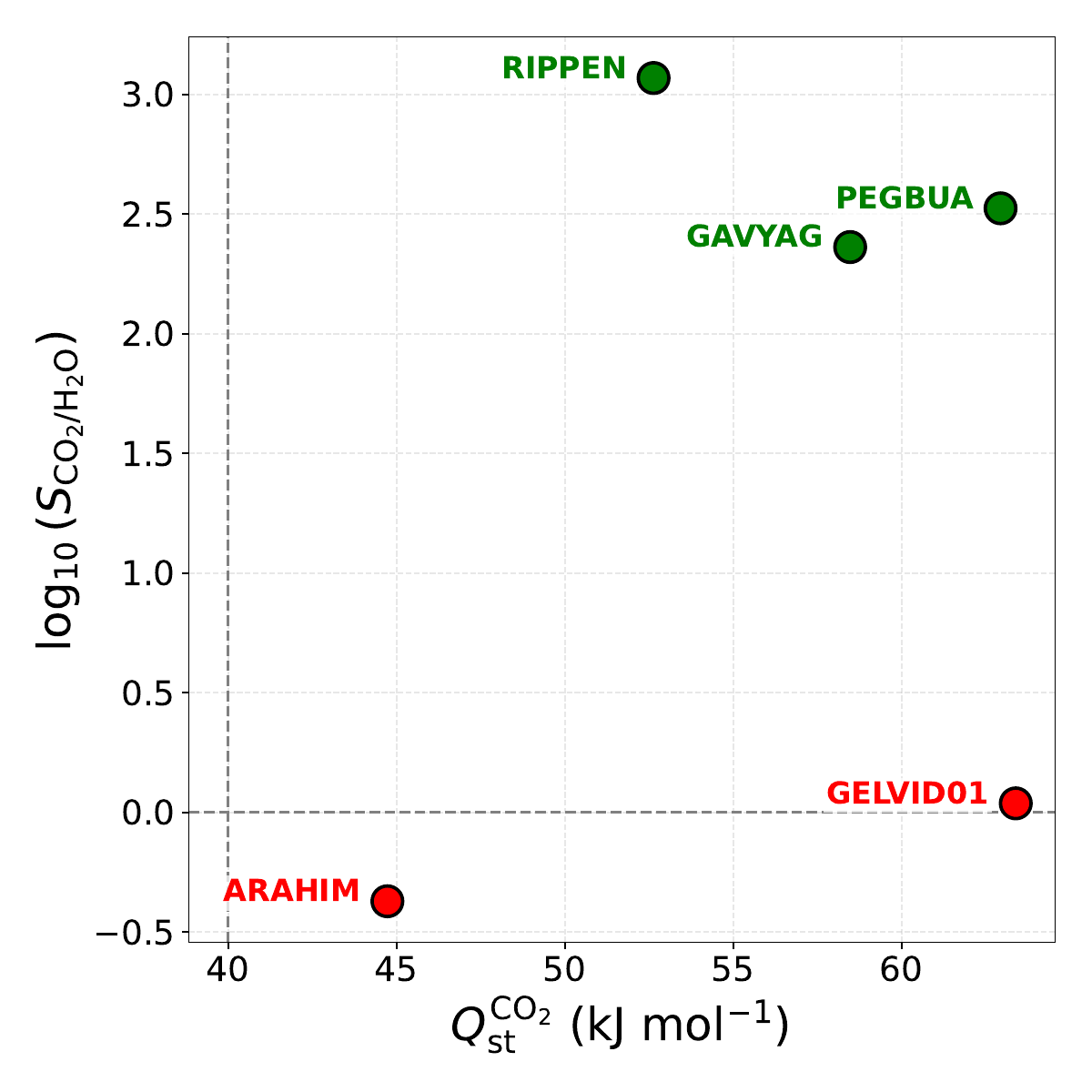}
    \caption{\textbf{Adsorption properties obtained from Widom insertion calculations.} Scatter plot showing the relationship between the CO$_2$ isosteric heat of adsorption, $Q_{\mathrm{st}}^{\mathrm{CO_2}}$, and the CO$_2$/H$_2$O Henry selectivity, $\log_{10}(S_{\mathrm{CO_2/H_2O}})$, for five representative MOFs taken from ref.\citenum{lim2025accelerating}. Green points indicate MOFs whose predicted adsorption properties agree well with the reference values, whereas red points indicate MOFs that show modest deviations. }
    \label{fig:widom_insertion_qst_vs_selectivity}
\end{figure}

\subsection{Agentic benchmark of MLIP models}

\prompt{

Please write a Python script that uses the available tools and functions in our \texttt{MLFF\_workflow} to benchmark MatterSim, Orb-OMAT, and MACE-OMAT for predicting the bulk modulus and shear modulus of each structure in [insert folder\_path]. Use the corresponding file name as the mp-id for each entry. Save the final results for each model to a CSV file containing at least the following columns: \texttt{mp\_id}, \texttt{bulk\_modulus}, and \texttt{shear\_modulus}. Generate a plot in your working space directory to compare the MLIP calculated results against the ground truth in \texttt{elasticity\_1000.csv} with the unit in GPa from [insert folder\_path].}

\begin{figure}[H]
    \centering
    \includegraphics[width=1.0\linewidth]{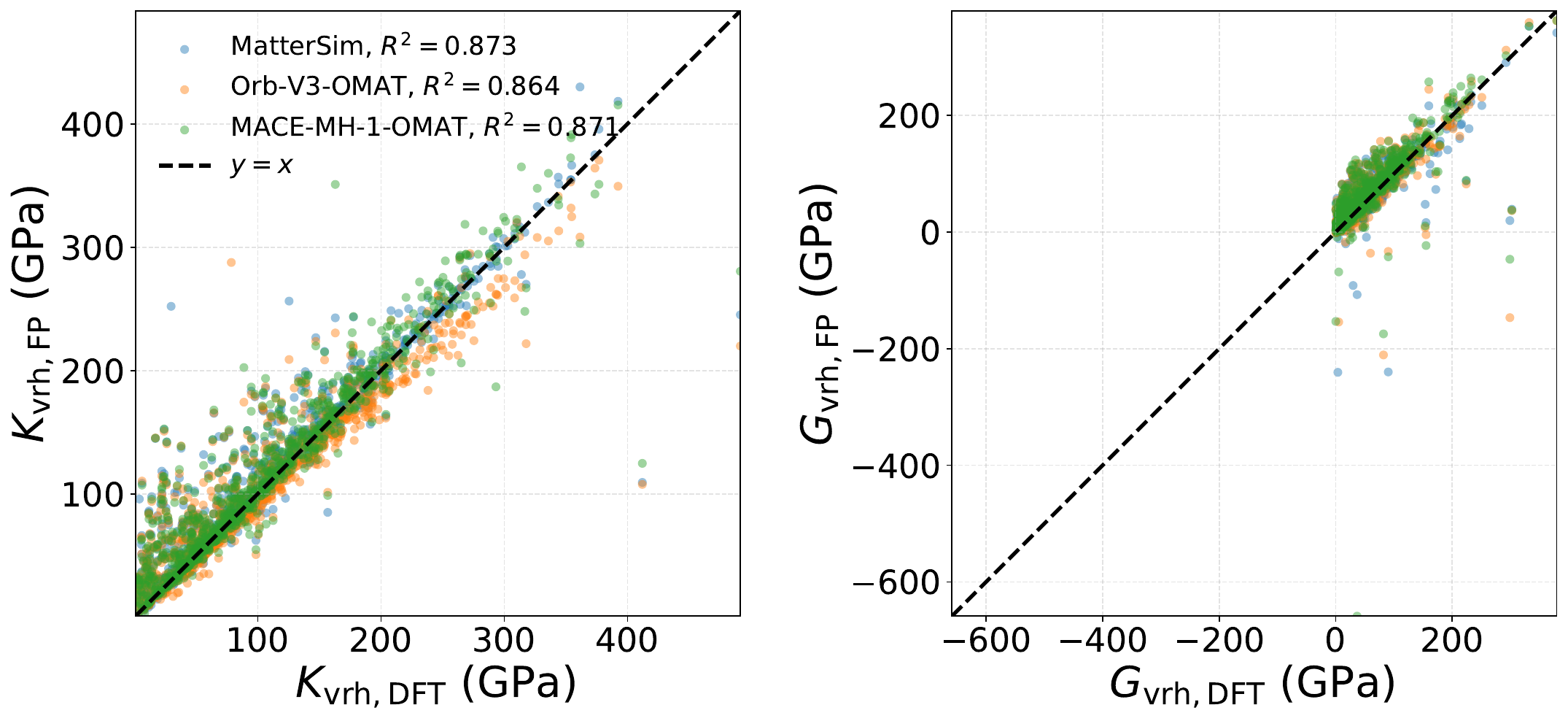}
    \caption{\textbf{Agentic benchmarks of MLIPs for elastic-property prediction. Parity plots of MLIP-predicted and DFT reference Voigt--Reuss--Hill bulk moduli ($K_{\mathrm{vrh}}$) and shear moduli ($G_{\mathrm{vrh}}$). Predictions from MatterSim, Orb-V3-OMAT, and MACE-MH-1-OMAT were obtained using the agentic workflow for automated structure relaxation and elasticity calculations. The dashed line denotes ideal agreement ($y=x$).}}
    \label{fig:elasticity_parity_plots}
\end{figure}

\subsection{Comparison with a skill-based coding agent}
\label{si_subsec:atomisticskills}
For all \potente{} runs in this benchmark, ChatGPT~5.5 was used as the primary \ac{llm} for interpreting the user request and coordinating workflow execution, while ChatGPT~4.1 was used by the routing agents. The comparison coding agent used Claude Code with \texttt{sonnet-4.6}.
\prompt{

Run UUID: [insert UUID]

Please read the MOF structures from [insert folder\_path] and perform GCMC simulations using CO2 molecules at 298 K and 1 bar. Run 5,000 equilibration steps followed by 5,000 production steps. Save the results into a CSV file.

\textcolor{blue}{\textit{\textbf{Note:} The specific UUIDs used for the five runs were c8f6ad74-3f17-5802-b4f9-5b8f1f6d4782, 9cbf7e6f-cd71-59e5-8f8d-c0c5e61f4c2e, f41f5f11-44bf-5a9d-aec0-94d7d6603f8b, 9993100d-046b-4211-98b1-50e74f4b696e, 5118d853-a2e9-4c11-b37d-2a7d2e48e044. These UUIDs were not expected to affect agent behaviour; they were added solely to invalidate user prompt caching, which could otherwise compromise the independence of repeated runs.}}
}

Across five independent runs by \potente{} for the processed HKUST-1 structure, C${72}$H${24}$Cu${12}$O${48}$, the MACE-DAC-1 model was consistently used for GCMC simulations. The predicted CO$_2$ uptake was therefore consistent, giving an average loading of 73.53 $\pm$ 3.30 g/L, corresponding to 4.60 $\pm$ 0.21 CO$_2$ molecules per cell. The average interaction energy was $-1.09 \pm 0.04$ eV, with an average interaction energy per adsorbate of $-0.247 \pm 0.002$ eV/CO$_2$. These values are physically reasonable for CO$_2$ adsorption in HKUST-1 under ambient conditions and indicate reproducible adsorption estimates across repeated short GCMC simulations. Because the production length was limited to 5,000 GCMC steps, these results represent screening-level adsorption predictions instead of fully converged adsorption isotherm data. 

We further submitted the same prompt to Claude Code equipped with the AtomisticSkills scientific skills library~\cite{deng2026harnessing}, which we use here as a representative open-ended skill-based coding agent. Fairchem was disabled so that the calculations used MACE-based models, improving comparability with \potente{}. Fairchem was disabled so that the values produced used MACE models and are more comparable to \potente{}. In all five runs, the agent completed the workflow by reading the HKUST-1 CIF file and the \texttt{chem-sorption-gcmc} skill, selecting the MACE-MH-1 model, and doing the GCMC calculations. They then checked the convergence traces and wrote the final CSV and JSON outputs. 
All the runs used MACE-MH-1 because it was the recommended default, which is mentioned as an in-line comment inside the MACE's wrapper module, \texttt{mace\_wrapper.py}\footnote{\url{https://github.com/learningmatter-mit/AtomisticSkills/blob/ca695e1b89ea48613291c88bd8c73e71af2025c7/src/utils/mlips/mace/mace_wrapper.py\#L18}}, listed beside other MACE models. 
Run CC\#2 additionally relaxed the structure using the \texttt{relax\_structure} skill from the MACE MCP server. Run CC\#4 thought about relaxing the structure before deciding to skip the step. The other runs did not consider the skill.
The five completed runs gave an average CO$_2$ loadings of 1.25 $\pm$ 0.57 mmol/g, with an average estimated heats of adsorption of 13.81 $\pm$ 0.77 kJ/mol. The resulting 8--9\% variation in loading is consistent with the expected stochastic variation from short GCMC production windows. The discrepancy between the AtomisticSkills and \potente{} results is attributed to the use of different MACE models.

\begin{table}[H]
\centering
\small
\caption{Comparison of GCMC runs (HKUST-1, CO$_2$, 298\,K, 1\,bar) when prompting \potente{} \textit{vs.} Claude Code equipped with AtomisticSkills.}
\label{si_tab:gcmc_runs}
\begin{tabularx}{\linewidth}{l*{7}{>{\centering\arraybackslash}X}}
\toprule
\textbf{Run} & \textbf{Session time (min)} & \textbf{Total tool calls} & \textbf{\ac{llm} API calls} & \textbf{CO$_2$ loading (mmol/g)} & \textbf{$Q_{\mathrm{st}}$ (kJ/mol)} & \textbf{Total cost (USD)} & \textbf{Total tokens} \\
\midrule
P\#1   & 6.35 & 1 & 2 & 2.080 & 22.40 & 0.150 & 26,574 \\
P\#2   & 6.18  & 1 & 2 & 1.804 & 22.56 & 0.154 & 26,851 \\
P\#3   & 6.34  & 1 & 2 & 2.031 & 22.30  & 0.149  & 26,614 \\
P\#4  & 6.39  & 1 & 2 & 1.824  & 22.94 & 0.158 & 26,982 \\
P\#5   & 6.43 & 1 & 2 & 1.717 & 23.77 & 0.150 & 26,599 \\
\hline
CC\#1   & 19.13 & 21 & 17 & 2.16 & 14.41 & 0.47 & 656,240 \\
CC\#2   & 17.97 & 21 & 18 & 0.763 & 12.46 & 0.65 & 851,115 \\
CC\#3   & 15.73 & 17 & 15 & 1.399 & 13.54 & 0.65 & 828,348 \\
CC\#4   & 17.50 & 30 & 27 & 1.385 & 14.06 & 0.88 & 1,253,094 \\
CC\#5   & 18.58 & 24 & 21 & 0.54 & 14.6 & 0.54 & 820,080\\
\hline
Potente Avg. & 6.41 & 1 & 3 & 1.865 & 17.21 & 0.12 & 22,541\\
CC Avg. & 17.78 & 22.60 & 19.60 & 1.249 & 13.81 & 0.64 & 881,775 \\
\bottomrule
\end{tabularx}
\end{table}

\subsection{Reproducibility benchmarks}

\begin{figure}[H]
    \centering
    \includegraphics[width=1.0\linewidth]{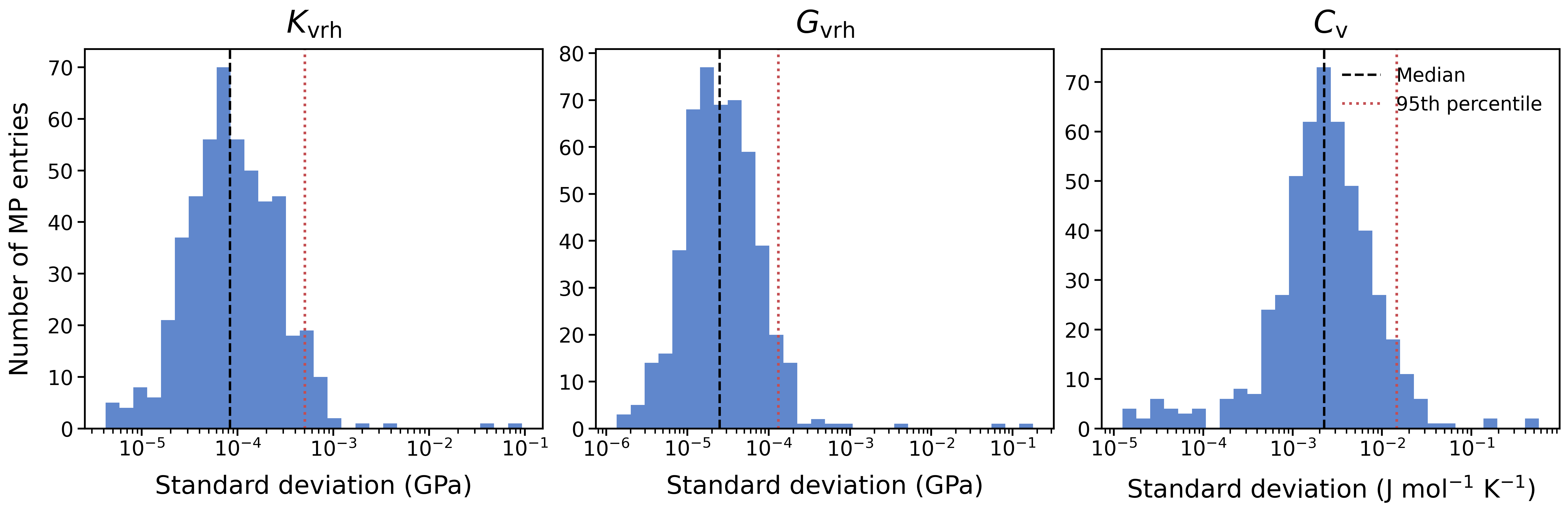}
    \caption{\textbf{Reproducibility of \elagenteP{} for high-throughput calculations.} Distributions of the standard deviations obtained across five independent trials for the Voigt--Reuss--Hill bulk modulus ($K_{\mathrm{VRH}}$), shear modulus ($G_{\mathrm{VRH}}$), and harmonic heat capacity at 300~K ($C_V$). The dashed and dotted vertical lines indicate the median and 95th percentile of the standard-deviation distributions, respectively.} 
    \label{fig:potente_repro_histogram}
\end{figure}

\begin{figure}
    \centering
    \includegraphics[width=1.0\linewidth]{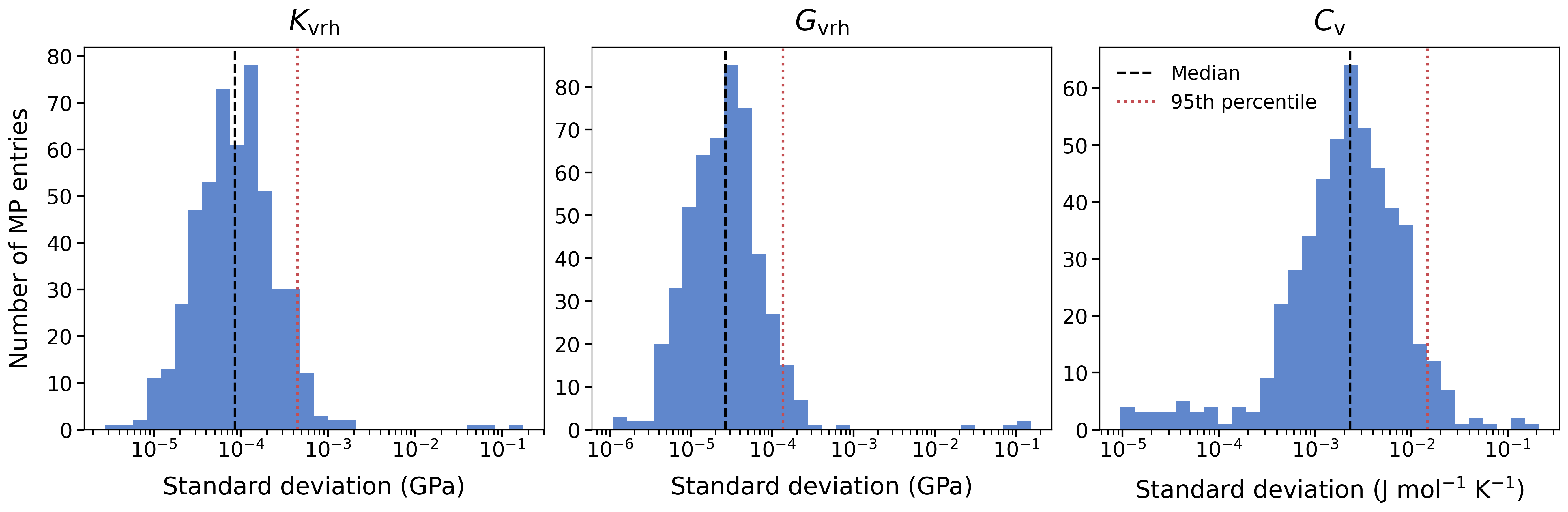}
    \caption{\textbf{Reproducibility of the manually scripted high-throughput workflow.}
    Distributions of the standard deviations obtained from five independent executions of the same Python script using identical computational parameters for the Voigt--Reuss--Hill bulk modulus ($K_{\mathrm{VRH}}$), shear modulus ($G_{\mathrm{VRH}}$), and harmonic heat capacity at 300~K ($C_V$). The dashed and dotted vertical lines indicate the median and 95th percentile of the standard-deviation distributions, respectively.}
    \label{fig:human_repro_histogram}
\end{figure}

\begin{figure}
    \centering
    \includegraphics[width=1.0\linewidth]{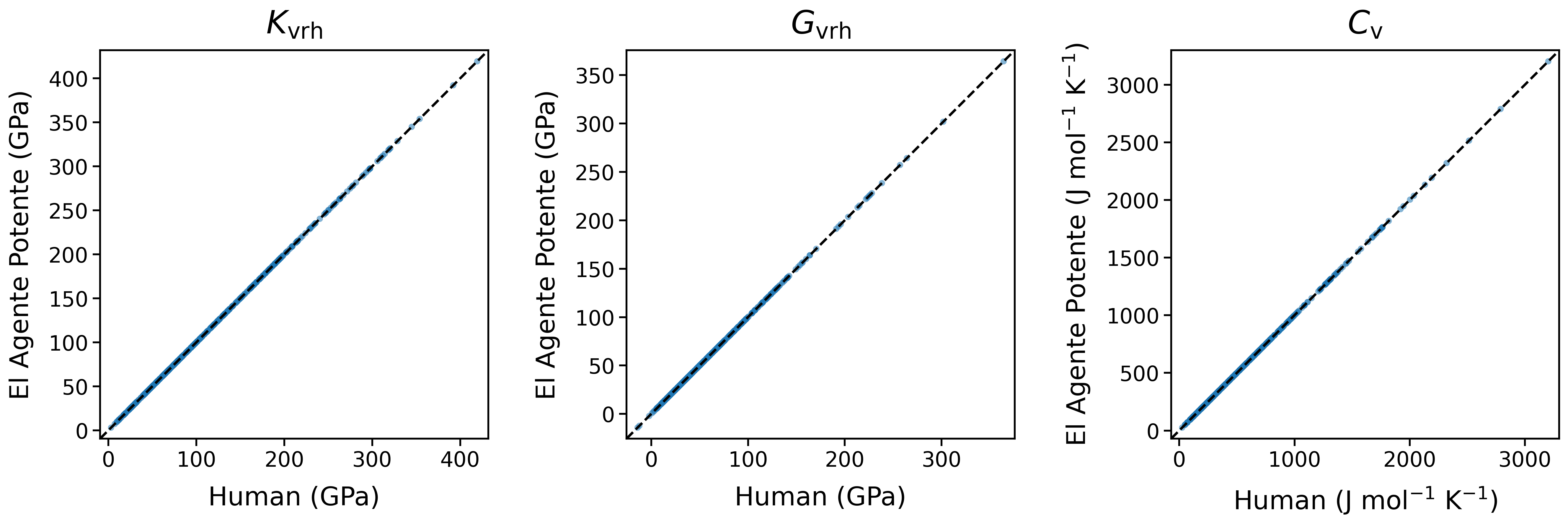}
    \caption{\textbf{Parity between \elagenteP{} and the manually scripted workflow for high-throughput calculations.} Comparison of the mean properties obtained over five independent executions of each workflow for the Voigt--Reuss--Hill bulk modulus ($K_{\mathrm{VRH}}$), shear modulus ($G_{\mathrm{VRH}}$), and harmonic heat capacity at 300~K ($C_V$). Each point represents the five-run average for one material, and the dashed line indicates perfect agreement ($y=x$).}
    \label{fig:parity_plots_high_throughout}
\end{figure}

\newpage
\section{Abridged chat transcripts}

The following abridged transcripts reproduce the user prompts, model-provided Thoughts, and final responses, while omitting tool calls and their return
payloads.

Local filesystem prefixes have been replaced with \path{<workspace_path_redacted>} and \path{<repository_path_redacted>}; filenames, run identifiers, and relative artifact paths are otherwise preserved.

\label{si_sec:chat_transcripts}
\subsection{Materials discovery}

\subsubsection{Perovskite discovery}
\label{si:chat_perovskite}
\prompt{
Search existing literatures and propose five novel and reasonable compositions for promising perovskite materials with the general formula ABX$_3$ using the \texttt{ALEX\_MP20\_CSP} structure generator. For each proposed composition, generate ten candidate structures, evaluate their phase stability. For each composition, select the structure with the lowest energy above the convex hull and calculate its bandgap.
}

\providecommand{\tightlist}{%
  \setlength{\itemsep}{0pt}\setlength{\parskip}{0pt}}

\potentereslong{
\begin{thoughtBox}
\faLightbulb\ \underline{\textbf{Thoughts}}

\textbf{Evaluating energy calculations}

I’m thinking about how to approach this. If the energies are enormous, I should use 5.0 eV/atom. This could help bypass filtering while still computing the hull. I need to mention that the selected lowest energy configuration isn't necessarily stable. It’ll be important to evaluate phase stability. I’m considering running five compositions in parallel, using commentary tools, and hopefully returning an output list. Let's call for that!
\end{thoughtBox}

\begin{thoughtBox}
\faLightbulb\ \underline{\textbf{Thoughts}}

\textbf{Evaluating material gaps and stability}

I’m working through some material data, focusing on gaps and stability. CsSnCl$_3$ has the best stability but a gap of 1.31, while RbGeI$_3$ has a balanced gap of 1.61 and low energy above the hull. KGeI$_3$ might be metastable but has a significant gap. I'm considering materials like RbGeI$_3$, KGeBr$_3$, and CsSnCl$_3$ for absorption purposes. I need to note the energy above hull thresholds, around 0.05, suggesting potential synthesizability. I'll mention my literature search with proper citations later.
\end{thoughtBox}

\begin{thoughtBox}
\faLightbulb\ \underline{\textbf{Thoughts}}

\textbf{Gathering source citations}

I need to include that I’ve searched existing literature. Lead-free ABX$_3$ halide perovskites are well-known, and I'm considering how cation size and tolerance factors stabilize 3D structures and influence bandgaps, with trends from Cl to I and comparing Sn to Pb. I should look for reliable sources for exact snippets. If I used the web for statements, I’ll need to cite those results, although the ``Materials Project'' is more of a tool reference rather than a web source, so no citations there. I'll mention the term ``novel'' too!
\end{thoughtBox}

I searched the recent ABX$_3$ perovskite literature and then ran the requested \texttt{ALEX\_MP20\_CSP} screening workflow. The literature basis I used was: lead-free ABX$_3$ halide perovskites remain an active optoelectronic materials space; A-site size/tolerance-factor arguments are still widely used to assess formability; and halide/B-site substitutions are known to strongly tune band gaps, e.g. Cl/Br/I and Sn/Ge trends~\cite{mitro2022electronic,rahman2022rbsnx,rahman2025first,ayalew2026exploring}. I therefore focused on \textbf{all-inorganic, Pb-free, charge-balanced ABX$_3$ halides} using K/Rb/Cs on A, Ge/Sn on B, and Br/I/Cl on X. I treat these as \textbf{reasonable underexplored screening proposals}, not as a claim that every formula has never appeared in any database or publication.

\paragraph{\textbf{Workflow executed}}
For each composition:

\begin{itemize}
\tightlist
\item Structure generator: \textbf{\texttt{ALEX\_MP20\_CSP}}
\item Candidates generated: \textbf{10}
\item Relaxation: \textbf{MACE-OMAT}, cell relaxed
\item Phase stability: MLFF-relaxed Materials Project chemical-system phase diagram
\item Selection: \textbf{lowest energy above hull} among the 10 generated candidates
\item Band gap: \textbf{PETMADDOS density-of-states band gap}
\item I used a high hull cutoff internally so that candidates were ranked rather than prematurely discarded; the reported selected structure is the lowest-energy-above-hull candidate.
\end{itemize}

\paragraph{\textbf{Results: selected lowest-hull structure per composition}}
\begin{center}
\begingroup
\scriptsize
\renewcommand{\arraystretch}{1.15}
\begin{tabularx}{\linewidth}{@{}lXcccX@{}}
\textbf{Composition} & \textbf{A/B/X rationale} & \textbf{Selected $E_\mathrm{hull}$ (eV/atom)} & \textbf{$E_\mathrm{hull}$ range (eV/atom)} & \textbf{Band gap (eV)} & \textbf{Selected structure IRI} \\
\hline
$\mathrm{CsSnCl}_3$ & Cs$^+$/Sn$^{2+}$/Cl$^-$, Pb-free Sn chloride analogue & \textbf{0.0151} & 0.0151--0.0185 & \textbf{1.31} & \path{ConceptualAtoms_ce3c8fb7-f8d3-428a-992b-4af354941774} \\
$\mathrm{RbGeI}_3$ & Rb$^+$/Ge$^{2+}$/I$^-$, Ge-based iodide absorber candidate & \textbf{0.0241} & 0.0241--0.0307 & \textbf{1.61} & \path{ConceptualAtoms_b86073d3-3fac-4bb7-9d5b-fd4114314b07} \\
$\mathrm{KGeBr}_3$ & K$^+$/Ge$^{2+}$/Br$^-$, smaller A with Br for moderate gap & \textbf{0.0306} & 0.0306--0.0362 & \textbf{1.51} & \path{ConceptualAtoms_9c40e224-0f57-40a5-b739-605ae4754239} \\
$\mathrm{RbSnBr}_3$ & Rb$^+$/Sn$^{2+}$/Br$^-$, Sn bromide narrow-gap candidate & \textbf{0.0351} & 0.0351--0.0417 & \textbf{0.774} & \path{ConceptualAtoms_b9eb7962-8969-4788-bd61-782ae16ec4ae} \\
$\mathrm{KGeI}_3$ & K$^+$/Ge$^{2+}$/I$^-$, compact-A Ge iodide variant & \textbf{0.0579} & 0.0579--0.0638 & \textbf{1.28} & \path{ConceptualAtoms_0395f82b-cf7d-432b-9eef-1326095eb759} \\
\end{tabularx}
\endgroup
\end{center}

\paragraph{\textbf{Shortlist interpretation}}
\textbf{Most promising from this screen: RbGeI$_3$ and KGeBr$_3$.} They combine low predicted energy above hull with band gaps in the \textasciitilde{}1.5--1.6 eV range, which is attractive for absorber-oriented screening. The general literature also supports Ge/Sn and halide substitution as a rational route for tuning lead-free ABX$_3$ perovskite electronic structure~\cite{mitro2022electronic,rahman2022rbsnx,rahman2025first,ayalew2026exploring}.

\textbf{Best phase-stability score: CsSnCl$_3$,} with the lowest selected energy above hull at \textasciitilde{}0.015 eV/atom, though it is less ``novel'' than the Ge-centered K/Rb proposals and chloride chemistry may have different optical tradeoffs.

\textbf{Lower-priority but still reasonable: RbSnBr$_3$} has good predicted metastability but a smaller computed gap, while \textbf{KGeI$_3$} has a useful \textasciitilde{}1.28 eV gap but the highest energy above hull among the five selected structures.

These are MLFF/DOS screening results; the next validation step would be higher-level DFT relaxation + band structure, ideally including spin--orbit coupling for Sn/I/Br-containing systems.
}

\paragraph{\textbf{Execution summary.}}
\potente{} used a built-in literature search to identify chemically reasonable Pb-free ABX$_3$ halide candidates and then launched five independent MLFF workflows, one for each proposed composition. Each workflow generated ten candidate structures with \texttt{ALEX\_MP20\_CSP}, relaxed them with MACE-OMAT, evaluated their energies above the convex hull, retained the lowest-hull candidate, and predicted its density-of-states band gap with PETMADDOS. All five workflow calls completed successfully. Intermediate search results, tool calls, tool-return payloads, internal paths, and call identifiers are omitted from this abridged transcript.

\subsubsection{Superionic lithium-ion conductor discovery}
\label{si:chat_superionic_conductors}
\prompt{
Search the existing literature and propose five novel and reasonable compositions for Li-ion superionic conductors with the formula Li$_{12}$M$_4$X$_{16}$ using the structure generator. For each proposed composition, generate 10 candidate structures and evaluate their phase stability.

For each composition, select the structure with the lowest energy above the convex hull and calculate its ionic conductivity at 300 K. To perform the conductivity calculation, replicate the unit cell by a factor of 2 along the x, y, and z directions, then run NPT MD simulations with 100 ps of equilibration followed by a 100 ps production run. Use a logging interval of 10 steps.
}

\providecommand{\tightlist}{%
  \setlength{\itemsep}{0pt}\setlength{\parskip}{0pt}}

\potentereslong{
\begin{thoughtBox}
\faLightbulb\ \underline{\textbf{Thoughts}}

\textbf{Gathering final citations}

I need to finalize my results with proper citations since I used web sources. It’s essential to reference current literature to support my findings. The search results may not be perfect, but I can cite information like Li$_3$PS$_4$ thio-LISICON, the high-temperature phase, and ORNL Li$_{3+x}$[Si$_x$P]S$_4$, which has higher conductivity. I’ll also look into design principles and find more direct sources for selenium or soft anions. Let’s search for these additional references!
\end{thoughtBox}

\begin{thoughtBox}
\faLightbulb\ \underline{\textbf{Thoughts}}

\textbf{Choosing threshold wording}

I think I should aim to avoid using the word “typically” or possibly rephrase it to say “using a pragmatic 0.1 threshold.” But, wait, I just realized that the threshold I'm referencing wasn't actually used by the user. So, it might be better to clarify or rethink how I present this information. I'm trying to ensure that I communicate effectively and accurately for the user’s understanding.
\end{thoughtBox}

I searched the Li-ion solid-electrolyte literature, then used the structure generator/MLFF workflow to screen five Li$_{12}$M$_4$X$_{16}$ candidates. I interpreted Li$_{12}$M$_4$X$_{16}$ as the 4-formula-unit cell of \textbf{Li$_3$MX$_4$}, with M mostly group-15 / pnictogen cations and X = Se/Te chalcogenides.

\textbf{Literature basis.} I avoided the canonical benchmark \textbf{Li$_3$PS$_4$ / Li$_{12}$P$_4$S$_{16}$} because it is already heavily studied: Li$_3$PS$_4$ is a thio-LISICON solid electrolyte, nanoporous $\beta$-Li$_3$PS$_4$ has reported strongly enhanced room-temperature Li conductivity, and Li$_{3+x}$[Si$_x$P$_{1-x}$]S$_4$ variants can exceed 1 mS/cm via Li-site splitting/disorder. (\url{https://www.ornl.gov/publication/anomalous-high-ionic-conductivity-nanoporous-b-li3ps4?utm_source=openai}) I also used the broader thio-LISICON lesson that replacing oxide frameworks with more polarizable sulfide frameworks can raise conductivity by orders of magnitude, and that pnictogen substitution in Li$_3$SbS$_4$-type thio-LISICONs can strongly alter Li transport. (\url{https://pubs.rsc.org/en/content/articlehtml/2021/ma/d1ma00082a?utm_source=openai}) Based on that, I proposed heavier / softer \textbf{selenide and telluride Li$_3$MX$_4$ analogues} as reasonable exploratory superionic-conductor candidates.

\paragraph{\textbf{Workflow run details}}
For each composition I generated \textbf{10 candidate structures}, relaxed them with \textbf{MACE-OMAT}, evaluated energy above the convex hull, retained the \textbf{single lowest-$E_\mathrm{hull}$} structure, then ran \textbf{Li ionic conductivity at 300 K} using:

\begin{itemize}
\tightlist
\item \textbf{$2 \times 2 \times 2$ cell replication}
\item \textbf{NPT Nose--Hoover-chain MD}
\item \textbf{1 fs timestep}
\item \textbf{100 ps equilibration = 100,000 steps}
\item \textbf{100 ps production = 100,000 steps}
\item \textbf{log interval = 10 steps}
\item direct 300 K conductivity estimate from MD
\end{itemize}

I set a high phase-stability cutoff internally so that all generated candidates could be ranked; only the \textbf{top-1 lowest energy-above-hull structure per composition} was passed to MD.

\paragraph{\textbf{Results}}
\begin{center}
\begingroup
\scriptsize
\renewcommand{\arraystretch}{1.15}
\begin{adjustbox}{max width=\linewidth}
\begin{tabular}{rlllrrrr}
\textbf{Rank by stability} & \textbf{Proposed composition} & \textbf{Reduced form} & \textbf{Lowest $E_\mathrm{hull}$ (eV/atom)} & \textbf{$E_\mathrm{hull}$ range (eV/atom)} & \textbf{Candidates $\leq 0.10$ eV/atom} & \textbf{$\sigma_\mathrm{ion}$, 300 K (mS/cm)} & \textbf{Li diffusion coefficient (cm$^2$/s)} \\
\hline
1 & \textbf{Li$_{12}$Sb$_4$Se$_{16}$} & Li$_3$SbSe$_4$ & \textbf{0.0618} & 0.0618--0.1110 & 7/10 & \textbf{6.56} & $6.71 \times 10^{-8}$ \\
2 & \textbf{Li$_{12}$As$_4$Se$_{16}$} & Li$_3$AsSe$_4$ & \textbf{0.0728} & 0.0728--0.1602 & 5/10 & \textbf{9.14} & $1.01 \times 10^{-7}$ \\
3 & \textbf{Li$_{12}$Bi$_4$Se$_{16}$} & Li$_3$BiSe$_4$ & \textbf{0.0755} & 0.0755--0.1146 & 5/10 & \textbf{5.89} & $6.16 \times 10^{-8}$ \\
4 & \textbf{Li$_{12}$As$_4$Te$_{16}$} & Li$_3$AsTe$_4$ & \textbf{0.0993} & 0.0993--0.1402 & 1/10 & \textbf{43.16} & $5.56 \times 10^{-7}$ \\
5 & \textbf{Li$_{12}$Sb$_4$Te$_{16}$} & Li$_3$SbTe$_4$ & \textbf{0.1038} & 0.1038--0.1495 & 0/10 & \textbf{65.44} & $8.70 \times 10^{-7}$ \\
\end{tabular}
\end{adjustbox}
\endgroup
\end{center}

\paragraph{\textbf{Interpretation}}
\textbf{Best balanced candidates:}

\begin{enumerate}
\tightlist
\item \textbf{Li$_{12}$Sb$_4$Se$_{16}$ / Li$_3$SbSe$_4$} --- lowest energy above hull in this screen, with respectable 300 K conductivity of \textbf{6.56 mS/cm}.
\item \textbf{Li$_{12}$As$_4$Se$_{16}$ / Li$_3$AsSe$_4$} --- second-lowest hull energy and slightly higher conductivity, \textbf{9.14 mS/cm}.
\item \textbf{Li$_{12}$Bi$_4$Se$_{16}$ / Li$_3$BiSe$_4$} --- also near-hull, but Bi(V)-selenide chemistry should be treated cautiously despite the reasonable MLFF screening result.
\end{enumerate}

\textbf{Highest conductivity but higher stability risk:}

\begin{itemize}
\tightlist
\item \textbf{Li$_{12}$As$_4$Te$_{16}$} and \textbf{Li$_{12}$Sb$_4$Te$_{16}$} gave very high predicted conductivities, \textbf{43.2} and \textbf{65.4 mS/cm}, respectively. However, their hull energies are at or above \textasciitilde{}0.10 eV/atom, especially Sb--Te, so I would treat these as high-risk/high-reward telluride leads.
\end{itemize}

\paragraph{\textbf{Selected structures / artifacts}}
\begin{itemize}
\item \textbf{Li$_{12}$As$_4$Se$_{16}$} selected structure: \url{https://elagente.ca/grafico/ConceptualAtoms_8166f92b-0e2e-4bc7-99b1-348e1ebb5e03}\\
Output JSON: \path{<workspace_path_redacted>/call_mIuB9hPJt1jaT8c8dYM4YzRG/structures/As4Li12Se16_b4b20e82/output.json}
\item \textbf{Li$_{12}$As$_4$Te$_{16}$} selected structure: \url{https://elagente.ca/grafico/ConceptualAtoms_0347cf14-1233-4d40-a774-5fa467d05619}\\
Output JSON: \path{<workspace_path_redacted>/call_6Qv34GYur6HHKtpl5w1GrpNy/structures/As4Li12Te16_087347a8/output.json}
\item \textbf{Li$_{12}$Sb$_4$Se$_{16}$} selected structure: \url{https://elagente.ca/grafico/ConceptualAtoms_6bd83854-77e1-452c-8b94-52ca5a3a4430}\\
Output JSON: \path{<workspace_path_redacted>/call_AnCpe03Qil4ebIr5FoUBDVN2/structures/Li12Sb4Se16_e137b5f9/output.json}
\item \textbf{Li$_{12}$Sb$_4$Te$_{16}$} selected structure: \url{https://elagente.ca/grafico/ConceptualAtoms_c03f2753-5466-4c23-aaad-06663826d343}\\
Output JSON: \path{<workspace_path_redacted>/call_xl6dLwr9hvlkpmzq7kCkg2F0/structures/Li12Sb4Te16_1277ee0d/output.json}
\item \textbf{Li$_{12}$Bi$_4$Se$_{16}$} selected structure: \url{https://elagente.ca/grafico/ConceptualAtoms_64725590-15df-49b1-90dc-4d1099aa5d0c}\\
Output JSON: \path{<workspace_path_redacted>/call_ZxPU7zVfNZpSf6bpgjb7RSUd/structures/Bi4Li12Se16_d59d2432/output.json}
\end{itemize}

\textbf{Recommendation:} prioritize \textbf{Li$_{12}$Sb$_4$Se$_{16}$} and \textbf{Li$_{12}$As$_4$Se$_{16}$} for follow-up, because they combine near-hull stability with mS/cm-scale predicted 300 K Li conductivity.
}

\paragraph{\textbf{Execution summary.}}
\potente{} used three built-in literature-search operations and then launched five independent MLFF workflows, one for each proposed composition. Each workflow generated ten structures, performed MACE-OMAT relaxation and convex-hull assessment, retained the lowest-hull candidate, replicated it into a $2 \times 2 \times 2$ supercell, and ran 100 ps of NPT equilibration followed by 100 ps of NPT production dynamics for the direct 300 K conductivity estimate. All five workflow calls completed successfully. Intermediate search results, tool calls, and tool-return payloads are omitted from this abridged transcript.

\subsection{Molecular energy-landscape exploration}

\subsubsection{Low-energy conformer search and PySCF validation}
\label{si:chat_conformer_pyscf}
\prompt{
Perform low energy conformer search for \texttt{O=C(OCCOCCOCCOCCOC(=O)c1ccccc1)c1ccccc1} using Orb-OMol.
}

\providecommand{\tightlist}{%
  \setlength{\itemsep}{0pt}\setlength{\parskip}{0pt}}

\potentereslong{
\begin{thoughtBox}
\faLightbulb\ \underline{\textbf{Thoughts}}

\textbf{Calculating energy results}

I need to finalize the results with the lowest energy and present a top 10 table. While I could convert eV to kcal using 1 eV = 23.0605 kcal/mol, it's not necessary. I’ll mention the lowest energy values, like 0.023 eV = 0.54 kcal, but no additional calculations are needed. I'll include artifact paths for the conformers, specifically noting the \texttt{minimum\_001.xyz}. The top values will be calculated from MLFF potential energies, and I’ll mention that duplicates might exist.
\end{thoughtBox}

Completed the \textbf{low-energy conformer search} for:

\texttt{O=C(OCCOCCOCCOCCOC(=O)c1ccccc1)c1ccccc1}

using \textbf{Orb-OMOL} as the MLFF backend.

\textbf{System:} neutral singlet, \textbf{C$_{22}$H$_{26}$O$_7$}\\
\textbf{Method:} multiple-minimum Monte Carlo conformer search\\
\textbf{Conformers returned:} 84\\
\textbf{Lowest-energy conformer:} \texttt{minimum\_001.xyz}\\
\textbf{Lowest Orb-OMOL energy:} \textbf{$-37567.706940$ eV}

\paragraph{\textbf{Top low-energy conformers}}
\begin{center}
\renewcommand{\arraystretch}{1.15}
\begin{tabular}{rcrrr}
\textbf{Rank} & \textbf{File} & \textbf{Energy / eV} & \textbf{Relative / eV} & \textbf{Relative / kcal mol$^{-1}$} \\
\hline
1 & \texttt{minimum\_001.xyz} & $-37567.706940$ & 0.000000 & 0.00 \\
2 & \texttt{minimum\_002.xyz} & $-37567.683503$ & 0.023437 & 0.54 \\
3 & \texttt{minimum\_003.xyz} & $-37567.679596$ & 0.027344 & 0.63 \\
4 & \texttt{minimum\_004.xyz} & $-37567.624909$ & 0.082031 & 1.89 \\
5 & \texttt{minimum\_005.xyz} & $-37567.624909$ & 0.082031 & 1.89 \\
6 & \texttt{minimum\_006.xyz} & $-37567.617097$ & 0.089844 & 2.07 \\
7 & \texttt{minimum\_007.xyz} & $-37567.609284$ & 0.097656 & 2.25 \\
8 & \texttt{minimum\_008.xyz} & $-37567.593659$ & 0.113281 & 2.61 \\
9 & \texttt{minimum\_009.xyz} & $-37567.585847$ & 0.121093 & 2.79 \\
10 & \texttt{minimum\_010.xyz} & $-37567.578034$ & 0.128906 & 2.97 \\
\end{tabular}
\end{center}

\paragraph{\textbf{Output artifacts}}
\begin{itemize}
\item \textbf{Run directory:}\\
\path{<workspace_path_redacted>/call_MyMjmpZh1b3isjyggS5cdGqv}
\item \textbf{Run summary:}\\
\path{<workspace_path_redacted>/call_MyMjmpZh1b3isjyggS5cdGqv/run_summary.txt}
\item \textbf{Per-structure JSON:}\\
\path{<workspace_path_redacted>/call_MyMjmpZh1b3isjyggS5cdGqv/structures/C22H26O7_18900d31/output.json}
\item \textbf{Lowest conformer XYZ:}\\
\path{<workspace_path_redacted>/call_MyMjmpZh1b3isjyggS5cdGqv/structures/C22H26O7_18900d31/low_energy_conformers/minimum_001.xyz}
\end{itemize}

Note: these are \textbf{Orb-OMOL MLFF potential energies}, not DFT electronic energies or thermochemical free energies.
}

\paragraph{\textbf{Execution summary.}}
\potente{} invoked the MLFF workflow once for a multiple-minimum Monte Carlo conformer search using Orb-OMOL. The workflow returned 84 conformers and preserved the ranked structures and associated run artifacts. The tool call and its full return payload are omitted from this abridged transcript.

\prompt{
Use PySCF to rerank the ten lowest-energy conformers from \path{<workspace_path_redacted>/call_MyMjmpZh1b3isjyggS5cdGqv/structures/C22H26O7_18900d31/low_energy_conformers/}. Optimize the lowest-energy conformer and calculate its infrared and Raman spectra. Generate the corresponding plots and report the results.
}

\potentereslong{
\begin{thoughtBox}
\faLightbulb\ \underline{\textbf{Thoughts}}

\textbf{Evaluating computational workflow}

I'm thinking about how to set up the workflow with the \texttt{run\_pyscf\_workflow} function. It seems I need to identify multiple files, possibly doing ten parallel calls for the \texttt{SinglePointCalculation}. After that, I'll parse the energies and select the lowest one to run the workflow with optimization and frequency analysis.

I should clarify if the workflow supports the Raman method while ensuring I'm using the right settings. If the GPU runs into issues, I might need to opt for parallel single-point calculations, but I'll check the output carefully before proceeding. Let’s start with getting the energies and see how it goes from there.
\end{thoughtBox}

\begin{thoughtBox}
\faLightbulb\ \underline{\textbf{Thoughts}}

\textbf{Planning to generate plots}

I need to create plots using Python, possibly with CSV files. I have spectra arrays from previous analysis, but I’m concerned about input size. Instead of huge inputs, I could write arrays from results manually, as Python can't access the assistant variable. Perhaps I can copy selected arrays for plotting? It seems like I need full spectra for effective visuals. I could also consider generating plot data in a CSV format, as the arrays are around 159 in length.
\end{thoughtBox}

\begin{thoughtBox}
\faLightbulb\ \underline{\textbf{Thoughts}}

\textbf{Creating reduced frequency table}

I need to focus on including only the necessary data, like frequency, IR intensity, and Raman activity while omitting imaginary frequencies. I’m considering whether to paste arrays directly into a Python script or parse them from a saved JSON file. It seems like I might not have a JSON path available, but I could potentially use a chkfile and PySCF. If that's not possible, I can manually construct a reduced table or generate plots, maybe as SVG or ASCII. I’ll aim to copy the arrays from the output into Python; they might be large but manageable.
\end{thoughtBox}

\begin{thoughtBox}
\faLightbulb\ \underline{\textbf{Thoughts}}

\textbf{Analyzing imaginary frequencies}

It looks like I've found that there are no imaginary frequencies above the specified threshold, which is set at 50 cm$^{-1}$. However, I noticed that my frequency object has small imaginary values like 43j, 28j, and 8j, which are the low-frequency modes below that threshold. So, it's important to mention these three negative modes ($<50$ cm$^{-1}$) present in the raw spectrum as floppy modes, even though they aren't significant. I may want to exclude these negative frequencies when plotting.
\end{thoughtBox}

\begin{thoughtBox}
\faLightbulb\ \underline{\textbf{Thoughts}}

\textbf{Formatting frequency arrays}

I'm planning to paste positive frequency arrays, skipping negatives, and using the same positive frequencies as before. I'll hard code the frequencies and IR intensities while ensuring proper array lengths to avoid mismatches. Maybe Python can help check the lengths and truncate if necessary. I'll need to format the frequency, IR, and Raman activities as strings with whitespace-separated numbers. For the frequency count, I want to make sure it reaches 159. Let's start composing the code and utilize Matplotlib.
\end{thoughtBox}

Completed the PySCF reranking and spectroscopy workflow.

\textbf{Method used:} PySCF \textbf{wB97X-D4/def2-TZVP}, restricted singlet, neutral \textbf{C$_{22}$H$_{26}$O$_7$}.

\paragraph{\textbf{1. PySCF reranking of the 10 Orb-OMOL conformers}}
Single-point PySCF energies were computed on the ten lowest Orb-OMOL conformer geometries.

\begin{center}
\renewcommand{\arraystretch}{1.15}
\begin{tabular}{rcrrr}
\textbf{PySCF rank} & \textbf{Original conformer} & \textbf{Energy / Hartree} & \textbf{$\Delta E$ / eV} & \textbf{$\Delta E$ / kcal mol$^{-1}$} \\
\hline
1 & \texttt{minimum\_001.xyz} & $-1380.735863731819$ & 0.000000 & 0.000 \\
2 & \texttt{minimum\_002.xyz} & $-1380.734715595454$ & 0.031242 & 0.720 \\
3 & \texttt{minimum\_007.xyz} & $-1380.732312970846$ & 0.096621 & 2.228 \\
4 & \texttt{minimum\_004.xyz} & $-1380.732163548079$ & 0.100687 & 2.322 \\
5 & \texttt{minimum\_003.xyz} & $-1380.732126022578$ & 0.101708 & 2.345 \\
6 & \texttt{minimum\_006.xyz} & $-1380.732077070662$ & 0.103040 & 2.376 \\
7 & \texttt{minimum\_005.xyz} & $-1380.731990178972$ & 0.105405 & 2.431 \\
8 & \texttt{minimum\_008.xyz} & $-1380.730916692580$ & 0.134616 & 3.104 \\
9 & \texttt{minimum\_010.xyz} & $-1380.730896298119$ & 0.135171 & 3.117 \\
10 & \texttt{minimum\_009.xyz} & $-1380.730896162124$ & 0.135174 & 3.117 \\
\end{tabular}
\end{center}

\textbf{PySCF-lowest conformer:} \texttt{minimum\_001.xyz}

\paragraph{\textbf{2. Optimization and spectra for PySCF-lowest conformer}}
\texttt{minimum\_001.xyz} was then optimized with PySCF.

\begin{itemize}
\tightlist
\item \textbf{Initial single-point energy:} $-1380.735863731819$ Hartree
\item \textbf{Optimized electronic energy:} \textbf{$-1380.740477950003$ Hartree}
\item \textbf{Thermochemical total enthalpy, H:} $-1380.255766671495$ Hartree
\item \textbf{Thermochemical Gibbs free energy, G:} $-1380.336997404892$ Hartree
\end{itemize}

Frequency analysis found \textbf{no significant imaginary frequencies} under the 50 cm$^{-1}$ threshold used by the workflow. The raw list contains three very low-frequency negative/floppy modes at approximately \textbf{$-43.94$, $-28.29$, and $-8.36$ cm$^{-1}$}, which are below the threshold and typical for flexible large molecules.

\paragraph{\textbf{3. Strongest IR peaks}}
\begin{center}
\renewcommand{\arraystretch}{1.15}
\begin{tabular}{rrr}
\textbf{Rank} & \textbf{Frequency / cm$^{-1}$} & \textbf{IR intensity / km mol$^{-1}$} \\
\hline
1 & 1820.08 & 299.74 \\
2 & 1341.50 & 259.33 \\
3 & 1824.61 & 258.05 \\
4 & 1325.29 & 207.53 \\
5 & 1349.25 & 186.10 \\
6 & 1316.41 & 176.89 \\
7 & 1321.35 & 162.05 \\
8 & 1337.67 & 149.61 \\
9 & 1216.39 & 124.94 \\
10 & 3032.78 & 123.52 \\
11 & 740.37 & 117.44 \\
12 & 1181.12 & 115.63 \\
\end{tabular}
\end{center}

\paragraph{\textbf{4. Strongest Raman peaks}}
\begin{center}
\renewcommand{\arraystretch}{1.15}
\begin{tabular}{rrr}
\textbf{Rank} & \textbf{Frequency / cm$^{-1}$} & \textbf{Raman activity / \AA$^4$ amu$^{-1}$} \\
\hline
1 & 3237.75 & 306.42 \\
2 & 3238.29 & 256.31 \\
3 & 3031.04 & 228.94 \\
4 & 3114.09 & 154.60 \\
5 & 3044.64 & 143.03 \\
6 & 3093.74 & 131.95 \\
7 & 3085.42 & 121.72 \\
8 & 3215.16 & 103.86 \\
9 & 3091.02 & 97.72 \\
10 & 3218.00 & 93.56 \\
11 & 3224.52 & 79.07 \\
12 & 3108.65 & 74.73 \\
\end{tabular}
\end{center}

\paragraph{\textbf{5. Generated files}}
All plots and CSVs were written to:

\path{<workspace_path_redacted>/call_MyMjmpZh1b3isjyggS5cdGqv/structures/C22H26O7_18900d31/pyscf_rerank_spectra/}

Key files:

\begin{itemize}
\item \textbf{Reranking CSV:}\\
\path{pyscf_conformer_rerank.csv}
\item \textbf{Reranking plot:}\\
\path{pyscf_conformer_rerank.png}
\item \textbf{IR peak table:}\\
\path{optimized_minimum_001_ir_raman_peaks.csv}
\item \textbf{IR spectrum plot:}\\
\path{ir_spectrum_minimum_001.png}
\item \textbf{Raman spectrum plot:}\\
\path{raman_spectrum_minimum_001.png}
\item \textbf{Machine-readable summary:}\\
\path{summary.json}
\end{itemize}
}

\paragraph{\textbf{Execution summary.}}
\potente{} launched ten PySCF single-point calculations in parallel to rerank the Orb-OMOL conformers, selected \texttt{minimum\_001.xyz}, and then invoked a further PySCF workflow for geometry optimization and vibrational analysis at wB97X-D4/def2-TZVP. It subsequently executed a Python plotting script to produce the reranking, IR, and Raman plots and their tabulated data. The twelve tool calls and their return payloads are omitted from this abridged transcript.

\subsubsection{Two-dimensional enhanced sampling of alanine dipeptide}
\label{si:chat_metadynamics}
\prompt{
Run a two-dimensional enhanced-sampling calculation for alanine dipeptide using Orb-OMOL by writing a Python script. Use the supplied \path{alanine_c7ax.xyz} structure from \path{<repository_path_redacted>/adaptive_sampling/examples/JCTC_2025_opeseabf/alanine_dipeptide/test_potente/}. Use the NVT ensemble at 300 K with a 0.5 fs timestep, 1,000 equilibration steps, and 100,000 production steps. Record frames every 100 steps. Define the collective variables and other enhanced-sampling settings, and report the results.
}

\providecommand{\tightlist}{%
  \setlength{\itemsep}{0pt}\setlength{\parskip}{0pt}}

\potentereslong{
\begin{thoughtBox}
\faLightbulb\ \underline{\textbf{Thoughts}}

\textbf{Considering automatic index detection}

I'm thinking about using automatic index detection in a script with the ASE neighborlist to identify the phi/psi angles of the alanine dipeptide backbone, which would eliminate the need for prior inspection. The user has asked about defining CVs and settings, and I can set those torsions based on a chemical graph. I’ll call the Python REPL with the proper imports and use the API for constructing a \texttt{MLFFWorkflowRequest}, ensuring the schema field names match the tool. I’ll operate the script asynchronously and enhance the sampling configuration.
\end{thoughtBox}

\begin{thoughtBox}
\faLightbulb\ \underline{\textbf{Thoughts}}

\textbf{Preparing concise report}

I need to create a final report that includes important details like the script path, settings, result paths, and frames. It's essential to mention the initial psi of 302.6, which is equivalent to $-57.387$. I might consider including a summary of the minimum free energy as well. To gather this data, I could use Python to read the CSV for min/max values. It seems reasonable to utilize the \texttt{python\_repl} for a quick summary. Overall, I should ensure the report includes these essential components.
\end{thoughtBox}

\begin{thoughtBox}
\faLightbulb\ \underline{\textbf{Thoughts}}

\textbf{Deciding on features}

I need to focus on the last column samples that Bug chose. It looks like I should minimize the feature engineering (FE) aspect to keep things simple. I’m considering how to best utilize the results from these choices. It’s interesting to see how different approaches can impact the decision-making process. I’ll ensure that I’m paying attention to the outcomes as I proceed with this task!
\end{thoughtBox}

Completed the requested \textbf{two-dimensional enhanced-sampling calculation for alanine dipeptide using Orb-OMOL} via a Python script.

\paragraph{\textbf{Script executed}}
Saved and executed:

\path{<workspace_path_redacted>/run_alanine_dipeptide_orb_omol_2d_enhanced_sampling.py}

Input structure used:

\path{<repository_path_redacted>/adaptive_sampling/examples/JCTC_2025_opeseabf/alanine_dipeptide/test_potente/alanine_c7ax.xyz}

A working copy was saved as:

\path{<workspace_path_redacted>/alanine_c7ax_input.xyz}

\paragraph{\textbf{Calculation settings}}
\begin{itemize}
\tightlist
\item \textbf{System:} alanine dipeptide
\item \textbf{Formula:} \texttt{C6H12N2O2}
\item \textbf{Atoms:} 22
\item \textbf{MLFF model:} \texttt{Orb-OMOL}
\item \textbf{Charge:} 0
\item \textbf{Spin multiplicity:} 1
\item \textbf{PBC:} non-periodic molecule
\item \textbf{Cell relaxation:} disabled, \texttt{relax\_cell=False}
\item \textbf{Ensemble:} NVT, Bussi thermostat
\item \textbf{Temperature:} 300 K
\item \textbf{Timestep:} 0.5 fs
\item \textbf{Equilibration:} 1,000 steps
\item \textbf{Production:} 100,000 steps
\item \textbf{Frame/log interval:} every 100 steps
\item \textbf{Enhanced sampling method:} 2D well-tempered metadynamics
\end{itemize}

\paragraph{\textbf{Collective variables}}
The script inferred the alanine-dipeptide backbone torsions from bonding and used zero-based atom indices:

\begin{center}
\renewcommand{\arraystretch}{1.15}
\begin{adjustbox}{max width=\linewidth}
\begin{tabular}{llll}
\textbf{CV} & \textbf{Definition} & \textbf{Atom indices} & \textbf{Initial value} \\
\hline
\texttt{phi} & C acetyl--N--C$\alpha$--C carbonyl & \texttt{[4, 6, 8, 14]} & $74.314^\circ$ \\
\texttt{psi} & N--C$\alpha$--C carbonyl--N methylamide & \texttt{[6, 8, 14, 16]} & $302.613^\circ$, equivalent to $-57.387^\circ$ \\
\end{tabular}
\end{adjustbox}
\end{center}

Enhanced-sampling grid/settings:

\begin{itemize}
\tightlist
\item \texttt{phi}, \texttt{psi} range: $[-180^\circ, 180^\circ]$
\item Bin width: $5^\circ \times 5^\circ$
\item Total bins: 5184
\item Gaussian hill height: 1.0 kJ/mol
\item Hill widths: $10^\circ$ for both CVs
\item Hill deposition frequency: every 100 steps
\item Bias factor: 10.0
\item Periodic torsion bounds: $[-180^\circ, 180^\circ]$ for both CVs
\end{itemize}

\paragraph{\textbf{Results}}
The workflow completed successfully.

Main run directory:

\path{<workspace_path_redacted>/alanine_dipeptide_orb_omol_2d_es}

Key output artifacts:

\path{<workspace_path_redacted>/alanine_dipeptide_orb_omol_2d_es/structures/C6H12N2O2_f5b4c1ed/enhanced_sampling/trajectory.xyz}

\path{<workspace_path_redacted>/alanine_dipeptide_orb_omol_2d_es/structures/C6H12N2O2_f5b4c1ed/enhanced_sampling/md.log}

\path{<workspace_path_redacted>/alanine_dipeptide_orb_omol_2d_es/structures/C6H12N2O2_f5b4c1ed/enhanced_sampling/free_energy_001_metadynamics.csv}

Compact summary JSON:

\path{<workspace_path_redacted>/alanine_dipeptide_orb_omol_enhanced_sampling_summary.json}

Request/settings JSON:

\path{<workspace_path_redacted>/alanine_dipeptide_orb_omol_request_settings.json}

\paragraph{\textbf{Free-energy surface summary}}
The free-energy surface file has columns:

\texttt{phi\_degree, psi\_degree, free\_energy\_kj\_mol, samples}

Summary:

\begin{itemize}
\tightlist
\item Grid size: 5184 points
\item Visited bins: $2429 / 5184$
\item Total samples: 100001
\item Free-energy unit: kJ/mol
\item Minimum free energy: 0.0 kJ/mol
\item Minimum located at approximately:
  \begin{itemize}
  \tightlist
  \item \texttt{phi} = $-77.5^\circ$
  \item \texttt{psi} = $102.5^\circ$
  \item samples in bin: 43
  \end{itemize}
\end{itemize}

Lowest free-energy grid points:

\begin{center}
\renewcommand{\arraystretch}{1.15}
\begin{tabular}{rrrr}
\textbf{phi / $^\circ$} & \textbf{psi / $^\circ$} & \textbf{Free energy / kJ mol$^{-1}$} & \textbf{Samples} \\
\hline
$-77.5$ & 102.5 & 0.000 & 43 \\
$-82.5$ & 102.5 & 0.161 & 165 \\
$-147.5$ & 157.5 & 0.172 & 89 \\
$-82.5$ & 97.5 & 0.283 & 117 \\
$-77.5$ & 107.5 & 0.291 & 178 \\
$-147.5$ & 162.5 & 0.305 & 99 \\
$-77.5$ & 97.5 & 0.334 & 63 \\
$-147.5$ & 152.5 & 0.390 & 93 \\
$-72.5$ & 102.5 & 0.475 & 56 \\
$-72.5$ & 107.5 & 0.494 & 205 \\
\end{tabular}
\end{center}

The workflow reported \texttt{trajectory\_frames = 1001}, consistent with recording frames every 100 steps over the production trajectory.
}

\paragraph{\textbf{Execution summary.}}
\potente{} wrote and executed a Python script that inferred the alanine-dipeptide backbone torsions, constructed the Orb-OMOL enhanced-sampling request, and launched the two-dimensional metadynamics workflow. It then used two additional Python calls to inspect and summarize the resulting free-energy CSV. The three tool calls and their return payloads are omitted from this abridged transcript.

\prompt{
Generate plots of the sampling distribution and free-energy surface.
}

\potentereslong{
\begin{thoughtBox}
\faLightbulb\ \underline{\textbf{Thoughts}}

\textbf{Generating plots with Python}

I need to create plots using Python and possibly avoid web tools for this. First, I should inspect the CSV file and look at the trajectory data. I'll generate a PNG file, and I might consider creating a sampling distribution from a grid of samples along with their free energy surfaces (FES). Combining these elements could also be useful. I'll be using the Python REPL for this task!
\end{thoughtBox}

Generated the requested plots from the enhanced-sampling free-energy CSV.

\paragraph{\textbf{Plot files}}
\textbf{Sampling distribution:}

\path{<workspace_path_redacted>/alanine_dipeptide_orb_omol_2d_es_plots/sampling_distribution_phi_psi.png}

\textbf{Free-energy surface:}

\path{<workspace_path_redacted>/alanine_dipeptide_orb_omol_2d_es_plots/free_energy_surface_phi_psi.png}

\textbf{Combined side-by-side plot:}

\path{<workspace_path_redacted>/alanine_dipeptide_orb_omol_2d_es_plots/sampling_and_fes_phi_psi.png}

\paragraph{\textbf{Plot-generation script}}
Saved and executed:

\path{<workspace_path_redacted>/plot_alanine_dipeptide_enhanced_sampling.py}

\paragraph{\textbf{Source data}}
Plots were generated from:

\path{<workspace_path_redacted>/alanine_dipeptide_orb_omol_2d_es/structures/C6H12N2O2_f5b4c1ed/enhanced_sampling/free_energy_001_metadynamics.csv}

\paragraph{\textbf{Summary}}
\begin{itemize}
\tightlist
\item Total grid bins: 5184
\item Visited bins: 2429
\item Total samples: 100001
\item Free-energy minimum:
  \begin{itemize}
  \tightlist
  \item \texttt{phi} = $-77.5^\circ$
  \item \texttt{psi} = $102.5^\circ$
  \item $F = 0.0$ kJ/mol
  \item samples in minimum bin: 43
  \end{itemize}
\end{itemize}

A JSON summary of the plotting outputs was also saved here:

\path{<workspace_path_redacted>/alanine_dipeptide_orb_omol_2d_es_plots/plot_summary.json}
}

\paragraph{\textbf{Execution summary.}}
\potente{} wrote and executed a Python plotting script that read the enhanced-sampling free-energy CSV and produced separate sampling-distribution and free-energy-surface plots, a combined side-by-side figure, and a JSON plot summary. The tool call and its return payload are omitted from this abridged transcript.

\subsection{Adsorption and catalytic-reaction workflows}

\subsubsection{Adsorption-site search for \ce{H2} on Fe(100)}
\label{si:chat_adsorption_search}
\prompt{
Write a python script to find the lowest-energy adsorption configurations of H$_2$ on the sufficiently large Fe(100) slab. Report adsorption and relative energies and save all candidate geometries.
}

\providecommand{\tightlist}{%
  \setlength{\itemsep}{0pt}\setlength{\parskip}{0pt}}

\potentereslong{
\begin{thoughtBox}
\faLightbulb\ \underline{\textbf{Thoughts}}

\textbf{Considering adsorbate behavior}

I’m thinking about how the workflow treats intact adsorbates, and I wonder if MACE relaxation could allow for H--H dissociation if it’s unconstrained. During the search for adsorption sites, intact H$_2$ is placed, but I'm curious if all adsorbates could become mobile and dissociate during relaxation. I shouldn’t assume that the final state will remain intact, so it's essential to include potential paths in my considerations.
\end{thoughtBox}

Done --- I wrote and executed the Python workflow script:

\texttt{h2\_fe100\_adsorption\_search.py}

It generated a \textbf{bcc Fe(100) slab} and ran an \textbf{MLFF adsorption-site search for H$_2$} using \textbf{MACE-OC20}.

\textbf{Slab assumptions used}
\begin{itemize}
\tightlist
\item Fe phase: bcc $\alpha$-Fe
\item Lattice parameter: 2.866 \AA
\item Surface: Fe(100)
\item Slab size: \textbf{$4 \times 4$ lateral repeat}
\item Thickness: \textbf{5 layers}
\item Vacuum: \textbf{15 \AA}
\item Slab atoms: \textbf{Fe160}
\item Cell kept fixed during adsorption relaxations
\item Top 2 slab layers + H$_2$ allowed to relax
\end{itemize}

The workflow found:
\begin{itemize}
\tightlist
\item \textbf{49 raw adsorption sites}
\item \textbf{4 symmetry-reduced sites}
\item \textbf{12 relaxed site/orientation candidates}
\item \textbf{10 distinct relaxed minima retained}
\end{itemize}

\paragraph{\textbf{Lowest-energy H$_2$/Fe(100) adsorption configurations}}
\begin{center}
\renewcommand{\arraystretch}{1.15}
\begin{tabular}{rrr}
\textbf{Rank} & \textbf{Adsorption energy / eV} & \textbf{Relative energy / eV} \\
\hline
1 & $-2.5177812576$ & 0.0000000000 \\
2 & $-2.5148515701$ & 0.0029296875 \\
3 & $-2.5141191483$ & 0.0036621094 \\
4 & $-2.0657548904$ & 0.4520263672 \\
5 & $-1.8260087967$ & 0.6917724609 \\
6 & $-1.8163652420$ & 0.7014160156 \\
7 & $-1.7849931717$ & 0.7327880859 \\
8 & $-1.6230058670$ & 0.8947753906 \\
9 & $-1.4433183670$ & 1.0744628906 \\
10 & $-1.1463212967$ & 1.3714599609 \\
\end{tabular}
\end{center}

The best adsorption energy reported by the workflow, \textbf{$-2.5177812576$ eV}, matches the minimum candidate adsorption energy.

\paragraph{\textbf{Saved outputs}}
Compact energy report:

\path{<workspace_path_redacted>/h2_fe100_adsorption_outputs/h2_fe100_candidate_energy_report.csv}

Adsorption result JSON:

\path{<workspace_path_redacted>/h2_fe100_adsorption_outputs/adsorption_result.json}

Raw MLFF output JSON:

\path{<workspace_path_redacted>/h2_fe100_adsorption_outputs/raw_mlff_outputs.json}

Input slab CIF:

\path{<workspace_path_redacted>/h2_fe100_adsorption_outputs/Fe100_bcc_4x4_5layer_vac15.cif}

\paragraph{\textbf{Candidate relaxed geometries}}
All retained relaxed minima were saved as \texttt{.extxyz} files. The lowest-energy one is:

\path{<workspace_path_redacted>/h2_fe100_adsorption_search/structures/Fe160_d6801a36/adsorption_site_search/H__H/relaxed_minima/minimum_01.extxyz}

The corresponding relaxation trajectory is:

\path{<workspace_path_redacted>/h2_fe100_adsorption_search/structures/Fe160_d6801a36/adsorption_site_search/H__H/relaxation_trajectories/site_candidate_01.extxyz}

All other retained minima are in the same \path{relaxed_minima/} directory as \path{minimum_02.extxyz} through \path{minimum_10.extxyz}.
}

\paragraph{\textbf{Execution summary.}}
\potente{} wrote and executed a Python workflow that constructed and validated a 160-atom bcc Fe(100) slab, launched a MACE-OC20 adsorption-site search for H$_2$, ranked the relaxed candidates, and saved the structures, trajectories, raw results, and compact energy report. The tool call and its return payload are omitted from this abridged transcript.

\subsubsection{Transition-state search on Fe(110)}
\label{si:chat_transition_state_search}
\prompt{
Write a python script to perform transition state search for the structures from \path{<workspace_path_redacted>/}. Fix the bottom two layers and compare the results using MACE-OMAT and MACE-OC20. Generate a plot of the energy profile along the optimized transition-state pathway.
}

\providecommand{\tightlist}{%
  \setlength{\itemsep}{0pt}\setlength{\parskip}{0pt}}

\potentereslong{
\begin{thoughtBox}
\faLightbulb\ \underline{\textbf{Thoughts}}

\textbf{Evaluating Python script for structure search}

I need to respond using a Python script that performs a transition state search for structures in a given directory. First, I have to check the files and generate a script that discovers these structures. I need to ensure that I maintain the same order of atoms for two endpoints. It seems like I might want to inspect the directory initially. Additionally, I should consider how the script can robustly find ASE-readable structures while handling constraints in the workflow.
\end{thoughtBox}

\begin{thoughtBox}
\faLightbulb\ \underline{\textbf{Thoughts}}

\textbf{Creating workflow script for energy profiling}

I need to use the \texttt{execute\_mlff\_workflow} function with \texttt{MLFFWorkflowRequest} for two attempts. The goal is to compare MACE-OMAT and MACE-OC20 and plot the energy profile along the optimized pathway. The script will run two workflow requests, ensuring the endpoint has exactly two structures. If there are more than two files in the directory, I can use their names to select likely reactants or products. I'll implement the script to handle constrained endpoints, detect pairs automatically, and manage potentially long computations.
\end{thoughtBox}

\begin{thoughtBox}
\faLightbulb\ \underline{\textbf{Thoughts}}

\textbf{Identifying bottom layers in structure}

I’m contemplating how to identify the bottom two layers in my structure. It seems like clustering by unique z-values could work for separating reactants and products. If I have the same slab with adsorbates, I might need to fix atoms in the bottom two surface layers. However, it's essential to note that the adsorbate might not actually be within those layers. I should also consider identifying the bottom layers based on their elements.
\end{thoughtBox}

Done --- I wrote and executed the Python script.

\textbf{Script saved at:}

\path{<workspace_path_redacted>/transition_state_compare_mace.py}

It discovered:

\path{reactant.xyz}\par
\path{product.xyz}

and generated constrained endpoint files with the \textbf{bottom two z-layers fixed}:

\path{<workspace_path_redacted>/ts_mace_omat_vs_oc20/reactant_bottom_two_layers_fixed.extxyz}

\path{<workspace_path_redacted>/ts_mace_omat_vs_oc20/product_bottom_two_layers_fixed.extxyz}

Assumptions used:

\begin{itemize}
\tightlist
\item Slab normal: Cartesian \textbf{z} direction.
\item Bottom two layers identified by z-coordinate clustering.
\item Fixed atoms: \textbf{32 Fe atoms}, indices \texttt{0--31}.
\item Cell was kept fixed: \texttt{relax\_cell=False}.
\item NEB/TS pathway used \textbf{7 images}.
\item Initial refinement method: \texttt{sella}.
\item Models compared: \textbf{MACE-OMAT} and \textbf{MACE-OC20}.
\end{itemize}

\paragraph{\textbf{Results}}
\textbf{MACE-OMAT}

Energy profile relative to reactant, in eV:

\texttt{[0.000, 0.186, 0.899, 1.789, 0.659, -1.987, -2.278]}

Barrier:

\texttt{1.789 eV}

Reaction energy:

\texttt{-2.278 eV}

Pathway file:

\path{<workspace_path_redacted>/ts_mace_omat/structures/C6H7Fe64O_6145b6ac/transition_state_search/transition_state_search.extxyz}

\textbf{MACE-OC20}

Energy profile relative to reactant, in eV:

\texttt{[0.000, 0.118, 0.367, 0.607, 1.095, 0.539, 0.528]}

Barrier:

\texttt{1.095 eV}

Reaction energy:

\texttt{+0.528 eV}

Pathway file:

\path{<workspace_path_redacted>/ts_mace_oc20/structures/C6H7Fe64O_e5863526/transition_state_search/transition_state_search.extxyz}

\paragraph{\textbf{Plot generated}}
The comparison plot was saved here:

\path{<workspace_path_redacted>/ts_mace_omat_vs_oc20/mace_omat_vs_oc20_ts_energy_profile.png}

Additional summary files:

\path{<workspace_path_redacted>/ts_mace_omat_vs_oc20/energy_profiles.json}

\path{<workspace_path_redacted>/ts_mace_omat_vs_oc20/workflow_results_summary.json}

\path{<workspace_path_redacted>/ts_mace_omat_vs_oc20/ts_search_assumptions.json}

In this run, \textbf{MACE-OC20 predicted a lower barrier} than \textbf{MACE-OMAT}: about \textbf{1.095 eV vs. 1.789 eV}.
}

\paragraph{\textbf{Execution summary.}}
\potente{} wrote and executed a Python script that located the reactant and product structures, identified and fixed the bottom two Fe layers, ran separate seven-image transition-state searches with MACE-OMAT and MACE-OC20, and plotted the two optimized energy profiles. The tool call and its return payload are omitted from this abridged transcript.

\subsection{Widom insertion for metal-organic frameworks}
\label{si_subsec:chat_widom_insertion}

\subsubsection{CO$_2$ Widom insertion}
\prompt{
Could you please perform Widom insertion calculations for the MOF structures located in \path{<repository_path_redacted>/potente/paper/cifs/} using CO$_2$ molecules, with 10,000 insertion steps at 300 K?
}

\providecommand{\tightlist}{%
  \setlength{\itemsep}{0pt}\setlength{\parskip}{0pt}}

\potentereslong{
\begin{thoughtBox}
\faLightbulb\ \underline{\textbf{Thoughts}}

\textbf{Considering response clarity}

I need to respond concisely and include a caveat regarding the 10,000 limit. The tool doesn't show steps for confirmation, which might feel a bit unclear. It only returns the formula, not filenames. I could mention artifacts and suggest inspecting the summary CSV to check for filenames. If the user requested specific structures, let's review the summary CSV, as it may include input names. I wonder if I should use bash cat for that?
\end{thoughtBox}

Done --- I ran CO$_2$ Widom insertion calculations for all 5 MOF CIF structures found in:

\path{<repository_path_redacted>/potente/paper/cifs/}

Settings used:
\begin{itemize}
\tightlist
\item \textbf{Adsorbate:} CO$_2$
\item \textbf{Widom insertions:} 10,000
\item \textbf{Temperature:} 300 K
\item \textbf{Workflow:} structure relaxation followed by Widom insertion
\item \textbf{MLFF model:} MACE-DAC
\end{itemize}

\begin{center}
\begingroup
\scriptsize
\renewcommand{\arraystretch}{1.15}
\begin{adjustbox}{max width=\linewidth}
\begin{tabular}{llrrr}
\textbf{Structure} & \textbf{Formula} & \textbf{Henry coefficient} & \textbf{Avg. interaction energy} & \textbf{Heat of adsorption} \\
\hline
ARAHIM\_clean & C$_{32}$H$_{32}$F$_{20}$N$_{16}$Nb$_4$Ni$_4$O$_4$ & 0.00465891 mol/kg Pa & $-0.437736$ eV & 44.7294 kJ/mol \\
GAVYAG\_clean & C$_{52}$H$_{32}$N$_{20}$O$_{16}$Zn$_4$ & 0.883432 mol/kg Pa & $-0.580164$ eV & 58.4716 kJ/mol \\
RIPPEN\_clean & C$_{252}$H$_{180}$N$_{72}$Zn$_{18}$ & 0.377877 mol/kg Pa & $-0.519659$ eV & 52.6338 kJ/mol \\
GELVID01\_clean & C$_{42}$H$_{24}$Ni$_6$O$_{30}$ & 1.808970 mol/kg Pa & $-0.631131$ eV & 63.3893 kJ/mol \\
PEGBUA\_clean & C$_{56}$H$_{64}$N$_{48}$Zn$_8$ & 17.820547 mol/kg Pa & $-0.626451$ eV & 62.9377 kJ/mol \\
\end{tabular}
\end{adjustbox}
\endgroup
\end{center}

The highest Henry coefficient among these was for \textbf{PEGBUA\_clean}, with \textbf{17.8205 mol/kg Pa}.

Outputs were written here:

\path{<workspace_path_redacted>/call_BX2wfG4qybWwCndU7WHXTQ1T/}

Summary files:
\begin{itemize}
\item Run summary:\\
\path{<workspace_path_redacted>/call_BX2wfG4qybWwCndU7WHXTQ1T/run_summary.txt}
\item CSV summary:\\
\path{<workspace_path_redacted>/call_BX2wfG4qybWwCndU7WHXTQ1T/structures/property_summary.csv}
\end{itemize}
}

\paragraph{\textbf{Execution summary.}}
\potente{} invoked one MACE-DAC workflow over the five CIF structures, relaxed each framework, performed CO$_2$ Widom insertion at 300 K, and inspected the resulting run and CSV summaries before reporting the adsorption descriptors. The workflow identifier \texttt{MACE-DAC} corresponds to the MACE-DAC-1 model named in the manuscript. The Widom routine used its default two-fold protocol, performing 10,000 trial insertions per fold (20,000 total attempts per MOF--adsorbate pair); \potente{} received and reported the statistics aggregated over the two folds. The two tool calls and their return payloads are omitted from this abridged transcript.

\subsubsection{H$_2$O Widom insertion}
\prompt{
Could you please perform Widom insertion calculations for the MOF structures located in \path{<repository_path_redacted>/potente/paper/cifs/} using H$_2$O molecules, with 10,000 insertion steps at 300 K?
}

\providecommand{\tightlist}{%
  \setlength{\itemsep}{0pt}\setlength{\parskip}{0pt}}

\potentereslong{
\begin{thoughtBox}
\faLightbulb\ \underline{\textbf{Thoughts}}

\textbf{Summarizing results table}

I need to have a final concise summary that includes the results table paths, confirming that everything is completed. I'm a bit concerned because the tool result doesn't clarify if it really reached 10,000. The query had that info, but I can't see the steps. About the units --- are they henry coefficient units, like mol/kg/Pa? I should avoid mentioning units unless they're returned. Also, it seems like heat values in kJ/mol correspond to 62. I'll state these as reported by the workflow and include the output paths.
\end{thoughtBox}

Completed the \textbf{H$_2$O Widom insertion calculations} for the MOF CIF structures in:

\path{<repository_path_redacted>/potente/paper/cifs/}

Settings used:
\begin{itemize}
\tightlist
\item \textbf{Adsorbate:} H$_2$O
\item \textbf{Temperature:} 300 K
\item \textbf{Insertion steps requested:} 10,000
\item \textbf{MLFF model:} MACE-DAC
\item \textbf{Structure handling:} CIFs loaded from directory, structures relaxed before Widom insertion
\end{itemize}

Run artifacts:
\begin{itemize}
\item \textbf{Run directory:} \path{<workspace_path_redacted>/call_esKijtjfmvDw8YhS2hMRkmyB}
\item \textbf{Run summary:} \path{<workspace_path_redacted>/call_esKijtjfmvDw8YhS2hMRkmyB/run_summary.txt}
\item \textbf{CSV summary:} \path{<workspace_path_redacted>/call_esKijtjfmvDw8YhS2hMRkmyB/structures/property_summary.csv}
\end{itemize}

\begin{center}
\begingroup
\scriptsize
\renewcommand{\arraystretch}{1.15}
\begin{adjustbox}{max width=\linewidth}
\begin{tabular}{lrrr}
\textbf{MOF formula} & \textbf{Henry coefficient} & \textbf{Avg. interaction energy} & \textbf{Heat of adsorption} \\
\hline
C$_{42}$H$_{24}$Ni$_6$O$_{30}$ & 1.659492 & $-0.618802$ & 62.199618 \\
C$_{52}$H$_{32}$N$_{20}$O$_{16}$Zn$_4$ & 0.0038395 & $-0.403087$ & 41.386351 \\
C$_{32}$H$_{32}$F$_{20}$N$_{16}$Nb$_4$Ni$_4$O$_4$ & 0.0109601 & $-0.457209$ & 46.608320 \\
C$_{252}$H$_{180}$N$_{72}$Zn$_{18}$ & 0.000322884 & $-0.294943$ & 30.951988 \\
C$_{56}$H$_{64}$N$_{48}$Zn$_8$ & 0.0533368 & $-0.493991$ & 50.157254 \\
\end{tabular}
\end{adjustbox}
\endgroup
\end{center}

Per-structure Widom trajectory files were also written under each structure’s \path{widom_insertion/H2O/} subdirectory in the run artifact folder.
}

\paragraph{\textbf{Execution summary.}}
\potente{} invoked one MACE-DAC workflow over the five CIF structures, relaxed each framework, and performed H$_2$O Widom insertion at 300 K before reporting the adsorption descriptors and saved artifacts. The workflow identifier \texttt{MACE-DAC} corresponds to the MACE-DAC-1 model named in the manuscript. The Widom routine used its default two-fold protocol, performing 10,000 trial insertions per fold (20,000 total attempts per MOF--adsorbate pair); \potente{} received and reported the statistics aggregated over the two folds. The tool call and its return payload are omitted from this abridged transcript.

\subsection{Agentic benchmarking of machine-learned interatomic potentials}

\subsubsection{Elasticity benchmark}
\label{si:chat_elasticity_benchmark}

\prompt{
Please write a python script to benchmark the performance of the following MLIP models MatterSim/Orb-OMAT/MACE-OMAT available in our MLFF workflow on the bulk modulus and shear modulus for each structure from \path{<repository_path_redacted>/potente/User_Cases_El_Agente_Potente/User_Cases/High_Throughput/Elasticity/sample_1000/cifs/}. Use the corresponding file name as the mp-id for each entry. Save the final results for each model to a CSV file containing at least the following columns: mp\_id, bulk\_modulus, and shear\_modulus. Generate a plot in your working space directory to compare the MLIP calculated results against the ground truth in \path{elasticity_1000.csv} from \path{<repository_path_redacted>/potente/User_Cases_El_Agente_Potente/User_Cases/High_Throughput/Elasticity/sample_1000/}.
}

\providecommand{\tightlist}{%
  \setlength{\itemsep}{0pt}\setlength{\parskip}{0pt}}

\potentereslong{
\begin{thoughtBox}
\faLightbulb\ \underline{\textbf{Thoughts}}

\textbf{Planning plot generation}

I'm considering the ground truth CSV and its columns like mp\_id, bulk\_modulus, shear\_modulus, and possibly some elasticity values like \texttt{K\_VRH} or \texttt{G\_VRH}. I need to implement a way to detect these columns. The user asked me to generate a plot in the working space directory, so I’ll use a scatter plot with matplotlib for bulk and shear against the ground truth. Additionally, I might create a combined summary CSV for each model and perhaps use joblib to assist in this process.
\end{thoughtBox}

Below is a standalone Python script you can save as, for example:

\path{benchmark_mlff_elasticity.py}

It will:

\begin{itemize}
\tightlist
\item Read all \texttt{.cif} files from the requested directory.
\item Treat each CIF filename stem as the \texttt{mp\_id}.
\item Run the MLFF workflow for each CIF with each model:
  \begin{itemize}
  \tightlist
  \item \texttt{MatterSim}
  \item \texttt{Orb-OMAT}
  \item \texttt{MACE-OMAT}
  \end{itemize}
\item Request elastic properties through the workflow router.
\item Extract \texttt{bulk\_modulus} and \texttt{shear\_modulus} from the workflow output.
\item Save one CSV per model.
\item Load the ground-truth file \path{elasticity_1000.csv}.
\item Generate parity plots comparing MLIP predictions against ground truth.
\end{itemize}

\begin{lstlisting}[breaklines=true,breakatwhitespace=false,basicstyle=\ttfamily\scriptsize,columns=fullflexible,keepspaces=true,showstringspaces=false,literate={á}{{\'a}}1 {²}{{$^2$}}1]
#!/usr/bin/env python3

"""
Benchmark MLIP elastic properties using the El Agente MLFF workflow.

Models benchmarked:
    - MatterSim
    - Orb-OMAT
    - MACE-OMAT

For each CIF in:
    <repository_path_redacted>/potente/User_Cases_El_Agente_Potente/User_Cases/High_Throughput/Elasticity/sample_1000/cifs/

the script:
    1. Uses the CIF filename stem as mp_id.
    2. Runs the MLFF workflow with elasticity requested.
    3. Extracts bulk_modulus and shear_modulus.
    4. Saves one CSV per model.
    5. Compares predictions against elasticity_1000.csv.
    6. Saves parity plots in the working directory.

Usage:
    python benchmark_mlff_elasticity.py

Optional:
    python benchmark_mlff_elasticity.py --max-structures 10
    python benchmark_mlff_elasticity.py --output-dir ./mlff_elasticity_benchmark
"""

from __future__ import annotations

import argparse
import csv
import json
import math
import os
import sys
import traceback
from dataclasses import asdict, is_dataclass
from pathlib import Path
from typing import Any, Dict, Iterable, List, Optional, Tuple

import pandas as pd


# -------------------------------------------------------------------------
# Try to import the MLFF workflow.
#
# Depending on the local El Agente / Gráfico installation, the import path may
# differ slightly. Add or modify paths here if your repository exposes the
# workflow from a different module.
# -------------------------------------------------------------------------

def import_run_mlff_workflow():
    candidates = [
        "domains.mlff.mlff_workflow_tools",
        "domains.mlff.tools.mlff_workflow_tools",
        "domains.mlff.mlff_tools",
        "domains.mlff.tools",
        "domains.materials.mlff_workflow_tools",
    ]

    last_error = None

    for module_name in candidates:
        try:
            module = __import__(module_name, fromlist=["run_mlff_workflow"])
            return getattr(module, "run_mlff_workflow")
        except Exception as exc:
            last_error = exc

    raise ImportError(
        "Could not import run_mlff_workflow from known candidate modules. "
        "Please update import_run_mlff_workflow() with the correct local import path."
    ) from last_error


# -------------------------------------------------------------------------
# Utilities
# -------------------------------------------------------------------------

def to_plain_object(obj: Any) -> Any:
    """
    Convert dataclasses / pydantic models / arbitrary objects into plain Python
    dict/list/scalar containers where possible.
    """
    if obj is None:
        return None

    if isinstance(obj, (str, int, float, bool)):
        return obj

    if isinstance(obj, Path):
        return str(obj)

    if is_dataclass(obj):
        return to_plain_object(asdict(obj))

    if isinstance(obj, dict):
        return {str(k): to_plain_object(v) for k, v in obj.items()}

    if isinstance(obj, (list, tuple, set)):
        return [to_plain_object(x) for x in obj]

    if hasattr(obj, "model_dump"):
        try:
            return to_plain_object(obj.model_dump())
        except Exception:
            pass

    if hasattr(obj, "dict"):
        try:
            return to_plain_object(obj.dict())
        except Exception:
            pass

    if hasattr(obj, "__dict__"):
        try:
            return to_plain_object(vars(obj))
        except Exception:
            pass

    return obj


def recursive_find_numeric(
    obj: Any,
    candidate_keys: Iterable[str],
) -> Optional[float]:
    """
    Recursively search nested dict/list structures for the first numeric value
    whose key matches one of candidate_keys.

    This is intentionally robust because the exact MLFFOutput schema may differ
    between workflow versions.
    """
    candidate_keys_norm = {k.lower() for k in candidate_keys}

    def key_matches(k: str) -> bool:
        kk = k.lower()
        return kk in candidate_keys_norm

    def as_float(x: Any) -> Optional[float]:
        try:
            if x is None:
                return None
            val = float(x)
            if math.isfinite(val):
                return val
        except Exception:
            return None
        return None

    if isinstance(obj, dict):
        # First pass: direct key matches
        for k, v in obj.items():
            if key_matches(str(k)):
                val = as_float(v)
                if val is not None:
                    return val

        # Second pass: recursive search
        for v in obj.values():
            found = recursive_find_numeric(v, candidate_keys_norm)
            if found is not None:
                return found

    elif isinstance(obj, list):
        for item in obj:
            found = recursive_find_numeric(item, candidate_keys_norm)
            if found is not None:
                return found

    return None


def extract_elastic_moduli(workflow_output: Any) -> Tuple[Optional[float], Optional[float]]:
    """
    Extract bulk and shear modulus from MLFF workflow output.

    The script checks several common field names used for elastic properties.
    Modify candidate key lists if your workflow emits different names.
    """
    plain = to_plain_object(workflow_output)

    bulk_keys = [
        "bulk_modulus",
        "bulk_modulus_vrh",
        "bulk_modulus_voigt_reuss_hill",
        "k_vrh",
        "kv_rh",
        "K_VRH",
        "K_vrh",
        "vrh_bulk_modulus",
    ]

    shear_keys = [
        "shear_modulus",
        "shear_modulus_vrh",
        "shear_modulus_voigt_reuss_hill",
        "g_vrh",
        "gv_rh",
        "G_VRH",
        "G_vrh",
        "vrh_shear_modulus",
    ]

    bulk = recursive_find_numeric(plain, bulk_keys)
    shear = recursive_find_numeric(plain, shear_keys)

    return bulk, shear


def detect_ground_truth_columns(df: pd.DataFrame) -> Tuple[str, str, str]:
    """
    Detect mp-id, bulk modulus, and shear modulus columns from ground-truth CSV.
    """
    cols = list(df.columns)
    lower_to_original = {c.lower(): c for c in cols}

    mp_candidates = [
        "mp_id",
        "mp-id",
        "material_id",
        "material_id_mp",
        "id",
        "filename",
    ]

    bulk_candidates = [
        "bulk_modulus",
        "bulk_modulus_vrh",
        "k_vrh",
        "K_VRH".lower(),
        "vrh_bulk_modulus",
    ]

    shear_candidates = [
        "shear_modulus",
        "shear_modulus_vrh",
        "g_vrh",
        "G_VRH".lower(),
        "vrh_shear_modulus",
    ]

    def find_col(candidates: List[str], label: str) -> str:
        for cand in candidates:
            if cand.lower() in lower_to_original:
                return lower_to_original[cand.lower()]

        raise ValueError(
            f"Could not detect {label} column in ground-truth CSV. "
            f"Available columns are: {cols}"
        )

    mp_col = find_col(mp_candidates, "mp_id")
    bulk_col = find_col(bulk_candidates, "bulk modulus")
    shear_col = find_col(shear_candidates, "shear modulus")

    return mp_col, bulk_col, shear_col


def normalize_mp_id_from_filename(path: Path) -> str:
    """
    Use the corresponding CIF filename as mp-id.

    Example:
        /path/to/mp-1234.cif -> mp-1234
    """
    return path.stem


def maybe_strip_cif_suffix(x: Any) -> str:
    """
    Normalize IDs from CSV or filenames for joining.
    """
    s = str(x).strip()
    if s.endswith(".cif"):
        s = s[:-4]
    return s


# -------------------------------------------------------------------------
# Workflow execution
# -------------------------------------------------------------------------

def run_single_elasticity_calculation(
    run_mlff_workflow,
    cif_path: Path,
    model_name: str,
    relax_fmax: float,
    relax_steps: int,
    prefer_gpu: bool,
) -> Dict[str, Any]:
    """
    Run one MLFF workflow calculation for one CIF and one model.
    """
    mp_id = normalize_mp_id_from_filename(cif_path)
    cif_text = cif_path.read_text()

    result_row: Dict[str, Any] = {
        "mp_id": mp_id,
        "cif_file": str(cif_path),
        "model_name": model_name,
        "bulk_modulus": None,
        "shear_modulus": None,
        "status": "failed",
        "error": None,
        "run_artifact_dir": None,
        "output_json_path": None,
        "summary_csv_path": None,
    }

    try:
        workflow_result = run_mlff_workflow(
            summarised_user_query=(
                "Benchmark elasticity for this bulk crystal structure. "
                "Relax the structure and calculate elastic properties, including "
                "bulk modulus and shear modulus."
            ),
            identifier_type="cif",
            identifier=cif_text,
            model_name=model_name,
            relax_cell=True,
            relax_fmax=relax_fmax,
            relax_steps=relax_steps,
            prefer_gpu=prefer_gpu,
        )

        plain = to_plain_object(workflow_result)

        # MLFF workflow usually returns a list[MLFFOutput].
        # For a single CIF, use the first output.
        if isinstance(plain, list) and len(plain) > 0:
            first_output = plain[0]
        else:
            first_output = plain

        bulk, shear = extract_elastic_moduli(first_output)

        result_row["bulk_modulus"] = bulk
        result_row["shear_modulus"] = shear
        result_row["status"] = "success" if bulk is not None and shear is not None else "missing_moduli"

        if isinstance(first_output, dict):
            result_row["run_artifact_dir"] = first_output.get("run_artifact_dir")
            result_row["output_json_path"] = first_output.get("output_json_path")
            result_row["summary_csv_path"] = first_output.get("summary_csv_path")

    except Exception as exc:
        result_row["error"] = "".join(
            traceback.format_exception_only(type(exc), exc)
        ).strip()

    return result_row


def save_rows_csv(rows: List[Dict[str, Any]], csv_path: Path) -> None:
    csv_path.parent.mkdir(parents=True, exist_ok=True)

    fieldnames = [
        "mp_id",
        "cif_file",
        "model_name",
        "bulk_modulus",
        "shear_modulus",
        "status",
        "error",
        "run_artifact_dir",
        "output_json_path",
        "summary_csv_path",
    ]

    with csv_path.open("w", newline="") as f:
        writer = csv.DictWriter(f, fieldnames=fieldnames)
        writer.writeheader()
        for row in rows:
            writer.writerow({k: row.get(k) for k in fieldnames})


# -------------------------------------------------------------------------
# Plotting
# -------------------------------------------------------------------------

def make_parity_plots(
    output_dir: Path,
    model_csv_paths: Dict[str, Path],
    ground_truth_csv: Path,
) -> None:
    import matplotlib.pyplot as plt
    import numpy as np

    gt = pd.read_csv(ground_truth_csv)
    gt_mp_col, gt_bulk_col, gt_shear_col = detect_ground_truth_columns(gt)

    gt = gt.copy()
    gt["_mp_id_norm"] = gt[gt_mp_col].map(maybe_strip_cif_suffix)

    n_models = len(model_csv_paths)

    fig, axes = plt.subplots(
        nrows=2,
        ncols=n_models,
        figsize=(6 * n_models, 11),
        squeeze=False,
    )

    summary_rows = []

    for col_idx, (model_name, pred_csv) in enumerate(model_csv_paths.items()):
        pred = pd.read_csv(pred_csv)
        pred = pred.copy()
        pred["_mp_id_norm"] = pred["mp_id"].map(maybe_strip_cif_suffix)

        merged = pred.merge(
            gt[["_mp_id_norm", gt_bulk_col, gt_shear_col]],
            on="_mp_id_norm",
            how="inner",
        )

        merged_success = merged[
            merged["bulk_modulus"].notna()
            & merged["shear_modulus"].notna()
            & merged[gt_bulk_col].notna()
            & merged[gt_shear_col].notna()
        ].copy()

        merged_out = output_dir / f"{model_name}_merged_with_ground_truth.csv"
        merged_success.to_csv(merged_out, index=False)

        for row_idx, prop in enumerate(["bulk", "shear"]):
            ax = axes[row_idx][col_idx]

            if prop == "bulk":
                x = merged_success[gt_bulk_col].astype(float)
                y = merged_success["bulk_modulus"].astype(float)
                xlabel = "Ground truth bulk modulus"
                ylabel = f"{model_name} bulk modulus"
                title = f"{model_name}: bulk modulus"
            else:
                x = merged_success[gt_shear_col].astype(float)
                y = merged_success["shear_modulus"].astype(float)
                xlabel = "Ground truth shear modulus"
                ylabel = f"{model_name} shear modulus"
                title = f"{model_name}: shear modulus"

            if len(x) == 0:
                ax.set_title(title + "\nNo matched successful data")
                ax.set_xlabel(xlabel)
                ax.set_ylabel(ylabel)
                continue

            mae = float(np.mean(np.abs(y - x)))
            rmse = float(np.sqrt(np.mean((y - x) ** 2)))

            if len(x) > 1:
                r = float(np.corrcoef(x, y)[0, 1])
                r2 = r * r
            else:
                r2 = float("nan")

            min_val = float(min(x.min(), y.min()))
            max_val = float(max(x.max(), y.max()))
            pad = 0.05 * (max_val - min_val) if max_val > min_val else 1.0

            ax.scatter(x, y, s=18, alpha=0.75)
            ax.plot(
                [min_val - pad, max_val + pad],
                [min_val - pad, max_val + pad],
                linestyle="--",
                linewidth=1.5,
            )
            ax.set_xlim(min_val - pad, max_val + pad)
            ax.set_ylim(min_val - pad, max_val + pad)
            ax.set_xlabel(xlabel)
            ax.set_ylabel(ylabel)
            ax.set_title(
                f"{title}\n"
                f"N={len(x)}, MAE={mae:.3f}, RMSE={rmse:.3f}, R²={r2:.3f}"
            )
            ax.grid(True, alpha=0.3)

            summary_rows.append(
                {
                    "model_name": model_name,
                    "property": prop,
                    "n": len(x),
                    "mae": mae,
                    "rmse": rmse,
                    "r2": r2,
                }
            )

    fig.tight_layout()

    plot_path = output_dir / "mlff_elasticity_parity_plots.png"
    fig.savefig(plot_path, dpi=300)
    plt.close(fig)

    summary_df = pd.DataFrame(summary_rows)
    summary_df.to_csv(output_dir / "mlff_elasticity_benchmark_metrics.csv", index=False)

    print(f"Saved parity plot to: {plot_path}")
    print(f"Saved metrics to: {output_dir / 'mlff_elasticity_benchmark_metrics.csv'}")


# -------------------------------------------------------------------------
# Main
# -------------------------------------------------------------------------

def main() -> None:
    parser = argparse.ArgumentParser()

    parser.add_argument(
        "--cif-dir",
        type=Path,
        default=Path(
            "<repository_path_redacted>/potente/User_Cases_El_Agente_Potente/"
            "User_Cases/High_Throughput/Elasticity/sample_1000/cifs/"
        ),
    )

    parser.add_argument(
        "--ground-truth-csv",
        type=Path,
        default=Path(
            "<repository_path_redacted>/potente/User_Cases_El_Agente_Potente/"
            "User_Cases/High_Throughput/Elasticity/sample_1000/elasticity_1000.csv"
        ),
    )

    parser.add_argument(
        "--output-dir",
        type=Path,
        default=Path.cwd() / "mlff_elasticity_benchmark",
    )

    parser.add_argument(
        "--models",
        nargs="+",
        default=["MatterSim", "Orb-OMAT", "MACE-OMAT"],
        help="MLFF model names to benchmark.",
    )

    parser.add_argument(
        "--max-structures",
        type=int,
        default=None,
        help="Optional cap on number of CIFs for testing.",
    )

    parser.add_argument(
        "--relax-fmax",
        type=float,
        default=0.1,
        help="Relaxation force convergence threshold in eV/Angstrom.",
    )

    parser.add_argument(
        "--relax-steps",
        type=int,
        default=1000,
        help="Maximum geometry relaxation steps.",
    )

    parser.add_argument(
        "--no-gpu",
        action="store_true",
        help="Disable GPU preference.",
    )

    args = parser.parse_args()

    cif_dir: Path = args.cif_dir
    ground_truth_csv: Path = args.ground_truth_csv
    output_dir: Path = args.output_dir
    output_dir.mkdir(parents=True, exist_ok=True)

    if not cif_dir.exists():
        raise FileNotFoundError(f"CIF directory does not exist: {cif_dir}")

    if not ground_truth_csv.exists():
        raise FileNotFoundError(f"Ground-truth CSV does not exist: {ground_truth_csv}")

    cif_files = sorted(cif_dir.glob("*.cif"))

    if args.max_structures is not None:
        cif_files = cif_files[: args.max_structures]

    if len(cif_files) == 0:
        raise RuntimeError(f"No CIF files found in {cif_dir}")

    print(f"Found {len(cif_files)} CIF files.")
    print(f"Output directory: {output_dir}")

    run_mlff_workflow = import_run_mlff_workflow()

    model_csv_paths: Dict[str, Path] = {}

    for model_name in args.models:
        print("=" * 80)
        print(f"Running model: {model_name}")
        print("=" * 80)

        rows: List[Dict[str, Any]] = []

        model_output_csv = output_dir / f"{model_name}_elasticity_results.csv"
        model_csv_paths[model_name] = model_output_csv

        for i, cif_path in enumerate(cif_files, start=1):
            mp_id = normalize_mp_id_from_filename(cif_path)
            print(f"[{model_name}] [{i}/{len(cif_files)}] {mp_id}")

            row = run_single_elasticity_calculation(
                run_mlff_workflow=run_mlff_workflow,
                cif_path=cif_path,
                model_name=model_name,
                relax_fmax=args.relax_fmax,
                relax_steps=args.relax_steps,
                prefer_gpu=not args.no_gpu,
            )

            rows.append(row)

            # Save incrementally so progress is not lost if the run is interrupted.
            save_rows_csv(rows, model_output_csv)

            print(
                f"    status={row['status']}, "
                f"bulk={row['bulk_modulus']}, "
                f"shear={row['shear_modulus']}"
            )

            if row["error"]:
                print(f"    error={row['error']}")

        print(f"Saved model results to: {model_output_csv}")

    make_parity_plots(
        output_dir=output_dir,
        model_csv_paths=model_csv_paths,
        ground_truth_csv=ground_truth_csv,
    )


if __name__ == "__main__":
    main()
\end{lstlisting}

Run it with:

\texttt{python benchmark\_mlff\_elasticity.py}

For a small test first:

\texttt{python benchmark\_mlff\_elasticity.py --max-structures 5}

The output directory will contain files such as:

\path{mlff_elasticity_benchmark/}\par
\texttt{|-- MatterSim\_elasticity\_results.csv}\par
\texttt{|-- Orb-OMAT\_elasticity\_results.csv}\par
\texttt{|-- MACE-OMAT\_elasticity\_results.csv}\par
\texttt{|-- MatterSim\_merged\_with\_ground\_truth.csv}\par
\texttt{|-- Orb-OMAT\_merged\_with\_ground\_truth.csv}\par
\texttt{|-- MACE-OMAT\_merged\_with\_ground\_truth.csv}\par
\texttt{|-- mlff\_elasticity\_benchmark\_metrics.csv}\par
\texttt{`-- mlff\_elasticity\_parity\_plots.png}
}

\paragraph{\textbf{Execution summary.}}
\potente{} inspected the available environment and then supplied a standalone benchmark script that routed each CIF through the general MLFF workflow. The two shell-inspection calls and their return payloads are omitted from this abridged transcript.

\prompt{
Can you just use the functions available for loading MLFF models and elasticity calculations, and then write a python script to do the benchmark?
}

\potentereslong{
\begin{thoughtBox}
\faLightbulb\ \underline{\textbf{Thoughts}}

\textbf{Inspecting environment and tools}

I need to respond with a script. First, I should inspect the environment since the user asked about using functions to load MLFF models and perform elasticity calculations. It seems I might need to check the repository. Earlier, I noticed the \path{/app} was missing, so I should locate the necessary functions. The user requested a script, not just to run a tool. I think I should look into Python imports and inspect the filesystem for more details.
\end{thoughtBox}

\begin{thoughtBox}
\faLightbulb\ \underline{\textbf{Thoughts}}

\textbf{Exploring package domains}

I'm thinking that the repository may contain packages that extend beyond just depth levels. I need to identify the specific domains that are included. It seems like a crucial step to ensure I’m looking at everything involved and not missing any important details. It will help in providing a comprehensive understanding of what’s available and how it all connects. Let's figure this out and see where it leads!
\end{thoughtBox}

\begin{thoughtBox}
\faLightbulb\ \underline{\textbf{Thoughts}}

\textbf{Considering relaxation in calculations}

It sounds like I might need a relaxation function since the user mentioned ``loading MLFF models and elasticity calculations'' without a clear workflow. I think we can write a script that imports \texttt{get\_mlff\_calculator} and runs the elasticity calculations. Maybe the elasticity run can be a part of the relaxation process. I wonder if they want something more direct. To help, I might need to check the ground truth columns by using \texttt{read\_csv head}.
\end{thoughtBox}

\begin{thoughtBox}
\faLightbulb\ \underline{\textbf{Thoughts}}

\textbf{Adjusting the workflow}

I think it's better to follow the MLFF workflow by first calling the relaxation function with the parameter \texttt{relax\_cell} set to True. After that, I can run the elasticity on the relaxed atoms while keeping \texttt{relax\_structure} set to False. It’s also important for me to inspect the signature of \texttt{relax\_structure} to ensure everything aligns properly. I want to be accurate in my approach, so this step is crucial!
\end{thoughtBox}

\begin{thoughtBox}
\faLightbulb\ \underline{\textbf{Thoughts}}

\textbf{Developing a script}

I need to create a script that includes direct functions. I'm thinking about using multiprocessing since loading calculations might not share easily and could run sequentially by model. For large datasets, processing 1000 structures might take a long time. I'll add features like resume options, max structures, and default values for norm or shear strain.

It's wise to cache the calculation state to avoid issues with failures, and for each CIF, I can assign \texttt{atoms.calc} to the calculator. Lastly, I want to save the results with necessary columns as requested by the user.
\end{thoughtBox}

\begin{thoughtBox}
\faLightbulb\ \underline{\textbf{Thoughts}}

\textbf{Considering final touches}

I want to finalize things by mentioning the path and maybe including usage details. I’m wondering if we should test the import to ensure everything works as expected. Perhaps running a command like \texttt{--max 0} would be impossible? That thought leads me to consider running \texttt{--help} for more clarity. I definitely want to ensure that I have everything covered before wrapping this up!
\end{thoughtBox}

\begin{thoughtBox}
\faLightbulb\ \underline{\textbf{Thoughts}}

\textbf{Finalizing the script}

I need to make sure the final version is concise. I wonder if I should include any harmless warnings, but I'm not sure it’s necessary unless the user cares about them. Maybe I could add a filter for those warnings? It's not critical. I definitely want to include the location commands and mention the functions directly. I’ll note that no citations are needed either. Let's get this wrapped up!
\end{thoughtBox}

Yes --- I wrote the direct benchmark script here:

\path{<workspace_path_redacted>/benchmark_mlff_elasticity_direct.py}

It \textbf{does not call the full \texttt{run\_mlff\_workflow}}. It directly uses:

\texttt{from domains.mlffs.mlff\_calculator import get\_mlff\_calculator}\par
\texttt{from domains.mlffs.ase\_utils import relax\_structure, run\_elasticity}

The script:

\begin{itemize}
\tightlist
\item Loads each MLIP model once:
  \begin{itemize}
  \tightlist
  \item \texttt{MatterSim}
  \item \texttt{Orb-OMAT}
  \item \texttt{MACE-OMAT}
  \end{itemize}
\item Reads CIFs from:
\end{itemize}

\path{<repository_path_redacted>/potente/User_Cases_El_Agente_Potente/User_Cases/High_Throughput/Elasticity/sample_1000/cifs/}

\begin{itemize}
\tightlist
\item Uses each CIF filename stem as the \texttt{mp\_id}.
\item Optionally relaxes each structure with \texttt{relax\_structure}.
\item Runs elasticity with \texttt{run\_elasticity}.
\item Saves one CSV per model with at least:
\end{itemize}

\texttt{mp\_id, bulk\_modulus, shear\_modulus}

\begin{itemize}
\tightlist
\item Loads ground truth from:
\end{itemize}

\path{<repository_path_redacted>/potente/User_Cases_El_Agente_Potente/User_Cases/High_Throughput/Elasticity/sample_1000/elasticity_1000.csv}

\begin{itemize}
\tightlist
\item Generates a parity plot comparing MLIP-calculated values to ground truth.
\end{itemize}

To run a quick smoke test:

\texttt{python benchmark\_mlff\_elasticity\_direct.py --max-structures 5}

To run the full benchmark:

\texttt{python benchmark\_mlff\_elasticity\_direct.py}

By default, outputs go to:

\path{./mlip_elasticity_direct_benchmark/}

Expected outputs include:

\path{mlip_elasticity_direct_benchmark/}\par
\texttt{|-- MatterSim\_elasticity\_results.csv}\par
\texttt{|-- Orb-OMAT\_elasticity\_results.csv}\par
\texttt{|-- MACE-OMAT\_elasticity\_results.csv}\par
\texttt{|-- MatterSim\_merged\_with\_ground\_truth.csv}\par
\texttt{|-- Orb-OMAT\_merged\_with\_ground\_truth.csv}\par
\texttt{|-- MACE-OMAT\_merged\_with\_ground\_truth.csv}\par
\texttt{|-- mlip\_elasticity\_metrics.csv}\par
\texttt{`-- mlip\_elasticity\_parity.png}

Useful options:

\texttt{\# Force CPU}\par
\texttt{python benchmark\_mlff\_elasticity\_direct.py --no-gpu}

\texttt{\# Use a specific GPU}\par
\texttt{python benchmark\_mlff\_elasticity\_direct.py --gpu-id 0}

\texttt{\# Skip initial relaxation}\par
\texttt{python benchmark\_mlff\_elasticity\_direct.py --no-relax}

\texttt{\# Re-run from scratch instead of resuming existing CSVs}\par
\texttt{python benchmark\_mlff\_elasticity\_direct.py --no-resume}

\texttt{\# Write relaxed CIFs}\par
\texttt{python benchmark\_mlff\_elasticity\_direct.py --write-relaxed-cifs}
}

\paragraph{\textbf{Execution summary.}}
\potente{} inspected the installed MLFF packages and relevant function signatures, identified \texttt{get\_mlff\_calculator}, \texttt{relax\_structure}, and \texttt{run\_elasticity}, and wrote a direct benchmark script with resume, device-selection, and output options. The eighteen shell calls and their return payloads are omitted from this abridged transcript.

\prompt{
please proceed
}

\potentereslong{
\begin{thoughtBox}
\faLightbulb\ \underline{\textbf{Thoughts}}

\textbf{Planning benchmark execution}

I think we need to run a script for a benchmark of $1000*3$, which sounds pretty intense. Should I start this now? I need to use the commentary tool and possibly run it in the background to avoid timeouts. The user is asking for us to proceed. I’m considering executing with nohup to keep it running smoothly, and maybe report the PID and log for reference. A smoke test could be a good idea before we launch the full benchmark in the background. I need to check the working directory too.
\end{thoughtBox}

\begin{thoughtBox}
\faLightbulb\ \underline{\textbf{Thoughts}}

\textbf{Evaluating import issues}

I think I need to address slow warnings. It seems like I shouldn’t continue buffering until the imports are complete and the model is loaded. I'm wondering if the process is getting stuck during the download phase. I might have to check the CPU usage of the process to see if that's contributing to the slowdown. It's definitely important to keep an eye on these factors to ensure everything runs smoothly!
\end{thoughtBox}

Proceeding.

I started the full direct MLIP elasticity benchmark in the background.

\texttt{PID: 238664}

Output directory:

\path{<workspace_path_redacted>/mlip_elasticity_direct_benchmark}

Log file:

\path{<workspace_path_redacted>/mlip_elasticity_direct_benchmark/benchmark_20260609_181646.log}

You can monitor it with:

\path{tail -f <workspace_path_redacted>/mlip_elasticity_direct_benchmark/benchmark_20260609_181646.log}

Check whether it is still running:

\texttt{ps -p 238664}

The process is currently running. Outputs will be written under:

\path{<workspace_path_redacted>/mlip_elasticity_direct_benchmark/}

Expected final files include:

\path{MatterSim_elasticity_results.csv}\par
\path{Orb-OMAT_elasticity_results.csv}\par
\path{MACE-OMAT_elasticity_results.csv}\par
\path{mlip_elasticity_metrics.csv}\par
\path{mlip_elasticity_parity.png}
}

\paragraph{\textbf{Execution summary.}}
\potente{} smoke-tested the direct benchmark, launched the full three-model calculation as a background process, and checked that it remained active. The three shell calls and their return payloads are omitted from this abridged transcript.

\prompt{
I think the unit between predicted and ground true is different, the predict values are in eV/A3, while the ground true is in GPa
}

\potentereslong{
\begin{thoughtBox}
\faLightbulb\ \underline{\textbf{Thoughts}}

\textbf{Adjusting calculations}

I need to respond and consider modifying the script since the user mentioned a different unit. I should verify the code to ensure it produces results in eV/A3. It looks like I might need to inspect older result values and potentially use bash for verification. I also have to adjust the script to convert eV/A3 to GPa, using a factor of 160.21766208. I should check the status of the process and make necessary adjustments to ensure everything runs smoothly.
\end{thoughtBox}

\begin{thoughtBox}
\faLightbulb\ \underline{\textbf{Thoughts}}

\textbf{Organizing data outputs}

I need to clarify things by creating corrected CSV files named like \path{MatterSim_elasticity_results_GPa.csv}, along with a corrected parity plot. I should also set the script to write in GPa by default moving forward. It might be a good idea to back up the old raw files too. I'll need to utilize the conversion factor from ASE units, checking if I can import \texttt{ase.units.GPa} in the process. Then, I'll execute the necessary unit conversions.
\end{thoughtBox}

Yes, you’re right. The direct \texttt{run\_elasticity} values are in \textbf{eV/\AA$^3$}, while \path{elasticity_1000.csv} is in \textbf{GPa}.

I applied the conversion:

\texttt{1 eV/\AA$^3$ = 160.21766208 GPa}

I also patched the benchmark script so future runs convert predictions to GPa automatically while preserving the raw eV/\AA$^3$ values.

Corrected outputs are now saved here:

\path{<workspace_path_redacted>/mlip_elasticity_direct_benchmark/}

New corrected CSVs:

\path{MatterSim_elasticity_results_GPa.csv}\par
\path{Orb-OMAT_elasticity_results_GPa.csv}\par
\path{MACE-OMAT_elasticity_results_GPa.csv}

Each contains:

\texttt{mp\_id}\par
\texttt{bulk\_modulus}\hfill\texttt{\# GPa}\par
\texttt{shear\_modulus}\hfill\texttt{\# GPa}\par
\texttt{bulk\_modulus\_vrh}\hfill\texttt{\# GPa}\par
\texttt{shear\_modulus\_vrh}\hfill\texttt{\# GPa}\par
\texttt{bulk\_modulus\_ev\_per\_a3}\hfill\texttt{\# raw}\par
\texttt{shear\_modulus\_ev\_per\_a3}\hfill\texttt{\# raw}

Corrected plot:

\path{mlip_elasticity_parity_GPa.png}

Corrected metrics:

\path{mlip_elasticity_metrics_GPa.csv}

Summary metrics after unit correction:

\texttt{MatterSim bulk:     MAE = 13.99 GPa, RMSE = 27.49 GPa, R$^2$ = 0.873}\par
\texttt{MatterSim shear:    MAE = 14.05 GPa, RMSE = 27.13 GPa, R$^2$ = 0.663}

\texttt{Orb-OMAT bulk:      MAE = 16.88 GPa, RMSE = 28.50 GPa, R$^2$ = 0.863}\par
\texttt{Orb-OMAT shear:     MAE = 13.94 GPa, RMSE = 28.67 GPa, R$^2$ = 0.624}

\texttt{MACE-OMAT bulk:     MAE = 15.69 GPa, RMSE = 27.70 GPa, R$^2$ = 0.871}\par
\texttt{MACE-OMAT shear:    MAE = 15.24 GPa, RMSE = 34.99 GPa, R$^2$ = 0.439}

I left the original unconverted files in place, but the \path{_GPa.csv} files and \path{_GPa.png} plot are the corrected ones to use.
}

\paragraph{\textbf{Execution summary.}}
\potente{} verified the unit mismatch, patched the direct benchmark to convert eV/\AA$^3$ predictions to GPa using 160.21766208, produced corrected per-model CSVs and parity metrics, and retained the raw values and original outputs. The four shell calls and their return payloads are omitted from this abridged transcript.

\prompt{
can you update the plot?
}

\potentereslong{
\begin{thoughtBox}
\faLightbulb\ \underline{\textbf{Thoughts}}

\textbf{Updating the plot}

Alright, so the user is asking for an updated plot, and I think we may need to overwrite the original plot too. I should use the tool to copy the GPa plot to the canonical name or create a nicer version.

When they say ``updated plot,'' it probably means replacing the current \path{mlip_elasticity_parity.png} with the corrected GPa version. I’ll go ahead and do that, and I’ll also think about capturing some additional info. I’ll use bash for this!
\end{thoughtBox}

Done --- I updated the default plot to use the \textbf{GPa-corrected predictions}.

Updated plot:

\path{<workspace_path_redacted>/mlip_elasticity_direct_benchmark/mlip_elasticity_parity.png}

The GPa-specific copy is also still available:

\path{<workspace_path_redacted>/mlip_elasticity_direct_benchmark/mlip_elasticity_parity_GPa.png}

I backed up the original unconverted plot as:

\path{<workspace_path_redacted>/mlip_elasticity_direct_benchmark/mlip_elasticity_parity_eV_per_A3_original.png}

The default metrics file was also updated to the GPa-corrected version:

\path{<workspace_path_redacted>/mlip_elasticity_direct_benchmark/mlip_elasticity_metrics.csv}
}

\paragraph{\textbf{Execution summary.}}
\potente{} replaced the canonical parity plot and metrics file with their GPa-corrected versions while retaining both the explicit GPa copy and a backup of the original unconverted plot. The shell call and its return payload are omitted from this abridged transcript.

\end{document}